\documentclass[journal]{IEEEtran}
\IEEEoverridecommandlockouts
\usepackage{cite}
\usepackage{amsmath,amsthm,amsfonts,amssymb, bm,mathrsfs}
\usepackage{units}
\usepackage{xr}
\usepackage{boondox-calo}
\usepackage{algorithmic}
\usepackage{algorithm}
\usepackage{booktabs}
\usepackage{graphicx}
\usepackage{subfig}
\usepackage{tabularx}
\usepackage{caption}
\usepackage{multirow}
\usepackage{threeparttable} 
\usepackage{float} 
\usepackage{wasysym,makecell}
\usepackage{textcomp}
\usepackage{xcolor}
\def\BibTeX{{\rm B\kern-.05em{\sc i\kern-.025em b}\kern-.08em
    T\kern-.1667em\lower.7ex\hbox{E}\kern-.125emX}}
\makeatletter
\newcommand*{\addFileDependency}[1]{
  \typeout{(#1)}
  \@addtofilelist{#1}
  \IfFileExists{#1}{}{\typeout{No file #1.}}
}
\makeatother
\newcommand*{\myexternaldocument}[1]{%
    \externaldocument{#1}%
    \addFileDependency{#1.tex}%
    \addFileDependency{#1.aux}%
}
\myexternaldocument{supp}
    
\begin{document}

\title{LLM-Empowered Agentic MAC Protocols: A Dynamic Stackelberg Game Approach}

\author{Renxuan Tan, Rongpeng Li, Fei Wang, Chenghui Peng, Shaoyun Wu, Zhifeng Zhao, and Honggang Zhang
\thanks{R. Tan and R. Li are with the College of Information Science and Electronic Engineering, Zhejiang University, Hangzhou 310027, China. (email: \{ttrx, lirongpeng\}@zju.edu.cn). 

Fei Wang, Chenghui Peng, and Shaoyun Wu are with Huawei Technologies Company Ltd., Shanghai 210026, China. (email: \{wangfei76, pengchenghui, wushaoyun\}@huawei.com).

Z. Zhao is with Zhejiang Lab as well as Zhejiang University, Hangzhou 310027, China. (email: zhaozf@zhejianglab.org). 

H. Zhang is with Macau University of Science and Technology, Macau, China (email: hgzhang@must.edu.mo).

Corresponding Author: Rongpeng Li
}
}
\newtheorem{theorem}{Theorem}
\newtheorem{definition}{Definition}
\newtheorem{assumption}{Assumption}
\newtheorem{lemma}{Lemma}
\newtheorem{proposition}{Proposition}
\newtheorem{corollary}{Corollary}
\renewcommand{\qedsymbol}{$\blacksquare$}

\newcommand\subscriptuet[1][i]{_{#1,t,{ u}}}
\newcommand\subscriptbst[1][t]{_{#1,{ b}}}
\newcommand\subscriptit[1][i]{_{#1,t}}
\newcommand\subscriptiu[1][i]{_{#1,u}}

\newcommand\subscriptbsuet[2][t]{^{{ b},u_{#2}}_{#1}}
\newcommand\UEn[1][i]{{u}_{#1}}
\newcommand\BS{\text{BS}}

\newcommand\ueact[1][i]{a^{\rm bit}_{{#1},t,{ u}}}
\newcommand\uemsg[1][i]{u_{{#1},t}}
\newcommand\bsmsg[1][i]{d_{{#1},t}}
\newcommand\bspolicy{\pi_b}
\newcommand\uepolicy[1][i]{\pi_{{#1},u}}

\newcommand{\tabincell}[2]{\begin{tabular}{@{}#1@{}}#2\end{tabular}}
\maketitle

\begin{abstract}
Medium Access Control (MAC) protocols, essential for wireless networks, are typically manually configured. While deep reinforcement learning (DRL)-based protocols enhance task-specified network performance, they suffer from poor generalizability and resilience, demanding costly retraining to adapt to dynamic environments. To overcome this limitation, we introduce a game-theoretic LLM-empowered multi-agent DRL framework, in which the uplink transmission between a base station and a varying number of user equipments is modeled as a dynamic multi-follower Stackelberg game (MFSG), capturing the network's natural hierarchical structure. {\color{black}Within this game, PPO-coordinated LLM agents synthesize task-grounded semantic MAC protocols for intent-aware coordination under network dynamics.} Protocol action grammar (PAG) is employed to ensure the reliability and efficiency of this process. Further, we theoretically establish the existence, convergence, and robustness of a Stackelberg equilibrium, ensuring stable learning dynamics of LLM-empowered policies under a changing number of followers. Simulations corroborate that our framework achieves up to $81.1\%$ and $46.6\%$ throughput gains over heuristic and state-of-the-art DRL baselines, respectively. Besides, it generalizes excellently to a fluctuating number of users without retraining or architectural changes. 
\end{abstract}




\section{Introduction}
\IEEEPARstart{T}he evolution towards next-generation (xG) wireless systems envisions artificial intelligence (AI)-native architectures wherein intelligent, resilient communication protocols autonomously emerge to manage unprecedented network dynamics \cite{Resilientsystem}. Central to this vision is the medium access control (MAC) protocol, which orchestrates channel access among numerous nodes. As network topologies become increasingly varying and heterogeneous, the prevailing paradigm of designing static, human-engineered MAC protocols is rendered obsolete, necessitating protocol emergence solutions that can learn and adapt in real-time \cite{10955466}. This need is particularly acute in challenging scenarios such as uplink data transmission scheduling (UDTS) within cellular networks \cite{MP-0, MP-maddpg}. 

Deep reinforcement learning (DRL) and its multi-agent counterpart (MARL) have become a promising approach in this domain, modeling network nodes -- such as base stations (BSs) and user equipments (UEs) -- as autonomous agents that learn coordinated control messages and policies through trial-and-error interaction \cite{MP-scale,mostafa2025,drx, FeasibleMAC}. Despite substantial potential for enhancing network key performance indicators (KPIs), up-to-date formulations typically resort to fully centralized or fully decentralized control paradigms but neglect the intrinsic game-theoretic relationships among network entities \cite{Massive,llm4mac}. 
Meanwhile, the UDTS in cellular networks is inherently characterized by resource competition under centralized coordination. Therefore, compared to alternative cooperative games \cite{9207866}, the Stackelberg game (SG), a hierarchical leader-follower game, can grant the flexibility of UEs under the guidance of the BS. Moreover, it can elegantly capture the underlying dynamics \cite{10264114} and guide policy convergence in MARL frameworks \cite{9444840}, rendering it a more natural and faithful solution. However, generalization in dynamic network environments remains a critical bottleneck \cite{Semantic-paper}, as there commonly emerges a fluctuating number of agents (i.e., attached UEs).

Recently, characterized by impressive comprehension and generative reasoning capabilities, large language models (LLMs) have emerged as a compelling solution to intractable protocol design challenges \cite{WiLLM}. Their inherent ability for processing variable-length sequences suggests a natural aptitude for operating in dynamic game environments, overcoming the architectural rigidity of non-LLM methods \cite{llm4mac}. Nevertheless, instantiating SG agents with LLMs introduces unique challenges. A fundamental discrepancy arises between the pre-training objective of an LLM -- generating plausible text -- and the strategic reasoning needed for a leader or follower in an SG, which is to maximize network utility. Addressing this significant gap requires functional alignment of model behavior with strategic goals. Existing LLM-based systems, however, typically depend on meticulously curated, domain-specific datasets or expert prompt engineering for domain adaptation \cite{LLM4Comunications}. This dependency constrains their capacity for exploratory learning, the very mechanism required for the emergence of novel, high-performance protocols. Furthermore, bridging the semantic gap between the LLMs' open-domain vocabulary and the game action space of MAC protocols is essential yet nontrivial. Reconciling generalization, exploration, and structured action generation is therefore critical \cite{GroundingLLM} for achieving stable and efficient policy convergence in such dynamic game settings. 

To overcome the limitations above, we propose an LLM-empowered MARL framework that conceptualizes the MAC protocol emergence for UDTS as a dynamic multi-follower Stackelberg game (MFSG). This game is instantiated through LLM-empowered agents, where the BS serves as the leader agent and broadcasts a macroscopic scheduling intent as downlink control messages (DCMs). The UEs, acting as followers, interpret this intent in the context of their local observations and articulate their best responses as uplink control messages (UCMs). Crucially, we reformulate the MARL paradigm by endowing both the BS and UEs with language-oriented policy models that naturally handle variable-length input and output sequences while retaining the exploratory capabilities of DRL. We employ proximal policy optimization (PPO) to facilitate a continuous, feedback-driven alignment of the agents' policies with evolving network dynamics. This synergistic design overcomes the architectural rigidity of conventional MARL and mitigates the data dependency of LLMs. We also incorporate a protocol action grammar (PAG) to guarantee reliability and efficiency by enforcing domain-specific constraints on agent decisions. To this end, we theoretically analyze the convergence of an equilibrium within this learning framework. We summarize the key differences between our algorithm and highly related literature in Table \ref{tab: summary}. 

Our key contributions are threefold, as follows. 
\begin{itemize}
	\item We establish a dynamic MFSG formulation for MAC protocol emergence in the UDTS scenario, explicitly capturing the inherently hierarchical interactions between BS and a dynamically varying number of UEs. As a pioneering effort, this game-theoretic model fundamentally differs from existing paradigms \cite{MP-0, MP-maddpg} by formulating asymmetric leader-follower relationships.
	\item We propose a novel LLM-empowered MARL framework that fundamentally reconciles the powerful generalization capabilities of LLMs with the exploratory learning essential for protocol discovery, driven by PPO. The concerted effort makes our design uniquely capable of adapting to a dynamic network without retraining. It also incorporates a PAG mechanism to further enhance efficiency and reliability. 
	\item We provide a theoretical analysis of our framework, establishing not only the guaranteed existence and local convergence of a Stackelberg equilibrium (SE) in the dynamic game, but also the robustness against information disparities. Extensive simulations have validated the superiority and robustness	of our proposed algorithm.
\end{itemize}

This paper is structured as follows. Section II details the related works. Section III presents the system model and formulates the problem. Section IV details the LLM-based decision maker and its PPO training. Section V evaluates performance via simulations, and the main conclusion is drawn in Section VI.
\section{Related Works}
\begin{table*}[tb]
	\centering
	\caption{The Summary of Differences with Related Literature on Protocol Emergence.}
	\label{tab: summary}
	\begin{tabular}{l|cccc|l} 
		\toprule
		& \tabincell{c}{Exploratory \\ Protocol Discovery} &  \tabincell{c}{Dynamic\\ Adaptability} & \tabincell{c}{Intrinsic \\ Hierarchy Modeling} & \tabincell{c}{Semantic \\ Protocols} & Brief Description \\
		
		\hline
		\tabincell{l}{\cite{MP-maddpg,MP-scale,mostafa2025,Massive}} & $\CIRCLE$ & $\RIGHTcircle$ & $\Circle$ & $\Circle$ & \tabincell{l}{Vanilla DRL for protocol emergence}   \\
		\hline
		
		\tabincell{l}{\cite{MP-new}\\\cite{ProtocolLLM,llmRANprotocol}} & \tabincell{l}{$\Circle$\\$\Circle$} & \tabincell{l}{$\CIRCLE$\\$\CIRCLE$} & \tabincell{l}{$\CIRCLE$\\$\Circle$} & \tabincell{l}{$\Circle$\\$\CIRCLE$} & \tabincell{l}{Static rules or datasets reliance result \\in insufficient exploration}\\
		\hline
		
		\tabincell{l}{\cite{FeasibleMAC,MultitaskMultipleAccess}} & $\CIRCLE$ & $\CIRCLE$ & $\RIGHTcircle$ & $\Circle$ & Ignoring signaling interpretability\\
		\hline
		
		\tabincell{l}{\cite{Kim2024KnowledgeDF,prolog,CPAgentNet}\\ \cite{kim2025resilientllmempoweredsemanticmac}} & \tabincell{l}{$\CIRCLE$\\$\CIRCLE$} & \tabincell{l}{$\RIGHTcircle$\\$\CIRCLE$} & \tabincell{l}{$\Circle$\\$\Circle$} & \tabincell{l}{$\CIRCLE$\\$\CIRCLE$} & \tabincell{l}{Ignoring intrinsic game theoretic \\formulation among network agents}\\
		\hline
		
		\textbf{Ours} & $\CIRCLE$ & $\CIRCLE$ & $\CIRCLE$ & $\CIRCLE$ & \tabincell{l}{Synergizes a dynamic Stackelberg\\ game model with an LLM-RL policy \\to learn semantic emergent protocols \\that are adaptive to network dynamics}\\
		
		\toprule
		
		\multicolumn{5}{>{\footnotesize\itshape}r}{Notations: \rm{${\Circle}$} \emph{indicates not included;} \rm{${\CIRCLE}$} \emph{indicates fully included;} \rm{${\RIGHTcircle}$} \emph{means partially included.}}
	\end{tabular}
\end{table*}
\subsection{DRL for Communication Protocol}
Leveraging its data-driven exploration and trial-and-error mechanisms, DRL is suited to optimize and even discover protocols for task-specific applications. One research line focuses on optimizing the functional blocks and operational parameters of existing protocols \cite{KeshtiarastML, 10758702, SpecNets, QLBT, MultitaskMultipleAccess}. For example, Ref. \cite{KeshtiarastML,10758702} introduces DRL to automatically revise the features of protocol blocks in response to various network conditions, demonstrating superior throughput and latency compared to traditional methods. Ref. \cite{SpecNets} proposes specialized wireless networks (SpecNets), wherein agents dynamically adjust settings such as contention windows and frame aggregation to meet diverse performance demands. Another research thrust focuses on augmenting the adaptability of communication protocols. Ref. \cite{QLBT} demonstrates the adaptability of the MARL-based protocol across various dynamic network conditions and coexistence scenarios. Ref. \cite{MultitaskMultipleAccess} employs multi-task learning and a scalable transformer-based critic to train a single, generalizable model that can adapt to dynamic network sizes and traffic conditions. In contrast, several works have considered protocol emergence through training from scratch. Ref. \cite{MP-0} establishes a DRL framework to learn channel access policies and signaling simultaneously. Subsequent works extend this with more advanced MARL algorithms and optimization tricks \cite{FeasibleMAC, MP-scale, Massive, mostafa2025}, achieving notable progress in the convergence, adaptability, and scalability of emergent protocols. Ref. \cite{prolog} even explores interpretability by encoding symbolic structure via a probabilistic logic programming language (ProbLog). Nevertheless, a persistent challenge for existing DRL-based solutions is their limited generalization \cite{CPAgentNet}, rendering them brittle in highly dynamic network environments and often necessitating costly retraining or architectural redesigns.

\subsection{LLMs for Communication Protocol}
The advent of LLMs is set to revolutionize wireless protocol design. Benefiting from their remarkable capabilities in understanding, reasoning, and generalization, alongside their extensive general knowledge bases, LLMs have achieved considerable success in various downstream tasks within wireless networking \cite{chen_netgpt_2024}. On the one hand, LLMs can serve as base models that are fine-tuned on communications-specific datasets to support resource management \cite{llmempoweredresource}, standard-protocol interpretation \cite{WirelessLLM}, and protocol design \cite{ProtocolLLM,llmRANprotocol}. On the other hand, LLMs also facilitate direct zero-shot or few-shot domain adaptation. Ref. \cite{Kim2024KnowledgeDF} leverages knowledge acquired via in-context learning and uses knowledge distillation to guide and accelerate protocol emergence in multi-agent navigation systems. Recently, Ref. \cite{kim2025resilientllmempoweredsemanticmac} employs prompt engineering to auto-tune LLM system prompts for generating semantic signaling tokens for BSs and UEs, enabling zero-shot adaptation to environmental dynamics. Ref. \cite{CPAgentNet} further showcases the coordination of agentic LLM workflows for autonomous and interpretable protocol design. While effective in achieving high performance, LLMs without post-training remain path-dependent on domain-specific datasets and highly engineered prompts, which constrain their ability to discover fundamentally new protocols through exploration. Our previous work \cite{llm4mac} makes an initial attempt to reconcile exploration with generalization by fine-tuning an LLM-based policy using PPO. Yet, it fell short of capturing the intrinsic game-theoretic interactions among agents.

\subsection{Stackelberg Game in Wireless Communication}
{\color{black}Distinct from broader computational game-theoretic paradigms, such as evolutionary games that typically characterize symmetric population-level adaptation \cite{Comment5}, SG features an intrinsic leader-follower structure and has been widely used to model hierarchical strategic interactions in wireless networks \cite{9099033,9582739,10530205}.} In \cite{9099033}, the SG framework effectively models the pricing strategies between service providers (leaders) and end-users (followers) in complex multi-tenant network slicing environments, demonstrating faster convergence and greater economic efficiency. Ref. \cite{9582739} employs an SG to model energy trading between a power station and a transmitter in an intelligent reflection surface-aided wireless-powered multicast system. The SG framework also delineates the fundamental dynamics governing network interactions and even accelerates convergence \cite{9444840}, guiding agents toward a more efficient equilibrium. However, to our best knowledge, MFSG has not been leveraged to develop MAC protocols.
\begin{figure}[tb]
	\centering
	\includegraphics[width=0.9\linewidth]{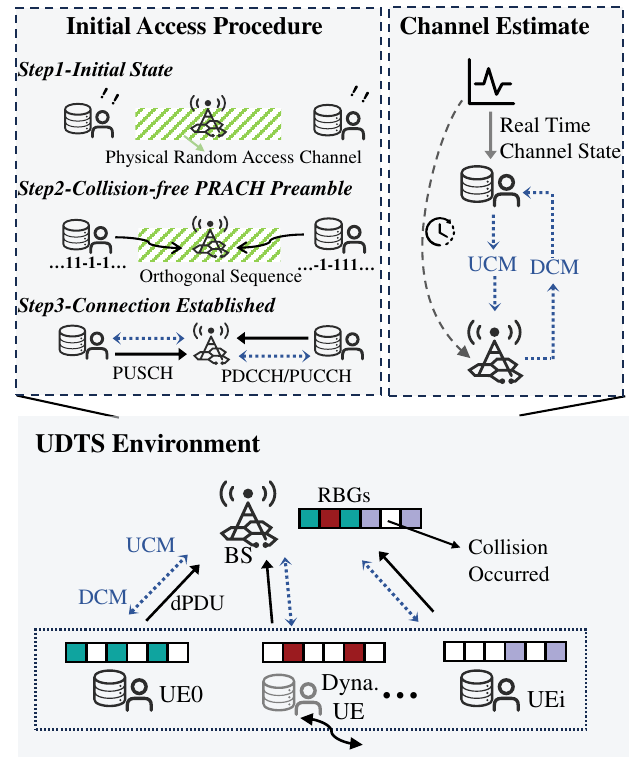}
	\caption{Scenario of interest: A BS serves dynamic UEs within an OFDMA system, UEs select RBGs for dPDU transmission, coordinating with the BS via UCMs and DCMs.}
	\label{fig:framework}
\end{figure}
\begin{table}[tb]
	\caption{Notations Mainly Used in This Paper.}
	\label{tab:notations}
	\begin{tabularx}{\columnwidth}{cX}
		\toprule
		Notation & Description  \\
		\midrule
		$I_t$, $\mathcal{I}_t$ & Number of active UEs and the set of UEs at time slot $t$ \\ 
		$M$ & Number of RBGs \\
		$T$ & Number of TTIs per episode\\
		$\xi^{\rm gen}$ & Generated dPDU size\\
		$\xi^{\rm rec}\subscriptuet$, $\xi^{\rm tx}\subscriptuet$ & dPDUs successfully received and total transmission attempts by UE $\UEn$ at time slot $t$\\
		$u\subscriptit$, $d\subscriptit$ & UCM and DCM of UE $\UEn$ at time slot $t$  \\
		$b\subscriptit$ & Buffer occupancy of UE $\UEn$ at time slot $t$  \\
		$\nu\subscriptit$ & Spectral efficiency of UE $u_i$ at time slot $t$  \\
		$c\subscriptuet$, $\hat{c}\subscriptuet$ & Locally channel quality of UE $u_i$ at time slot $t$ and corresponding BS's channel state estimate\\
		$x\subscriptuet$ & Cumulative RBGs utilized by UE $\UEn$ up to time $t$  \\
		$o\subscriptbst$, $\tilde{o}\subscriptbst$ & Observations of the BS at time slot $t$ \\
		$o\subscriptuet$, $\tilde{o}\subscriptuet$ & Observations of the UE $u_i$ at time slot $t$ \\
		$a\subscriptbst$, $\tilde{a}\subscriptbst$ & BS's action at time slot $t$, i.e., DCMs to UEs \\
		$a\subscriptuet$, $\tilde{a}\subscriptuet$ & UE $u_i$'s action at time slot $t$, comprising a transmission bitmap $a^{\rm bit}\subscriptuet$ and a UCM \\		
		$F\subscriptbst$ & Utility function of the leader (BS) at time slot $t$ \\
		$F\subscriptuet$ & Utility function of the follower UE $u_i$ at time slot $t$ \\
		$\pi_b$, $\pi\subscriptiu$ & Policies for the BS (leader) and UE $\UEn$ (follower) \\
		$\mathbb{D}(\cdot)$ & The probability distribution of the number of active UEs\\
		$\theta_b$, $\theta_u$ & Policy parameters for the BS and the shared UE network  \\
		$J_b(\cdot)$, $J\subscriptiu(\cdot)$ & Expected long-term utilities for the BS and UE $u_i$  \\
		$\phi$, $L(\cdot)$ &  Parameters of critic network and TD error for its update \\
		$V(\cdot)$, $A(\cdot)$ & State-value function and advantage function \\
		$\gamma$, $\lambda$ & Discount factor for long-term rewards and GAE \\
		$\alpha$, $\beta$ & Learning rate for actor and critic update \\
		$e$  & Training epoch \\
		$\psi_b$, $\psi_u$ & Meta-prompts for the BS and UE LLM agents \\
		$\tau_b$, $\tau\subscriptiu$ &Trajectory rollout of BS and UEs \\
		$\mathcal{B}_b$, $\mathcal{B}_u$ & Trajectory replay buffers for the leader and followers\\
		$\mathcal{W}$, $\mathcal{W}^-$ & LLM vocabulary and PAG-constrained vocabulary\\
		$\bm{\varTheta}$, $\bm{\Omega}$ & Joint policy parameters and update gradient field\\
		
		\bottomrule
	\end{tabularx}
\end{table}
\section{System Model and Problem Formulation}
\subsection{System Model}
\label{sec: system model}
\subsubsection{Basics of UDTS}

As illustrated in Fig. \ref{fig:framework}, we investigate the uplink data transmission scheduling (UDTS) within a single-cell orthogonal frequency division multiple access (OFDMA) system where a BS serves multiple UEs. The set of $I$ active UEs is denoted by $\mathcal{I}_t = \{\UEn[1], \cdots, \UEn[I]\}$ at time $t$, where $I$ may vary due to the arrivals and departures of UEs, reflecting the dynamic nature of the network topology. Consistent with 3GPP TS 38.321 \cite{3GPP38321}, the BS assigns each UE dedicated physical uplink/downlink control channels (PUCCH/PDCCH) for signaling and a shared contention-based channel (PUSCH) for MAC data protocol data unit (dPDU) transmission. The available bandwidth is divided into $M$ orthogonal resource block groups (RBGs) in the frequency domain, and time is discretized into slots of fixed duration, termed as transmission time intervals (TTIs). An operational cycle, or episode, spans $T$ TTIs. This scenario necessitates efficient MAC protocols that jointly optimize signaling and access strategies to mitigate RBG collisions while balancing KPIs like throughput, resource efficiency, and fairness. At time slot $t$, UE $\UEn$ constructs and transmits a UCM $\uemsg$, encapsulating its local state (e.g., buffer status report and interaction history) and transmission intent. In response, the BS integrates these UCMs with its global observations to issue a DCM $\bsmsg$ to each UE, conveying its macroscopic scheduling intent. These signaling exchanges, typically conveyed over the dedicated PUCCH and PDCCH, are crucial for coordinating access and establishing the operational context for data delivery.

\subsubsection{Delivery Model}
We model the arrival of new data at each UE as a Bernoulli process \cite{MP-0, MP-scale}, where a dPDU of a fixed size $\xi^{\rm gen}$ (in bits) is generated with probability $p_a$ in each TTI. Meanwhile, each UE is equipped with a first-in-first-out (FIFO) buffer of capacity $B_{\rm cap}$ to temporarily store dPDUs, with the buffer occupancy denoted by $b\subscriptit \in [0, B_{\rm cap}]$. The UE transmission action, denoted by $\ueact$, is an $M$-bit binary bitmap, where each bit indicates whether the $m$-th RBG is selected for data transmission. Following 3GPP TS 38.214 \cite{3GPP38214}, the maximum transport block size (TBS) of $\UEn$ in time slot $t$ is given by
\begin{equation}
	N\subscriptuet =  {\rm tr}(\bm{1}^T\ueact)\cdot N_{\rm rb}^{\rm rbg}\cdot N_{\rm sc}^{\rm rb} \cdot N_{\rm symbl}^{\rm sh} \cdot \nu\subscriptit
\end{equation}
 where $N_{\rm rb}^{\rm rbg}$ is the number of RBs per RBG, $N_{\rm sc}^{\rm rb}$ is the number of subcarriers per RB, $N_{\rm symbl}^{\rm sh}$ is the number of symbols per slot, and $\nu\subscriptit$ denotes spectral efficiency. Crucially, $\nu\subscriptit$ is a time-varying quantity that depends on the instantaneous signal-to-noise ratio (SNR) of the link and the corresponding modulation and coding scheme (MCS) selected by the UE. The system employs an automatic repeat request (ARQ) process for buffer management, ensuring UEs remove bits from their buffers after the BS confirms a successful data reception. 

\subsubsection{Channel Model}
 Without loss of generality, we model the PUSCH as a packet erasure channel with a nonzero transport block error rate (TBLER). If an RB is collision-free and the transmission is error-free, the BS successfully receives the transmitted data. Moreover, the channel quality for UE $\UEn$ at time $t$ is denoted by $c\subscriptuet$, which evolves between several discrete states. Its transitions reflect short-term channel fluctuations (e.g., fast fading) and thus affect the instantaneous spectral efficiency $\nu\subscriptit$. {\color{black}In parallel, the BS acquires corresponding channel state information (CSI), $\hat{c}\subscriptuet$, which may differ from the UE's locally true state. This mismatch between $c\subscriptuet$ and $\hat{c}\subscriptuet$ abstracts the aggregate leader-side information deficit caused by practical CSI acquisition imperfections, such as reciprocity errors, feedback/processing delays, and channel aging, rather than separately modeling their individual PHY-layer sources.}

\subsubsection{Decision Model}
A sequential decision-making process governs both UE-BS signaling and the UE's transmission behavior. 
For each UE $\UEn$, define its observation at time $t$ as $o\subscriptuet=(c\subscriptuet,b\subscriptit,\bsmsg,a_{i,t-1,u})$ whereas the taken action  $a\subscriptuet=(\ueact,\uemsg)$. Meanwhile, the BS's observation is given by $o\subscriptbst=(\{\hat{c}\subscriptuet\}_{i\in \mathcal{I}_t}),\{u_{i,t-1}\}_{i\in \mathcal{I}_t},\{d_{i,t-1}\}_{i\in \mathcal{I}_t})$ and its action is defined as $a\subscriptbst=(\bsmsg[1],\cdots,\bsmsg[I])$. 
{\color{black}Crucially, both BS and UE observations are primarily composed of the preceding interactive UCMs/DCMs, which allows the learned policies to condition subsequent RBG decisions on prior signaling and thereby induce task-grounded operational meanings to optimize system performance.} 
Each agent's behavior is dictated by a policy that maps its observations to a probability distribution over the action space. Let $\pi_b$ be the leader's policy and $\uepolicy$ the follower's policy, such that $a\subscriptbst \sim \pi_b(o\subscriptbst)$ and $a\subscriptuet \sim \uepolicy(o\subscriptuet)$.

\subsection{Stackelberg Game Formulation}
\begin{figure}
    \centering
    \includegraphics[width=\linewidth]{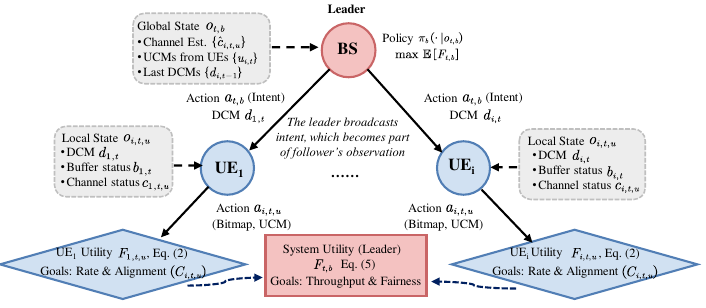}
    \caption{The decision tree of MFSG, which explicitly mapping the variable dependencies and utility formulations.}
    \label{fig: decisiontree}
\end{figure}
\label{sec:SG formulation}
We formally model the sequential and interactive decision-making process as a dynamic MFSG as illustrated in Fig. \ref{fig: decisiontree}. In this game, the BS (leader) enforces macroscopic decisions to optimize network-wide performance, while the UEs (followers) retain the flexibility to optimize fine-grained transmission parameters based on local information and BS's guidance. Our objective is to develop a framework that facilitates the emergence of leader–follower policies, driving the system toward a desirable SE.

As illustrated in Fig. \ref{fig: decisiontree}, the utility of each agent is coupled with the decisions of others\footnote{\color{black}While we formulate UEs as homogeneous expected-utility maximizers to preserve analytical tractability and strict convergence guarantees, future extensions could integrate prospect-theoretic primitives \cite{Comment4} to accommodate bounded rationality and behavioral heterogeneity in more practical scenarios.}. Specifically, the utility for a follower $\UEn\in\mathcal{I}_t$ at time slot $t$ is defined as a weighted sum of transmission efficiency and its consistency with the leader's DCM:
\begin{equation}
	\label{eq:ue unity}
	F\subscriptuet =  \varrho_1 \cdot \frac{\xi\subscriptuet^{\rm rec}}{\xi\subscriptuet^{\rm tx}} + \varrho_2 \cdot C\subscriptuet,
\end{equation}	
where $\varrho_1,\varrho_2$ are constants, $\xi\subscriptuet^{\rm rec}$ denotes the successfully received dPDUs from $\UEn$ at time slot $t$, and $\xi\subscriptuet^{\rm tx}$ represents the total number of transmission attempts made by $\UEn$. $C\subscriptuet$ quantifies the normalized Hamming similarity \cite{Hammingsimilarity} between the UE's transmission bitmap $a^{\rm bit}\subscriptuet$ and the BS's intended allocation bitmap $d^{\rm bit}\subscriptit$ (i.e., the bit-level representation of $d\subscriptit$ and can be directly derived from $a\subscriptbst$). In other words, 
\begin{equation}
	\label{eq: consistency term}
	C\subscriptuet=\frac{1}{M} \sum\nolimits_{m=1}^M\left(1-\left(a_{i, t, u, m}^{\rm bit} \oplus d_{i, t, m}^{\rm bit}\right)\right).
\end{equation}
Given the leader's policy $\pi_b$, each follower seeks to find its best-response policy $\uepolicy^\ast$ that maximizes its expected discounted long-term utility $J\subscriptiu$ over the distribution of UE numbers $\mathbb{D}(I)$,
\begin{equation}
	\label{eq: long-term ue unity}
	J\subscriptiu= \mathbb{E}_{I_t\sim\mathbb{D}(I)}\mathbb{E}_{\uepolicy, \pi_b, \uepolicy[-i]}\left[\sum\nolimits_{t=0}^{T-1} \gamma^t F\subscriptuet\right],
\end{equation}
where $\gamma\in [0,1]$ is the discount factor.
Conversely, the leader's utility $F\subscriptbst$ aligns with the system-level objectives, defined as a weighted sum of the average user throughput and inter-user fairness:
\begin{equation}
	\label{bs:ue unity}
	F\subscriptbst \triangleq \frac{1}{|\mathcal{I}_t|\cdot TTI}\sum_{i\in \mathcal{I}_t} \hat{\xi}\subscriptuet^{\rm rec}+\varepsilon\frac{(\sum_{i\in \mathcal{I}_t} x\subscriptuet)^2}{|\mathcal{I}_t|\cdot\sum_{i\in \mathcal{I}_t}(x\subscriptuet)^2},
\end{equation}
where the $\varepsilon$ balances the scaled throughput against fairness, which is quantified using Jain's Fairness Index (JFI) \cite{jfi}, $x\subscriptuet$ represents the cumulative number of RBGs utilized by UE $\UEn$ up to time $t$. At this equilibrium, the leader's policy is a universal solution to Eq. \eqref{eq: long-term bs unity}, anticipating that all followers will adhere to their best-response policies. 
\begin{equation}
	\label{eq: long-term bs unity}
	 \max_{\pi_b} J_b=\max_{\pi_b}\mathbb{E}_{I_t\sim\mathbb{D}(I)}\mathbb{E}_{\pi_b,\bm{\pi}^\ast_u}\left[\sum\nolimits_{t=0}^{T-1} \gamma^t F\subscriptbst\right],
\end{equation}
where $\bm{\pi}^\ast_u = (\{\pi^\ast\subscriptuet\}_{i\in\mathcal{I}_t})$.

Consequently, the problem is to find a set of policies $(\pi^\ast\subscriptbst,\{\pi^\ast\subscriptuet\}_{i\in\mathcal{I}_t})$ that constitutes an expected Stackelberg equilibrium (ESE) over $\mathbb{D}(I)$, where no player has an incentive to unilaterally deviate from their policy. 
Solving such an equilibrium is nontrivial due to the implicit coupling of utilities and the high-dimensional state–action space, which renders traditional optimization intractable. While MARL is a promising alternative, its fixed architectures fail to generalize to a dynamic environment. To address this, we propose an LLM-driven MARL framework that jointly leverages MARL's exploratory power and LLMs' generalization capability to learn robust equilibria, as detailed in the following section.

\section{PPO-based LLM Decision Maker for Protocol Emergence}
In this section, we recast the search for the SE as an LLM-empowered multi-agent learning process, where all agents operate in a semantic-generalized UTDS environment. Both the BS and UEs leverage parameterized LLM as their decision-making policies. Within each TTI, based on its observation, the BS initiates its action $a\subscriptbst$, and the UEs react accordingly. Ultimately, both agents receive their respective rewards from the environment. PPO with expert seeding is employed to ground the LLMs' performance, ensuring the policies are functionally aligned with the dynamic environment.
\subsection{Framework of RL-based Protocol Emergence}
\label{sec:background on marl}
We begin by outlining the conventional MARL approach for protocol emergence, establishing a baseline, and highlighting its limitations. Following the work \cite{mostafa2025}, the UDTS is formulated as a decentralized partially observable Markov decision process (Dec-POMDP), where agents are trained concurrently to learn both transmission and signaling actions using PPO. The UCM and DCM are treated as special discrete actions of the UEs and the BS, respectively. While these signaling actions do not directly alter the environment, they influence the decisions of other agents by being part of their observable states. Taking the BS as an example, we illustrate its training process. As generally defined by the PPO framework, the BS owns an actor network to represent its policy $\pi_b$ and a critic network $V(o\subscriptbst)$ for its value function, parameterized by $\theta_b$ and $\phi_b$ respectively. The collective goal is to find a policy that maximizes the objective function defined in \eqref{eq: long-term bs unity}. During the interaction with the environment, the BS agent collects a series of experience trajectories $\tau$ composed of transition tuples and 
\begin{equation}
	\label{eq:rollout}
	\tau_b = \left\{ \left(o\subscriptbst,a\subscriptbst,F\subscriptbst,\pi_b(a\subscriptbst \lvert o\subscriptbst),V(o\subscriptbst),A\subscriptbst
	\right)\right\}_{t=0}^{\lvert T-1\rvert}
\end{equation}
captures current observation, action, reward, policy probability, value estimate, and the advantage estimate. The advantage $A_t$ is computed via generalized advantage estimation (GAE) 
\begin{equation}
	\label{eq:GAE}
	A\subscriptbst = \sum\nolimits_{l=0}^{T}(\gamma\lambda)^l(F\subscriptbst[t+l] + \gamma V(o\subscriptbst[t+l+1]) - V(o\subscriptbst[t+l])),
\end{equation}
indicating whether an action taken is better or worse than the policy's average, and $\lambda$ denotes the GAE factor. Updating the critic network $\phi_b$ involves minimizing the squared temporal-difference (TD) residuals
\begin{equation}
	\label{eq:critic_loss}
	L(\phi_b)=\mathbb{E}_{\tau_b\sim\mathcal{B}_b}\Big[ \sum\nolimits_{t=1}^T \big(V(o\subscriptbst)-\hat{R}\subscriptbst\big)^2\Big],
\end{equation}
where $\mathcal{B}_b$ denotes experience buffers, $\hat{R}_t = \sum_{t^\prime = t}^{T} \gamma^{t^{\prime}-t} F\subscriptbst[t^\prime]$ is the discounted reward-to-go. The actor network $\theta_b$\footnote{As the policy $\pi
$ is a direct mapping from $\theta$, updating the actor network's parameters constitutes an update to the policy itself.} is updated via the sampled policy gradient from the clipped surrogate objective function 
\begin{align}
		&J(\theta_b) =\mathbb{E}_{\tau_b\sim\mathcal{B}_b}\bigg[\sum_{t=1}^T \big[\min \big(\eta\subscriptbst A\subscriptbst, \operatorname{clip}(\eta\subscriptbst, 1\!-\!\epsilon, 1\!+\!\epsilon) A\subscriptbst\big) \nonumber\\
		& -D_\text{KL}\big(\bspolicy^{\rm old}(\cdot|o\subscriptbst)\lvert\lvert\bspolicy(\cdot|o\subscriptbst)\big)  + H\big((\bspolicy(\cdot|o\subscriptbst)\big)\big]\bigg],	\label{eq:actor_loss}
\end{align}
where $\eta\subscriptbst =  \frac{\bspolicy(a\subscriptbst \lvert o\subscriptbst)}{\bspolicy^{\rm old}(a\subscriptbst \lvert o\subscriptbst)}$ denotes the important ratio, the operator $D_\text{KL}(\cdot\|\cdot)$ computes the KL-divergence, the policy entropy $H$ encourages exploration, meanwhile both the clipping function $\text{clip}(\cdot)$ and the hyperparameter $\epsilon$ prevent excessively large policy updates.
\subsection{Semantic-Generalized MFSG}
\begin{figure*}[t]
	\centering
	\includegraphics[width=0.75\linewidth]{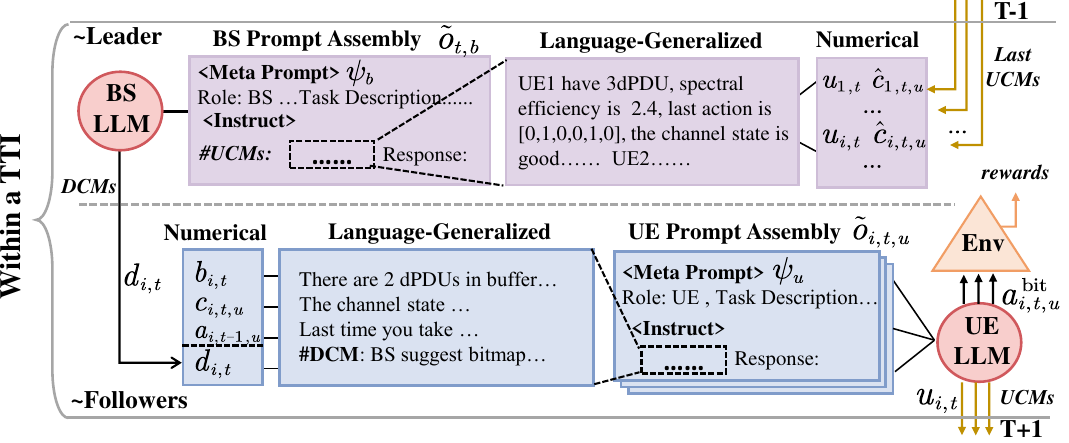}
	\caption{The execution workflow of the LLM-empowered agents. This diagram details the prompt assembly, numeric-to-text conversion, and the generative decision-making process within a TTI.}
	\label{fig:workflow}
\end{figure*}
It can be observed that the traditional policy $\bspolicy$'s input and output dimensions, which process the UCM and DCM vectors, scale with the total number of UEs. This generates an incompatibility with standard fixed-neuron architectures (e.g., multi-layer perceptron (MLP)), as any environment shift may invalidate the learned policy, forcing costly redesign and retraining from scratch. To address this kind of generalization challenge in dynamic environments, we take advantage of LLMs, which inherently process variable-length input and output sequences. Under this paradigm, fluctuations in the number of UEs merely alter the prompt's length rather than disrupting the fundamental training pipeline. 

Specifically, we introduce a natural language vocabulary $\mathcal{W}$ and a mapping function set $\bm{f}$ to adapt to the linguistic modality. We assume both the BS and UEs utilize the same family of LLMs, thereby sharing $\mathcal{W}$. As depicted in Fig. \ref{fig:workflow}, raw numerical observations $o\subscriptbst,o\subscriptuet$ are transformed into a comprehensive LLM-processable prompt through numeric-to-text conversion and meta-prompt assembly:
\begin{subequations}
\label{eq: pag embedding}
    \begin{align}
	\Tilde{o}\subscriptuet &\triangleq f_{o,u}(o\subscriptuet) \oplus \psi_u ,\ \Tilde{o}\subscriptuet \subset \mathcal{W}\\
	\Tilde{o}\subscriptbst &\triangleq f_{o,b}(o\subscriptbst) \oplus \psi_b ,\ \Tilde{o}\subscriptbst \subset \mathcal{W},
    \end{align}
\end{subequations}
where the tilde notation distinguishes the linguistic space from the primitive numerical space, $f_{o,u},f_{o,b}\in \bm{f}$ convert the numerical observation vector into their linguistic representation, and $\psi_u,\psi_b$ are meta-prompts containing natural language task descriptions and agent identity identifiers. The LLM then autoregressively generates a structured response sequence that encapsulates the agent's action and signaling. {\color{black}Rather than open-domain text, these outputs function as task-grounded semantic control messages \cite{kim2025resilientllmempoweredsemanticmac}, whose operational meanings are strictly induced by leader-follower interactions and spatial RBG allocations\footnote{See Appendix \ref{app: semantic} for further clarification.}.} 
Notably, this textual output must be parsed via an inverse mapping to be executed within the environment. 	
\subsection{Generative Policy Architecture with PAG}
\begin{figure}[t]
	\centering
	\includegraphics[width=0.8\linewidth]{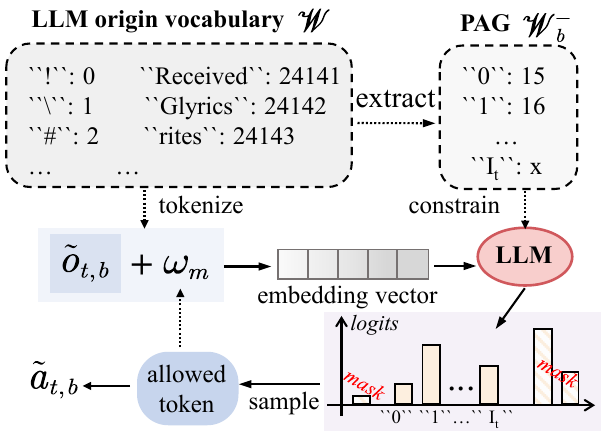}
	\caption{LLM policy action generation with PAG.}
	\label{fig:PAG}
\end{figure}
To ensure the reliability of protocol emergence, we employ PAG to constrain the LLM's generative process within the valid protocol-action subspace. As depicted in Fig. \ref{fig:PAG}, unlike standard stochastic sampling on an open-domain vocabulary $\mathcal{W}$ that may yield unparseable tokens or hallucinations \cite{11303187}, PAG enforces a strict mapping from the hidden states to the admissible token set $\mathcal{W}^-_b,\mathcal{W}^-_u\subset \mathcal{W}$.
Precisely, the BS's output is constrained to only the numerical tokens within $\mathcal{W}^-_b=\{``0", ``1",\cdots, ``I_t "\}$ that indicate the allocation of corresponding RBGs. Here, ``0" indicates no allocation and others specify the allocation to the $i$-th UE. Each DCM action is then represented as a sequence $\Tilde{a}\subscriptbst = \{w_0,\cdots,w_{M}\}$, $w\in\mathcal{W}^-_b$ of fixed length $M$, corresponding to the maximum number of allocable RBs.
 The probability of generating a specific, valid DCM is factorized autoregressively
\begin{equation}
	\label{eq:llm_generation}
	\mathbb{P}(\Tilde{a}\subscriptbst \lvert \Tilde{o}\subscriptbst ) = \prod\nolimits_{j=0}^{\lvert \Tilde{a}\subscriptbst \rvert} \mathbb{P}(w_j \lvert \Tilde{o}\subscriptbst,w_{<j} )
	,
\end{equation}
where $\mathbb{P}(w_j \lvert \Tilde{o}\subscriptbst,w_{<j} )$ represents the probability of generating token $w_j$ given prompt $\Tilde{o}\subscriptbst$ and preceding tokens $w_{<j}$. 
Crucially, to enforce the PAG constraints in practice, we manipulate the conditional probability $\mathbb{P}(w_j | \cdot)$ via logit masking. Let $\mathbf{z}_j \in \mathbb{R}^{|\mathcal{W}|}$ denote the raw logits output by the LLM at step $j$. We define a constraint mask $\mathbf{M} \in \mathbb{R}^{|\mathcal{W}|}$ such that $\mathbf{M}_k = 0$ if token $k \in \mathcal{W}^-_b$ and $-\infty$ otherwise. The valid conditional probability is then derived by
\begin{equation}
    \label{eq: PAG logit masking}
    \mathbb{P}(w_j | \tilde{o}_{t,b}, w_{<j}) = \frac{\exp(\mathbf{z}_{j, w_j} + \mathbf{M}_{w_j})}{\sum_{k \in \mathcal{W}} \exp(\mathbf{z}_{j, k} + \mathbf{M}_k)}.
\end{equation}
This operation forces the probability of invalid tokens to zero, effectively pruning the search space. To achieve computational efficiency, we obtain the log-probabilities from Eq. \eqref{eq:llm_generation} and subsequently normalize these scores using a softmax function. This yields a policy distribution over the action space $\Tilde{\mathcal{A}}_b$ as
\begin{equation}
	\bspolicy(\Tilde{a}\subscriptbst \lvert \Tilde{o}\subscriptbst ) = 
	\frac{\exp\{\sum_{j=0}^{\lvert \Tilde{a}\subscriptbst \rvert} \log\mathbb{P}(w_j \lvert \Tilde{o}\subscriptbst,w_{<j})\}}
	{\sum_{\Tilde{a}\subscriptbst \in \Tilde{\mathcal{A}}_b} \exp\{\sum_{j=0}^{\lvert \Tilde{a}\subscriptbst \rvert} \log\mathbb{P}(w_j \lvert \Tilde{o}\subscriptbst,w_{<j})\}
	}.
\end{equation}
Similarly, we can specify $\mathcal{W}^-_u=\{``0",``1"\}$ to cater to the UE' action $a^{\rm bit}\subscriptuet$ and derive the UE policy $\uepolicy$. Such policy formulation fully exploits the LLM's pre-trained generative priors to score the plausibility of actions, thereby obviating the need for auxiliary, randomly initialized output layers. It is thus robust to diverse action spaces without necessitating architectural modifications.
\subsection{PPO-driven Functional Alignment and Adaptation}
 \begin{algorithm}[tb]
 	\caption{PPO Alignment of LLM-Agents for Protocol Emergence in Dynamic MFSG}
  \label{alg:pscudo-code}
  \renewcommand{\algorithmicrequire}{\textbf{Input:}}
  \renewcommand{\algorithmiccomment}[1]{\hfill $\triangleright$ #1}
  \begin{algorithmic}[1]
       \REQUIRE Discount factor $\gamma$ and $\lambda$, learning rate $\alpha$ and $\beta$, time duration of each episode $T$.
       \STATE  Initialize $(\theta_b,\phi_b)$ and $(\theta_{u},\phi_{u})$.
       \STATE Initialize empty trajectory buffers $\mathcal{B}_b$ for leader and $\mathcal{B}_u$ for followers, respectively.
       \WHILE{epoch $e \leq e^{\rm max} $}
       \STATE Reset the UDTS environment and get initial observations $\tilde{o}\subscriptbst[0]$, $\{\tilde{o}_{i,0,u}\}_{i \in \mathcal{I}_t}$.
            \FOR{each time slot $t \in \{0,1,\cdots,T\}$}
               \STATE // \textit{Coupled experience generation within a time slot}
               \STATE Leader (BS) selects action $\tilde{a}\subscriptbst \sim \pi_b(\cdot|\tilde{o}\subscriptbst;\theta_b)$ that determines the DCM.\COMMENT{\textit{Leader acts first}}
               
               \STATE Each follower (UE $i$) observes local state and the DCM received from BS, selects action $\tilde{a}\subscriptuet \sim \pi_u(\cdot|\tilde{o}\subscriptuet;\theta_u)$, comprising the transmission bitmap and the UCM.\COMMENT{\textit{Followers react}}
               
               \STATE Environment executes physical transmission actions $\{\tilde{\bm{a}}\subscriptuet^{\text{bit}}\}_{i \in \mathcal{I}_t}$.
               \STATE Environment returns the individual rewards $F_{t,b}$, $\{F\subscriptuet\}_{i \in \mathcal{I}_t}$ and next observations. \COMMENT{\textit{$\tilde{o}\subscriptbst[t+1]$ now contains the UCMs $\{\tilde{u}\subscriptit\}_{i \in \mathcal{I}_t}$}}
            \ENDFOR
            \STATE Store the trajectory rollout $\tau_b$ into buffer $\mathcal{B}_b$.
            \STATE Aggregate the trajectory rollout from all UEs $\bigcup_{i \in \mathcal{I}_t}\tau\subscriptiu$, and store into buffer $\mathcal{B}_u$.
            \STATE // \textit{Independent policy updates}
            \FORALL{the leader and the followers}
          	  \IF {$\lvert \mathcal{B} \rvert \geq {\rm Buffer\ Size}$}
	          	  \STATE Minimize its critic's loss as detailed in Eq. \eqref{eq:critic_loss}.
	          	  \STATE Calculate the sampled policy gradient of Eq. \eqref{eq:actor_loss}.
	          	  \STATE Actor update $\theta^{(e+1)} \gets \theta^{(e)} + \alpha \cdot \nabla J(\theta^{(e)})$.
	          	  \STATE Critic update $\phi^{(e+1)} \gets \phi^{(e)} - \beta \cdot \nabla L(\phi^{(e)})$.
	          	  \STATE Clear experience buffer $\mathcal{B}$.
          	  \ENDIF
            \ENDFOR
           
       \ENDWHILE
   \end{algorithmic}
\end{algorithm}

		
		
The training mechanism follows the principles from Section \ref{sec:background on marl}, but we tailor it substantially to accommodate the unique hierarchical structure of our dynamic MFSG. We instantiate LLM-enpowered actor $(\theta_b, \theta_u)$ for BS and UEs, respectively, where dynamic UEs utilize the same neural network by parameter sharing. Their corresponding critic networks are realized by replacing the final layer of the initial LLM decoder block with a single-value head MLP. The training procedure, delineated in Algorithm \ref{alg:pscudo-code}, is orchestrated around the hierarchical experience coupling. Specifically, after an initial environment reset (\texttt{line 5}), the process unfolds within each TTI to precisely capture the game's sequential dependencies. The leader BS first executes its action $a\subscriptbst$ (\texttt{line 7}), which consequently alters the observational space for all followers UEs. The followers then react based on this updated context (\texttt{line 8}), and their joint actions collectively determine the subsequent observations and rewards (\texttt{lines 9-10}). Although the trajectory data collected from this process is inherently coupled by the hierarchical interaction (\texttt{lines 12-13}), the leader and followers independently update their respective policy and value networks (\texttt{lines 16-21}). This process enables the policies to progressively align with the environment dynamics by grounding each update in causally correct, interdependent experiences. 

\subsection{Proof of Convergence}
The co-adaptation of leader and follower policies in our dynamic MFSG creates an inherently non-stationary learning dynamic. This section therefore provides a formal proof that the gradient-based PPO alignment guides the framework to an equilibrium. We begin with fundamental assumptions and definitions.
\begin{definition}[Expected Stackelberg Equilibrium {[}ESE{]}]
	\label{def:ese}
	Let $\Theta_b$ and $\Theta_{u}=\Theta\subscriptiu[1]\times\cdots\times\Theta\subscriptiu[I_t]$ be the policy parameter sets of BS and all UEs, respectively. A policy profile $(\theta_b^\ast,\bm{\theta}_u^\ast)$ is a dynamic follower ESE if the followers’ policy $\bm{\theta}_u^\ast$ constitutes an expected NE $h(\theta_{b}^\ast)$ in response to
	$\theta_b^\ast$, where
	\begin{equation}
	 h(\theta_{b}^\ast)	=\left\{\bm{\theta}_{u}^\ast\in \Theta_{u}|J\subscriptiu(\bm{\theta}_{u}^\ast,\theta_b)\geq J\subscriptiu(\theta\subscriptiu,\theta\subscriptiu[-i]^\ast,\theta_b^\ast),\forall i \right\},
	\end{equation}
	And the leader's policy maximizes its expected utility, anticipating the followers' best response from $ h(\theta_{b}^\ast)$:
	\begin{equation}
		\theta_{b}^\ast \in \arg\max_{\theta_b \in \Theta_b}\inf_{\bm{\theta}_u \in h(\theta_{b})} J_b(\theta_b,\bm{\theta}_u) .
	\end{equation}
\end{definition}
Consistent with standard practice in learning-in-games \cite{Stackel-ActorCritic, Stackel-dynamics}, we only focus on local properties of the equilibrium and thus DSE is adopted.
\begin{definition}[Differential Stackelberg Equilibrium {[}DSE{]} \cite{Stackel-dynamics}]
	\label{def:dse}
	A policy profile $\bm{\varTheta}^\ast=(\theta_b^\ast,\bm{\theta}_u^\ast)$ constitutes a DSE, characterized by the following first- and second-order properties: 
	\begin{itemize}
		\item[a)] For UEs, $\bm{\theta}_u^\ast$ satisfies the differential Nash equilibrium (DNE) condition \cite{DNE}, expressed as $\nabla_{\theta\subscriptiu} J\subscriptiu \left(\theta_b^\ast, \bm{\theta}_u^\ast\right)=0$ and $\nabla_{\theta\subscriptiu}^2 J\subscriptiu\left(\theta_b^\ast, \bm{\theta}_u^\ast\right) \prec 0$;
		\item [b)] For BS, $\theta_b^\ast$ satisfies $\nabla_{\theta_b} \tilde{J}_b\left(\theta_b^\ast,h(\theta_b^\ast)\right)=0$ and $\nabla_{\theta_b}^2 \tilde{J}_b\left(\theta_b^\ast,h(\theta_b^\ast) \right) \prec 0$, where $h(\theta_b^\ast)$ is the UEs' NE response.
	\end{itemize} 
\end{definition}
To analyze the convergence of our PPO-based alignment, we also define the following policy update dynamics \cite{Stackel-dynamics, Stackel-conjec}.
\begin{definition}[Policy Update Dynamics]
\label{def: update_rule}
    Let the joint policy parameters be denoted by the vector $\bm{\varTheta} = (\theta_b,\bm{\theta}_u)^\top$. The learning process constitutes a discrete-time dynamical system governed by the update rule
    \begin{equation}
    \label{eq:stackel update rule}
        \bm{\varTheta}^{(e+1)} = \bm{\varTheta}^{(e)} + \alpha_b\cdot\bm{\Omega}(\bm{\varTheta}^{(e)}),
    \end{equation}
    where $\bm{\Omega}(\bm{\varTheta}^{(e)})=\big(\nabla_{\theta_b} J^{(e)}_b,\iota_u\nabla_{\theta_u} J^{(e)}\subscriptiu[1],\cdots,\iota_u\nabla_{\theta_u} J^{(e)}\subscriptiu[I]\big)^\top$
is the update gradient vector, $\iota_u$ denotes the time scale separation factor and $\alpha_b\iota_u=\alpha_u$. We assume all followers (UEs) adopt an identical time scale separation factor $\iota_u$ for simplicity.
\end{definition}
Furthermore, we make the following assumptions.
\begin{assumption}
	\label{assum: soomth and compact}
    The policy parameter spaces $\Theta_b$ and $\Theta_u$ are non-empty compact sets. And the agents' utility functions, $J_b$ and $J\subscriptiu$ are sufficiently smooth, i,e, $J_b,J\subscriptiu\in C^2({\bm \Theta},\mathbb{R})$. 
\end{assumption}
\begin{assumption}
	\label{assum: bounded utility}
    The system-wide utility function is bounded, i.e., $F_{t,b} \in [0, F_{\max}]$ for all states.
\end{assumption}
\begin{assumption}
	\label{assum: weak coupling}
    {\color{black}
    Near the equilibrium $\bm \varTheta^\ast$, the MFSG operates in a weak coupling regime where the second-order inter-player effects are locally dominated by the corresponding self-curvature terms\footnote{ \color{black}This weak-coupling condition is imposed strictly in a local neighborhood of the equilibrium $\bm \varTheta^\ast$ rather than arbitrary overloaded contention regimes; it characterizes a coordinated operational regime where the BS's intents and UEs' best responses have effectively mitigated persistent RBG overlap, thereby structurally decoupling the agents' utilities.}. 
    That is, for any distinct agents $i,j,k\in \{\UEn[1],\cdots,\UEn[I_t]\}$, 
    \begin{equation}
    \Vert \nabla^2_{\theta_i}J_k(\bm \varTheta^\ast)\Vert \gg \Vert \nabla_{\theta_i,\theta_j}J_k(\bm \varTheta^\ast)\Vert 
    \end{equation}
    }
\end{assumption}
\begin{assumption}
	\label{assum: dirichlet drift}
    We assume that any environmental state distribution $\mathcal{D}$ is independently sampled from a Dirichlet distribution $\text{Dir}(\chi_0 , \chi_1,\cdots, \chi_{|\mathcal{S}|})$, where $\chi_1,\chi_2,\cdots,\chi_{|\mathcal{S}|}$ quantifies the environmental stability \cite{amorosa2025learningdecentralizedmediumaccess, ng2011dirichlet}.
\end{assumption}

To ensure that the protocol emergence process has a well-defined and reachable target, we begin by formally defining and proving the existence of such an equilibrium within our system. This result is formalized as Theorem \ref{thm:existence}.
\begin{theorem}[Existence Theorem]
	\label{thm:existence}
	In our LLM-driven MFSG system -- where BS acts as the leader and a dynamic number of UEs act as followers -- an ESE policy profile $(\theta^\ast_b,\bm{\theta}^\ast_u)$ is guaranteed to exist. 
\end{theorem}
As completely proven in Appendix \ref{app: existence}, we analyze the equilibrium hierarchically. First, we show that the follower subgame is a potential game \cite{9207866}. Hence, for any $I_t$, it admits an NE, i.e., the followers' best-response correspondence is nonempty. Next, in the dynamic-follower setting, we invoke Weierstrass' Theorem \cite{Weierstrass} to establish the existence of a solution to the leader's problem. Together, these results imply the existence of an ESE.

Based on the existence of equilibrium, we present a general convergence theorem, which states that the policy gradient update converges in a fixed-number MFSG. 
\begin{theorem}[Converge Theorem]
	\label{thm:converge}
    For the policy update dynamics described in the Definition \ref{def: update_rule}, and under Assumptions \ref{assum: soomth and compact} and \ref{assum: weak coupling}, there exist positive constants $\kappa_1$, $\kappa_2$ such that if the leader's learning rate is set to $\alpha_b = \frac{\kappa_1}{\kappa_2}$, the sequence of policy parameters $\{\bm{\varTheta}^{(e)}\}_{e\geq0}$ is guaranteed to converge locally to a DSE at a rate of $O\big((1-\frac{\kappa_1^2}{\kappa_2})^\frac{e}{2}\big)$.
\end{theorem}
We leave the proof in Appendix \ref{app: converge} while providing a proof sketch here. The logic is to show that the update map, $\Lambda(\bm{\varTheta})=\bm{\varTheta}+\alpha_b \Omega(\bm{\varTheta})$, is a local contraction around the DSE fixed point $\bm{\varTheta}^*$. The analysis hinges on the property, derived in the appendix from Assumptions \ref{assum: soomth and compact}-\ref{assum: weak coupling}, that the symmetric part of the gradient field's Jacobian, $M\left(\bm{\varTheta}^*\right)=\frac{1}{2}\left(\mathbf{J}_{\bm{\Omega}}^{\top}+\mathbf{J}_{\bm{\Omega}}\right)$, is negative definite ($M\left(\bm{\varTheta}^*\right) \prec 0$). This allows for the selection of a learning rate  $\alpha_b=\kappa_1 / \kappa_2$, where $\kappa_1=\sigma_{\min}\left(-M\left(\bm{\varTheta}^*\right)\right)$, $\kappa_2=\sigma_{\max }\left(\mathbf{J}_{\bm{\Omega}}^{\top} \mathbf{J}_{\bm{\Omega}}\right)$ and $\sigma$ denotes a singular value, which guarantees contraction by satisfying
\begin{equation}
    \left\|I+\alpha_b J_{\bm{\Omega}}\left(\bm{\varTheta}^*\right)\right\|_2^2 \leq 1-\kappa_1^2 / \kappa_2<1.
\end{equation}
This result confirms that the spectral radius $\rho\left(\nabla \Lambda\left(\bm{\varTheta}^*\right)\right)<1$, proving that $\Lambda$ is a contraction and implying that {\color{black}under local smoothness and boundedness assumptions, though it does not ensure a global optimum for arbitrary LLM policies, it yields local geometric convergence to the DSE at the claimed rate. }

In our dynamic MFSG setting, where the number of followers changes stochastically and all followers share a common LLM policy, we derive the following corollary.
\begin{corollary}
	\label{corol}
 Suppose all agents employ LLM-empowered policies for dynamic MFSG. Let the leader's policy be denoted by $\theta_b$, and let all followers share a common policy $\theta_u$. Regardless of the changes in the number of followers, Algorithm \ref{alg:pscudo-code} maintains the almost sure convergence properties established in Theorem \ref{thm:converge}.
\end{corollary}
We leave the proof in Appendix \ref{app: corollary 1}, which extends Theorem \ref{thm:converge} by analyzing the averaged learning dynamics over the distribution of followers. The simulation results in Fig. \ref{fig: convergence} further demonstrate that convergence is not only theoretically possible but also achievable in practice.

To quantify the uncertainty induced by network anomalies, we further investigate the robustness of the converged equilibrium under environmental shifts.

\begin{theorem}[Robustness of Equilibrium]
	\label{thm:robustness}
    
    Given the equilibrium $\boldsymbol{\theta}^*$ optimized under distribution $\mathcal{D}$, and let $\hat{J}_{\mathcal{D}'}^{(K)}(\boldsymbol{\theta}^*)$ denote the empirical utility over $K$ samples from the perturbed $\mathcal{D}'$. Under Assumptions 2-4, for any $\delta > 0$, the difference between the empirical and nominal utility $J_{\mathcal{D}}(\boldsymbol{\theta}^*)$ satisfies\footnote{We here adopt the leader's utility for robustness analysis as it explicitly encapsulates the system-wide performance. Here, $J_{\mathcal{D}}(\boldsymbol{\theta}^*) \triangleq \mathbb{E}_{\boldsymbol{s} \sim \mathcal{D}}[F_{t,b}(\boldsymbol{\theta}^*, \boldsymbol{s})]$, where the state $\boldsymbol{s}$ abstracts a joint state trajectory over an episode. In this context, network anomalies are rigorously modeled as stochastic perturbations to the trajectory distribution $\mathcal{D}$.}:
    \begin{equation}
        \mathbb{P}\left( \left| J_{\mathcal{D}} - \hat{J}_{\mathcal{D}'}^{(K)} \right| \ge \delta \right) \le \frac{F_{\max}^2}{2\delta^2 (1 + |\mathcal{S}|\chi_{\epsilon})} + 2 \exp \left( -\frac{2K \delta^2}{F_{\max}^2} \right),
    \end{equation}
    where $\chi_{\epsilon}=\min\{\chi_i\}_i^{|\mathcal{S}|}>0$ is the Dirichlet distribution hyperparameter, and $|\mathcal{S}|$ denotes the cardinality of the trajectory space.
\end{theorem}
Theorem \ref{thm:robustness} ensures the learned equilibrium maintains stability rather than suffering catastrophic failure under mild environment shifts. Moreover, a larger $K$, i.e., a longer interaction horizon during practical testing can reduce the empirical bias. We leave the proof in Appendix \ref{app: robustness proof}.

\subsection{Complexity Analysis}
We analyze the framework's efficiency by examining its computational complexity, space complexity, and the effective exploration complexity. Regarding computational time, the floating point operations (FLOPs) for a single forward pass of an LLM with $n$ layers and hidden dimension $\mathcal{d}_{\rm model}$ scales quadratically with sequence length $\mathcal{l}$ as $\mathcal{O}(\mathcal{n}\mathcal{l}^2\mathcal{d}_{\rm model})$. 
This scaling is intrinsically asymmetric. For the leader, the prompt length is a function of $I_t$, i.e.,
\begin{equation}
	\mathcal{l}\subscriptbst \approx \mathcal{l}_{\psi_b}+ \sum_{i=1}^{I_t} \mathcal{l}\subscriptiu \triangleq \mathcal{O}(I_t),
\end{equation}
where $\mathcal{l}_{\psi_b}$ is the constant length of the meta-prompt and $\mathcal{l}\subscriptiu$ is the length of the tokenized UCM from UE followers. And the UE sequence length is constant $\mathcal{l}\subscriptiu = \mathcal{O}(1)$. Consequently, generating an action sequence of $M$ incurs a cost of $\mathcal{O}(M \cdot I_t^2 \cdot \mathcal{n}\mathcal{d}_{\rm model})$ for the leader and $\mathcal{O}(M\cdot I_t\cdot \mathcal{n}\mathcal{d}_{\rm model})$ for all $I_t$ followers. With $I_t$ varying over time according to a distribution, the total training complexity over $e^{\rm max}$ epochs is best characterized by its expected value
\begin{equation}
	\mathcal{O}\Big(e^{\rm max} T\cdot M\mathcal{n}\mathcal{d}_{\rm model}\big(\mathbb{E}[I_t^2]+\mathbb{E}[I_t]\big)\Big).
\end{equation}
In contrast, the per-decision inference cost is an instantaneous measure dependent on the current $I_t$, given by the sum of the leader's and followers' forward generation costs $\mathcal{O}\big( M I_t^2\mathcal{n}\mathcal{d}_{\rm model}+I_tM\mathcal{n}\mathcal{d}_{\rm model}\big)$. 

The exploration efficiency is fundamentally dictated by the action space complexity. Standard decoding operates on a high-dimensional probability simplex (e.g., $|\mathcal{W}| \sim 10^5$ for Llama series \cite{grattafiori2024llama3herdmodels}), suffering from the curse of dimensionality. PAG functions as a dimension reduction operator, projecting this sparse optimization problem onto a compact, task-valid subspace where $|\mathcal{W}^-| \ll |\mathcal{W}|$. This $\mathcal{O}(10^4)$-fold dimensionality reduction regularizes the policy search landscape, transforming intractable open-ended generation into a tractable discrete control task.

The total space complexity during training is the sum of storage for all network parameters and experience replay buffers, totaling $\mathcal{O}\big(\mathcal{n}\mathcal{d}^2_{\rm model}+|\mathcal{B}_b|+|\mathcal{B}_u|\big)$. During inference, this requirement reduces to storing only the actor network parameters, yielding a space complexity of $\mathcal{O}\big(\mathcal{n}\mathcal{d}^2_{\rm model}\big)$.

\section{Simulation Results}
\begin{figure*}[tb]
	\centering
	\includegraphics[width=0.9\linewidth]{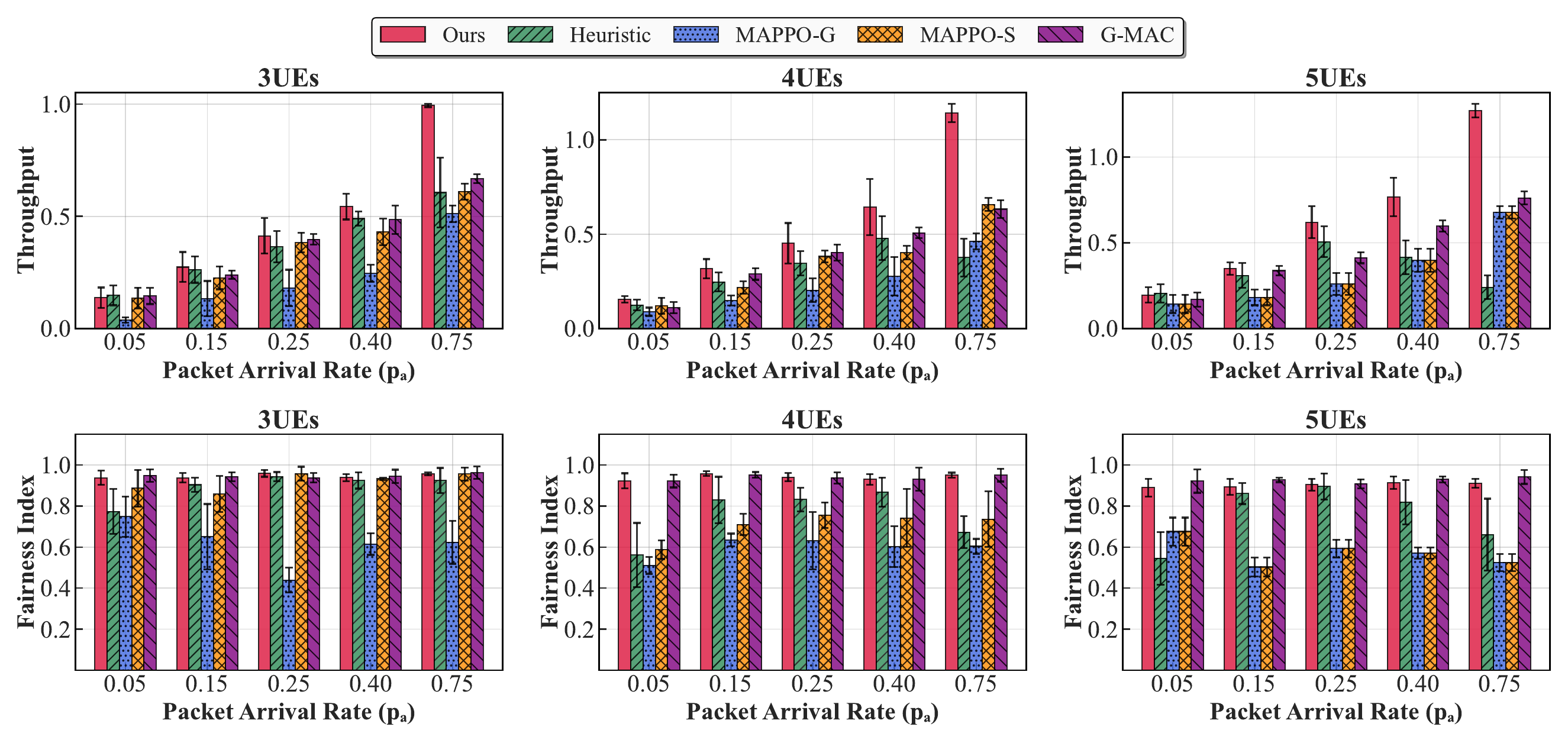}
	\caption{Performance comparison of throughput and fairness index under various packet arrival rates and network sizes, where the packet arrival rate $p_a$ of all UEs is set to the same value for each case. Results are averaged over three independent runs.}
	\label{fig: comparisons}
\end{figure*}
In this section, we first detail the simulation setup and the baseline methods. We then compare the performance of our proposed framework against the baselines under various scenarios. Additionally, we include ablation studies and hyperparameter sensitivity analysis.
\subsection{Simulation Setup}
We consider a UDTS environment where the number of UEs $I_t$ varies within the set $\{3, 4, 5\}$. Each episode spans $T=24$ TTIs, with a TTI duration of $5\times10^{-3}$ s. UEs commence with an empty and sufficiently large buffer to prevent overflow. By default, three distinct $p_a\in\{0.1,0.3,0.7\}$ are assigned to UEs cyclically to model heterogeneous traffic loads, with a fixed dPDU size $\xi^{\rm gen}=256$ bits. Each UE's channel transitions between ``poor", ``medium", and ``good" states with a changing frequency in $[1,10]$ ms, where each state maps to a spectral efficiency $\nu\subscriptit$ via a pre-defined MCS lookup table. Moreover, we set $N^{\rm rb}_{\rm sc}=12$, $N^{\rm sh}_{\rm symbol}=14$, aligning with the commonly deployed 3GPP new radio (NR) configuration \cite{3GPP38321,3GPP38214}.

We employ open-source Llama3-1B \cite{grattafiori2024llama3herdmodels} as the actor-critic backbone. The critic network is augmented with a three-layer value head ($1024$ hidden neurons per layer and ReLU activations) to produce scalar state values. The actor's temperature linearly decays from $3.0$ to $0.3$ during training. To facilitate online training, we perform $5$ epochs of PPO updates after collecting $10$ full episodes, after which the buffer $\mathcal{B}$ is cleared for new collections. The PPO updates utilize a mini-batch size of $128$, an Adam optimizer with a learning rate of $5\times10^{-4}$, and a cosine learning rate decay schedule. We totally train for $2,000$ epochs, periodically evaluating the performance in fixed-UE settings. Experiments are conducted on a server running Ubuntu 22.04.5, equipped with an AMD EPYC 7763 64-Core CPU, 1 TiB of RAM, and eight NVIDIA A800 GPUs. The software stack includes Python 3.12, CUDA 12.2, and PyTorch 2.5.
Furthermore, we consider the following baseline methods:
\begin{itemize}
	\item \textbf{Heuristic}: An enhanced S-ALOHA \cite{kim2025resilientllmempoweredsemanticmac} where each UE initially transmits a dPDU with a probability $0.5$, which is then adaptively decreased upon a collision.
	\item  \textbf{MAPPO-S}: Symbolic MAPPO algorithm for protocol emergence, as presented in \cite{mostafa2025}, which is retrained for each specific scenario to adapt to environmental shifts.
	\item  \textbf{MAPPO-G}: The same MAPPO architecture, but trained in one scenario and directly applied to others without retraining to evaluate its generalization capabilities.
    \item \textbf{G-MAC}: A state-of-the-art baseline inspired by \cite{GMAC}, which achieves structural generalization across varying numbers of devices by utilizing temporal policy serialization.
\end{itemize}
See Appendix \ref{app: details setup} for more detailed simulation setup and the baselines' implementation details. 
Finally, we introduce the following network KPIs to evaluate the performance.
\begin{itemize}
	\item[1)] \textit{Throughput}: We define the average throughput of our system as the number of dPDU bits received by BS per active UE per unit of time, computed over all TTIs by
	\begin{equation}
		\label{eq:TH}
		{\rm Th} = \frac{1}{T}\sum_{t=1}^{T}\Big(\frac{1}{|\mathcal{I}_t|}\sum_{i\in \mathcal{I}_t}\xi^{\rm rec}\subscriptuet\Big).
	\end{equation}
	\item[2)] \textit{Fairness Index}: We utilize the Jain's Fairness Index \cite{jfi} as detailed in the second term of Eq. \eqref{eq: long-term bs unity}.
	\item[3)] \textit{Resource Efficiency}: The resource efficiency is defined as the received dPDUs divided by the total RBG utilization
	\begin{equation}
		\label{eq:Eff}
		{\rm Eff} = \frac{\sum_{t=1}^{T}\sum_{i\in \mathcal{I}_t}\xi^{\rm rec}\subscriptuet}{\sum_{i\in \mathcal{I}_T}x_{i,T,u}}.
	\end{equation}
    \item[4)] \textit{Packets Success Rate}: We measure the ratio of successfully delivered dPDUs by $(\sum \xi^{\rm rec}) / (\sum \xi^{\rm tx})$.
    \item[5)] \textit{Latency}: We consider the average end-to-end delay (in slots) from dPDU generation to successful reception, as well as the 95\% tail latency. 
    \item[6)] \textit{Queue Stability}: This is evaluated via the average and maximum buffer occupancy across all UEs, serving as an indicator of congestion control.
\end{itemize}

\subsection{Numerical Results and Analysis}
\begin{table*}[ht]
\centering
  \caption{Comprehensive performance analysis of the proposed framework under heterogeneous traffic scenarios with $5$ UEs, including Bernoulli, periodic, and bursty flows. Results are averaged over three independent tests.}
  \label{tab:mixed_traffic_results}
  \renewcommand{\arraystretch}{1.2} 
  \setlength{\tabcolsep}{6pt}
  
  \begin{threeparttable}
    \begin{tabular}{lcccccc}
    \toprule
    \multirow{2}{*}{\textbf{Traffic Scenario}} & \multicolumn{2}{c}{\textbf{System Performance}} & \multicolumn{2}{c}{\textbf{Latency (Slots)}} & \multicolumn{2}{c}{\textbf{Queue Stability (dPDUs)}} \\
    \cmidrule(lr){2-3} \cmidrule(lr){4-5} \cmidrule(lr){6-7}
     & \textbf{Th.} (Mbps) & \textbf{Success Rate (\%)} & \textbf{Avg.} & \textbf{95\% Tail} & \textbf{Avg. Length} & \textbf{Max. Length} \\
    \midrule
    
    \textbf{Case 1 (5B)} & $1.140 \pm 0.037$ & $93.70 \pm 2.25$ & $2.95 \pm 0.08$ & $6.00 \pm 1.21$ & $2.01 \pm 0.13$ & $6.81 \pm 1.49$ \\
    \midrule
    \textbf{Case 2 (3B/2P)} & $1.028 \pm 0.026$ & $92.42 \pm 1.89$ & $2.87 \pm 0.12$ & $5.93 \pm 0.09$ & $1.63 \pm 0.28$ & $6.49 \pm 0.28$ \\
    
    \textbf{Case 3 (2B/3P)} & $0.935 \pm 0.035$ & $93.15 \pm 2.33$ & $2.56 \pm 0.50$ & $7.58 \pm 4.38$ & $1.52 \pm 0.41$ & $7.10 \pm 2.97$ \\
    \midrule
    \textbf{Case 4 (3B/2O)}  & $1.179 \pm 0.063$ & $89.29 \pm 1.34$ & $4.17 \pm 0.08$ & $10.03 \pm 1.06$ & $2.48 \pm 0.32$ & $8.13 \pm 1.52$ \\
    
    \textbf{Case 5 (2B/3O)} & $1.226 \pm 0.015$ & $89.46 \pm 1.48$ & $4.55 \pm 0.11$ & $9.62 \pm 0.87$ & $2.87 \pm 0.27$ & $8.85 \pm 0.88$ \\
    \midrule
    \textbf{Case 6 (3B/1P/1O)} & $1.109 \pm 0.045$ & $90.79 \pm 0.58$ & $3.39 \pm 0.13$ & $6.82 \pm 0.86$ & $2.00 \pm 0.42$ & $6.24 \pm 1.31$ \\
    
    \textbf{Case 7 (1B/2P/2O)} & $1.027 \pm 0.007$ & $89.86 \pm 2.12$ & $4.04 \pm 0.34$ & $9.67 \pm 2.36$ & $2.34 \pm 0.31$ & $8.42 \pm 1.77$ \\
    \bottomrule
  \end{tabular}
    \begin{tablenotes}
      \footnotesize
      \item[-] \textit{Metric}: \textbf{Th.}: Average throughput; \textbf{Success Rate}: Percentage of successfully received dPDUs relative to the transmitted; \textbf{95\% Tail}: The 95th percentile of queuing delay; \textbf{Queue Stability}: The average and maximum buffer occupancy per UE.
     \item[-] \textit{Traffic pattern abbreviations}: \textbf{B}: Bernoulli traffic; \textbf{P}: Periodic traffic; \textbf{O}: Bursty traffic (ON-OFF).
    \end{tablenotes}
  \end{threeparttable}
\end{table*}
\begin{figure}[t]
	\centering
	\includegraphics[width=0.9\linewidth]{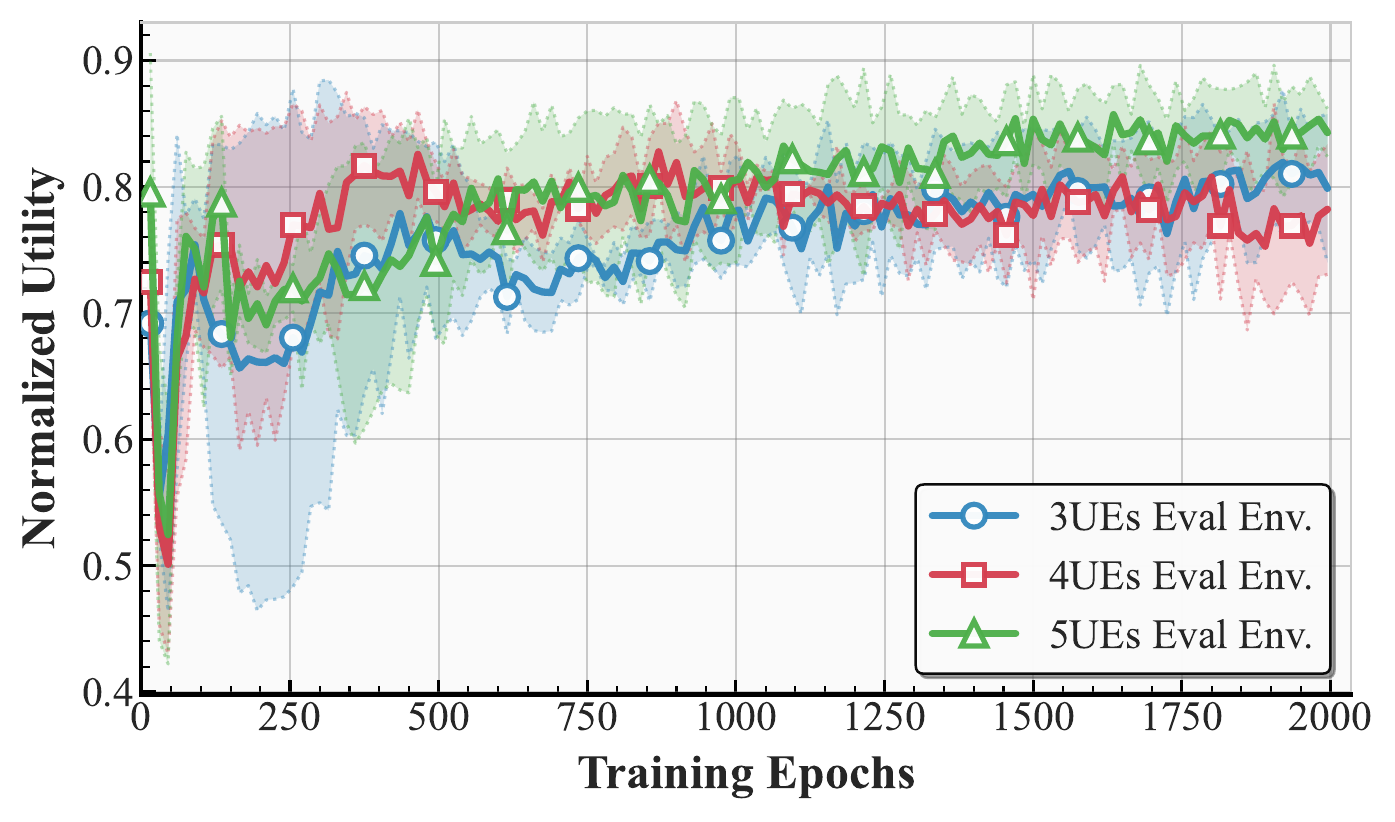}
	\caption{Convergence of normalized system utility during training in evaluation environments with varying UE numbers.}
	\label{fig: convergence}
\end{figure}
\begin{figure}[t]
	\centering
	\includegraphics[width=.9\linewidth]{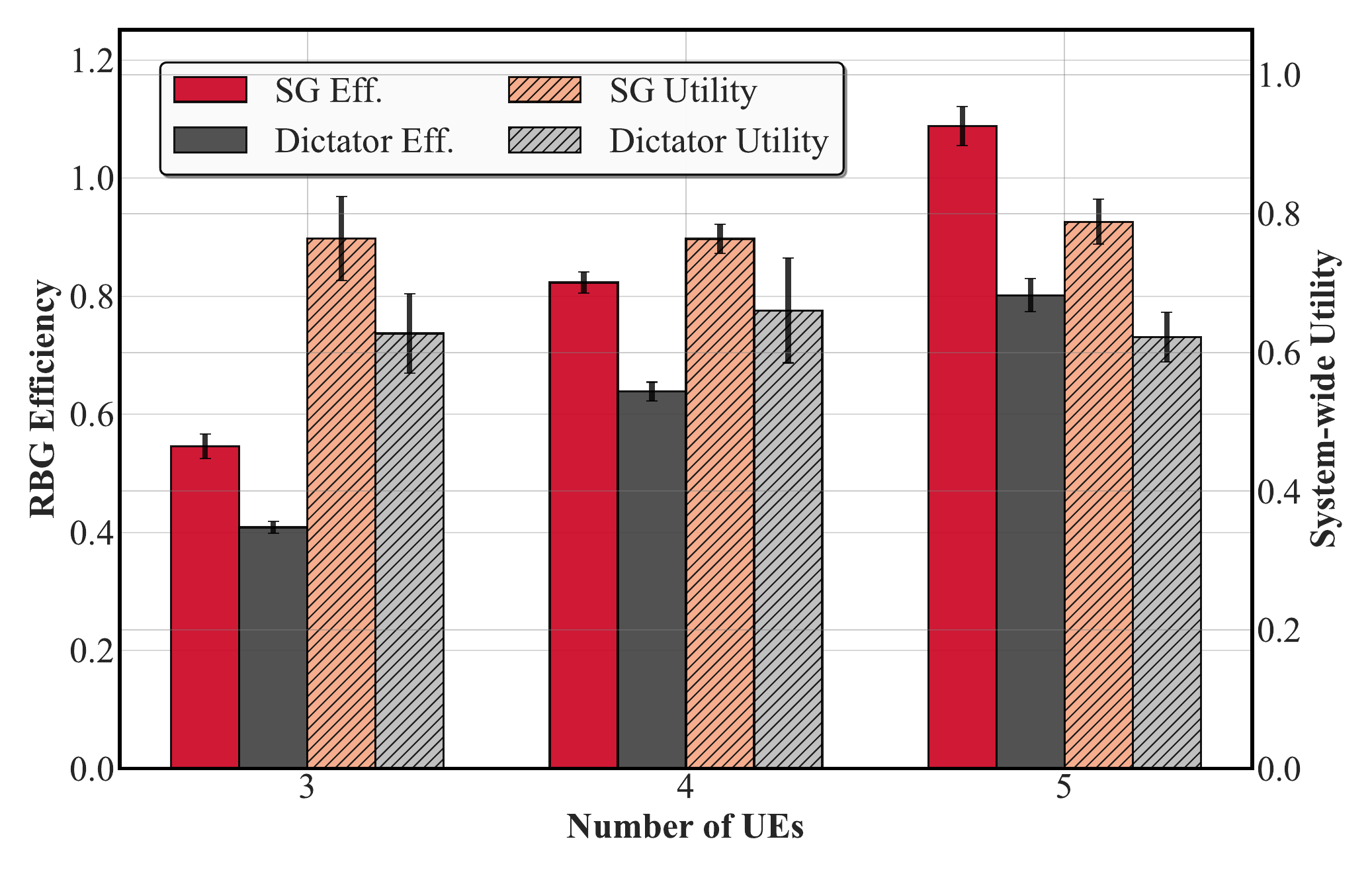}
	\caption{Ablation study on the impact of the hierarchical SG mechanism. Results are averaged over three independent runs.}
	\label{fig: SGcomparison}
\end{figure}
\begin{figure*}[tb]
	\centering
	\includegraphics[width=0.85\linewidth]{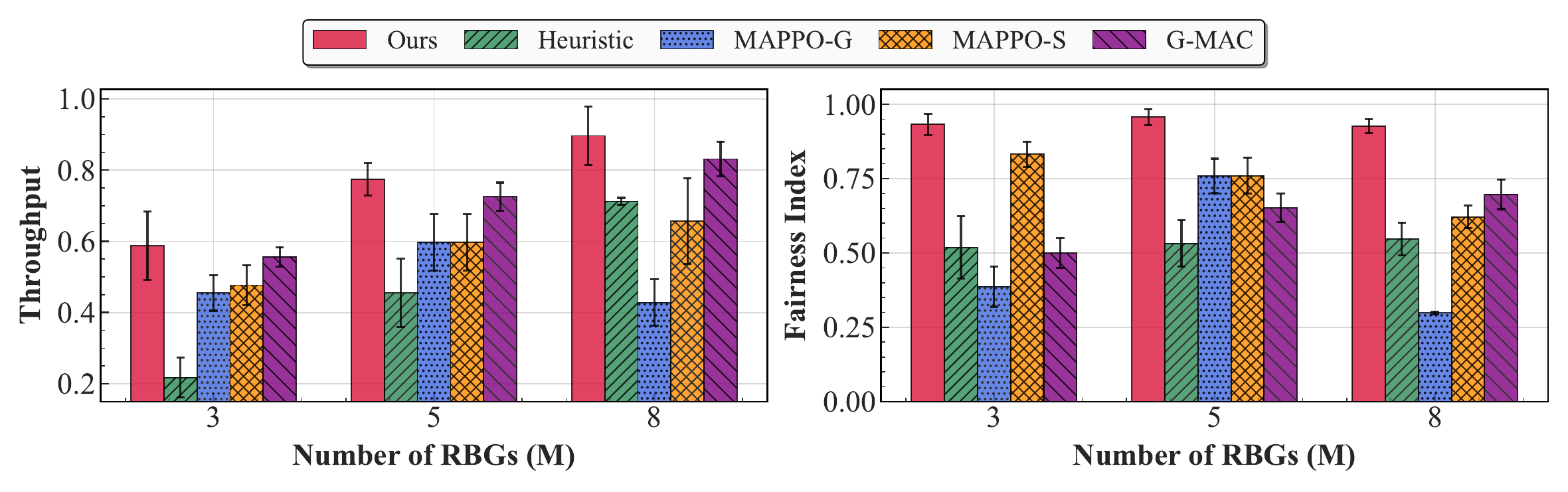}
	\caption{Performance comparison of throughput and fairness index with a varying number of RBGs, evaluated with $5$ UEs and averaged over three independent runs.}
	\label{fig:diffM}
\end{figure*}
\begin{figure}[t]
	\centering
	\includegraphics[width=0.9\linewidth]{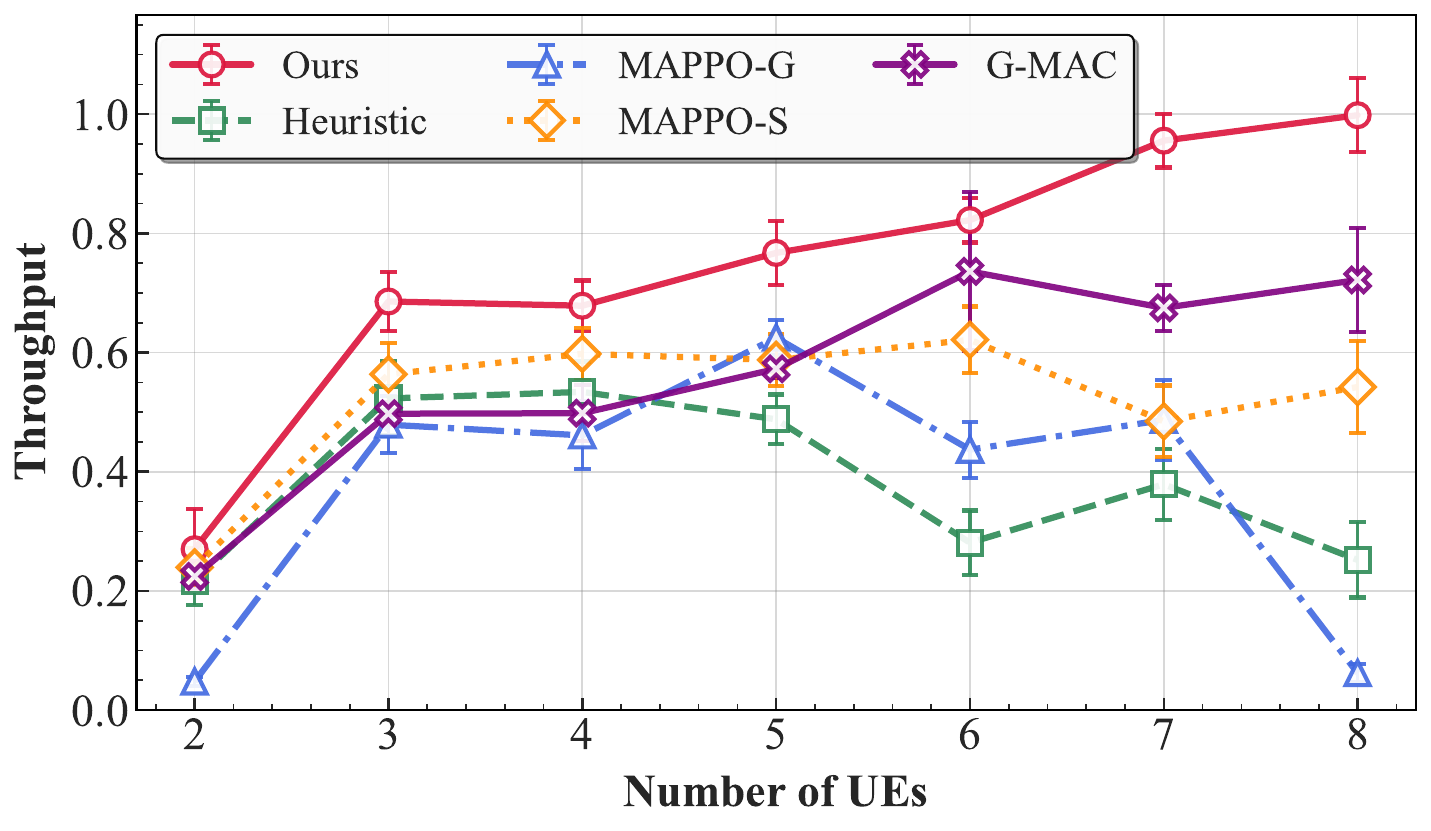}
	\caption{Performance scalability with an increasing number of UEs. Results are averaged over three independent runs.}
	\label{fig:uescale}
\end{figure}
\subsubsection{Comparative Analysis with Baselines}
Fig. \ref{fig: comparisons} presents a comparative evaluation of the proposed framework against baselines in terms of average throughput and fairness. Both the \texttt{MAPPO-G} and \texttt{G-MAC} are trained exclusively in the 5-UEs scenario. As observed, our method consistently dominates all baselines across varying configurations. In the high-contention regime with $5$ UEs and $p_a=0.75$, the proposed framework achieves a throughput of approximately $1.2$, surpassing the retrained \texttt{MAPPO-S} and \texttt{Heuristic} by $46.6\%$ and $81.1\%$, respectively. Notably, while the \texttt{G-MAC} effectively mitigates the catastrophic degradation seen in \texttt{MAPPO-G} by adopting a novel BS policy structure \cite{GMAC}, our method still outperforms it by approximately $38.5\%$, verifying a superior generalization of the LLM-empowered framework. In terms of fairness, as shown in the second row of Fig. \ref{fig: comparisons}, our approach maintains a near-optimal fairness index across all scenarios. In stark contrast, both the \texttt{Heuristic} and \texttt{MAPPO-G} exhibit sharp deterioration in the 5 UEs scenario, with fairness falling below $0.6$. Accordingly, the consistently high performance across different network scales demonstrates our framework's strong generalization capability, operating on dynamic network topologies without the need for retraining.

Fig. \ref{fig: convergence} illustrates the convergence of the system-wide utility during the PPO training. The utility steadily improves and converges to approximately $0.8$ across all evaluated scenarios after approximately $1,250$ training epochs, empirically validating the convergence guarantee established in Corollary \ref{corol}.
\subsubsection{Contribution of SG Mechanism}
To validate the superiority of our proposed MFSG mechanism, we conduct an ablation study comparing its performance against a centralized ``Dictator'' scheme. In this baseline, UEs (followers) forfeit their autonomy, rigidly adhering to the BS's (leader's) scheduling commands without local optimization. As depicted in Fig. \ref{fig: SGcomparison}, the SG framework yields average gains of approximately $24.74\%$ in RBG efficiency and $17.58\%$ in system utility. This performance gap confirms UEs are not merely passive executors. Unlike the rigid Dictator baseline, our framework empowers UEs to leverage their private local information for fine-grained decision-making, thereby handling heterogeneous network complexities more effectively.
\subsubsection{Impact of Heterogeneous Traffic}
To validate robustness under realistic conditions, we extend the traffic model beyond the standard Bernoulli process\footnote{For detailed experimental settings of these traffic models, please refer to Appendix \ref{app: details setup}.}. We introduce periodic traffic to represent URLLC tasks characterized by fixed arrival intervals, and bursty traffic to model eMBB tasks like web browsing or variable bit rate (VBR) video, formulated as an on-off process with high arrival rates during active periods \cite{embb}. Table \ref{tab:mixed_traffic_results} details the system performance under these traffic compositions with $5$ UEs. In the baseline homogeneous scenario (\textit{Case 1}), the system exhibits a low average latency of $2.95$ slots and consistent queue lengths. When deterministic periodic traffic is introduced (e.g., \textit{Case 3}), the average latency and queue occupancy further improve by approximately $2.7\%$ and $18.9\%$, respectively. This enhancement suggests that the framework may be implicitly adapting to the regularity of periodic flows, thereby mitigating contention for the remaining users. The inclusion of bursty traffic (e.g., \textit{Case 5}) imposes significant stress on the network, evidenced by the rise in the 95\% tail latency to $10.03$ slots and increased maximum queue lengths. Even in \textit{Case 6}, the fully heterogeneous setting, the protocol achieves a balanced performance with a moderate latency increase of $16.9\%$ and queue length increase of $13.6\%$ compared to the baseline. These results corroborate that the framework maintains system stability without performance collapse, demonstrating robust adaptability to diverse and complex traffic dynamics.
\subsubsection{Impact of Different RBGs}
Fig. 8 evaluates the framework’s scalability by varying the number of RBGs ($M$) from 3 to 8. The throughput analysis (left) confirms that our method scales most effectively with increased resources, peaking at approximately 0.90 when $M=8$. This performance significantly outperforms the strongest baseline, G-MAC (0.83), and surpasses the Heuristic and MAPPO-S methods by wide margins. Notably, MAPPO-G exhibits a generalization failure in high-resource settings, with throughput dropping sharply to 0.43 at $M=8$. Regarding fairness (right), our approach demonstrates superior robustness, consistently maintaining an index above 0.92 across all scenarios. In contrast, while G-MAC offers competitive throughput, its fairness remains comparatively low (peaking at $\approx 0.70$); similarly, MAPPO-S suffers a significant fairness decline as $M$ increases. Overall, our method achieves the optimal trade-off between aggregate throughput and equitable resource distribution.

Fig. \ref{fig:diffM} evaluates the framework's performance and adaptability when varying the number of RBGs while maintaining other configurations unchanged. The left chart indicates that our method scales most effectively with increased resources, yielding a substantial performance gain of $52.5\%$ as $M$ expands from $3$ to $8$. With $M=8$ RBGs, our method achieves the highest throughput, outperforming \texttt{G-MAC} and \texttt{MAPPO-S} by around $7.9\%$ and $36.4\%$, respectively. Conversely, \texttt{MAPPO-G} exhibits a severe generalization failure with the performance degrading paradoxically as resources increase. The fairness results on the right further highlight the robustness of our approach, which maintains a high fairness index regardless of the number of RBGs, ensuring equitable resource distribution. While \texttt{G-MAC} provides competitive throughput, its fairness fluctuates significantly, lagging behind our method by roughly $24.7\%$ at $M=8$.
\subsubsection{Impact of Number of UEs}
\begin{table}[tb]
	\centering
	\caption{Scalability analysis in a hyper-cellular hierarchical multi-cell environment. The results of throughput in Mbps (Mean $\pm$ Std) are averaged over three independent experiments.}
	\label{tab:multi_cell_scalability}
	\begin{tabular}{cccc}
		\toprule
		\multirowcell{2}{\textbf{Num. of Cells}\\(\textit{5 UEs per Cell})} & \multicolumn{3}{c}{\textbf{Tier-0 Resource Partitioning Strategy}} \\
		\cmidrule(lr){2-4}
		& \textbf{Equal} & \textbf{Proportional} & \textbf{Random} \\
		\midrule
		2 Cells & $1.32 \pm 0.058$ & $1.36 \pm 0.051$ & $1.25 \pm 0.138$ \\
		3 Cells & $1.95 \pm 0.132$ & $1.91 \pm 0.127$ & $1.82 \pm 0.234$ \\
		4 Cells & $2.42 \pm 0.910$ & $2.13 \pm 0.189$ & $2.57 \pm 0.121$ \\
		\bottomrule
	\end{tabular}
\end{table}
Fig. \ref{fig:uescale} evaluates the throughput scalability of the proposed framework against baselines as the network density increases from $2$ to $8$ UEs. Our method demonstrates superior robustness in all scenarios. Specifically, at the peak load of $8$ UEs, the proposed algorithm achieves substantial throughput gains of $38.31\%$ and $84.12\%$ over the strongest baselines, \texttt{G-MAC} and \texttt{MAPPO-S}, respectively. In contrast, \texttt{Heuristic} and \texttt{MAPPO-G} suffer severe degradation in dense networks, plummeting to throughput levels that are $74.76\%$ and $93.79\%$ lower than the proposed scheme, respectively. These results validate that the LLM-empowered protocol optimizes channel utilization more effectively than standard MARL and heuristic schemes under dynamic, large-scale conditions.

In Table \ref{tab:multi_cell_scalability}, we extend the evaluation to a multi-cell setting by adopting the hyper-cellular network (HCN) architecture \cite{zhou2016software}, which consists of one controlling BS (CBS) providing continuous macro coverage and $N$ data BSs (DBSs) deployed for localized capacity. Each DBS cell serves four local UEs and one shared edge UE with heterogeneous traffic intensities $p_{a}\in\{0.1,0.2,0.4,0.6,0.8\}$. This topology enforces active inter-cell coupling, as shared UEs create dynamic load dependencies between adjacent DBSs, necessitating coordinated resource partitioning by the CBS. The total resource pool scales linearly with the network size, maintaining an average of $M=5$ per cell. At the network level, the CBS partitions the global RBGs among the DBSs using three policies -- Equal (static), Proportional (demand-aware), and Random -- while each DBS acts as the local leader, performing inference and executing the Stackelberg game following the same procedure as in the single-cell experiments. By embedding our framework into the HCN architecture, the system demonstrates robust scalability with near-linear throughput growth, proving that the locally learned MAC policy transfers effectively to larger, CBS-coordinated multi-cell deployments. More related discussion is provided in Appendix \ref{app: Scalability}.

\subsubsection{Impact of Policy Backbones}
\label{sec: Policy Backbones}
\begin{figure*}[ht]
    \centering
    \subfloat{
		\includegraphics[width=0.63\linewidth]{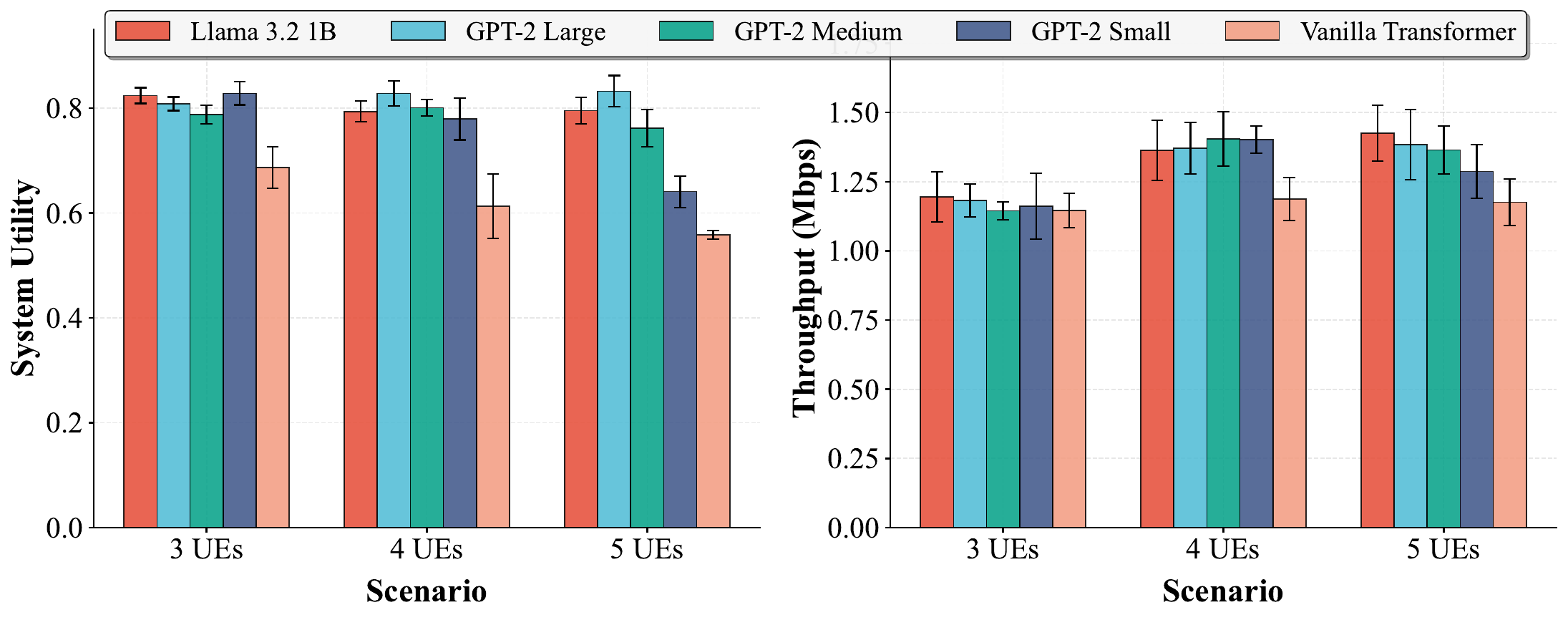}}
        \hspace{-0.3cm}
    \subfloat{
		\includegraphics[width=0.33\linewidth]{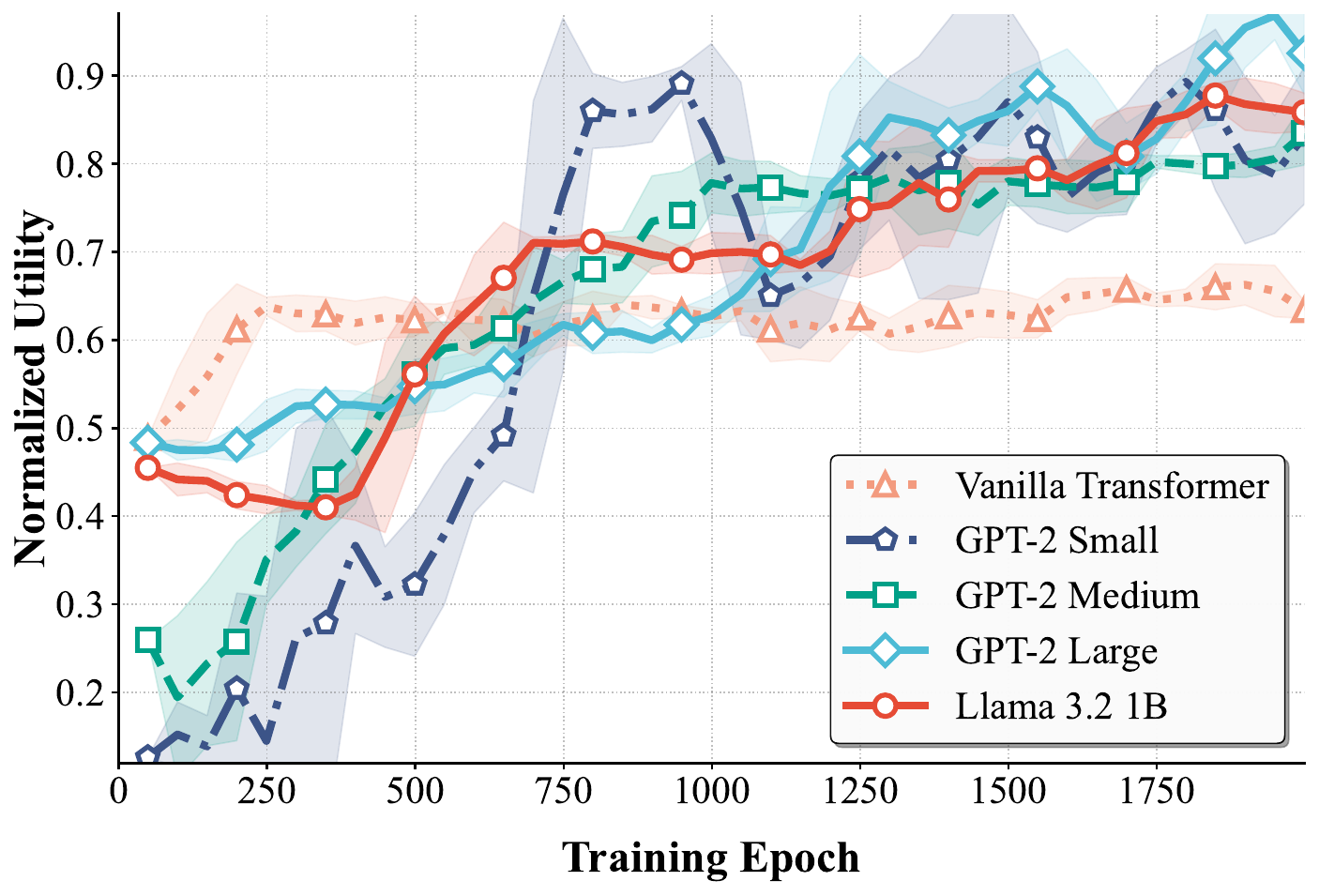}}
        \caption{Ablation study on policy backbones. (Left and Middle) Performance comparison of system utility and throughput  across different policy backbones and network scales. (Right) Convergence dynamics of normalized system utility for the $4$ UE scenario under different model architectures during training.}
        \label{fig: vanilla trans compare}
\end{figure*}
In Fig. \ref{fig: vanilla trans compare}, we investigate the impact of the underlying policy architecture on our framework. We introduce a ``Vanilla Transformer" baseline, which comprises standard, randomly initialized self-attention layers capable of processing variable-length sequences, and compare it against pre-trained LLMs ranging from varying parameter sizes. Detailed experimental setups are provided in the Appendix \ref{app: vanilla transformer}. As shown in Fig. \ref{fig: vanilla trans compare} (Left and Middle), the vanilla transformer consistently underperforms compared to other pre-trained counterparts, showing a tendency to overfit and poor generalization. This gap is further elucidated by the training dynamics in Fig. \ref{fig: vanilla trans compare} (Right), where the baseline appears to plateau earlier. This implies that pre-trained priors may offer a superior initialization, potentially steering the optimization trajectory away from the premature stagnation observed in the scratch-trained baseline. Moreover, the performance across different LLMs (from GPT-2 Small to Llama 3.2-1B) is remarkably consistent, suggesting that the task does not exhibit a strong scaling law; rather, a certain threshold of model capacity is sufficient.
\subsubsection{Impact of TBLER}
\begin{figure}[tb]
	\centering
	\includegraphics[width=.85\linewidth]{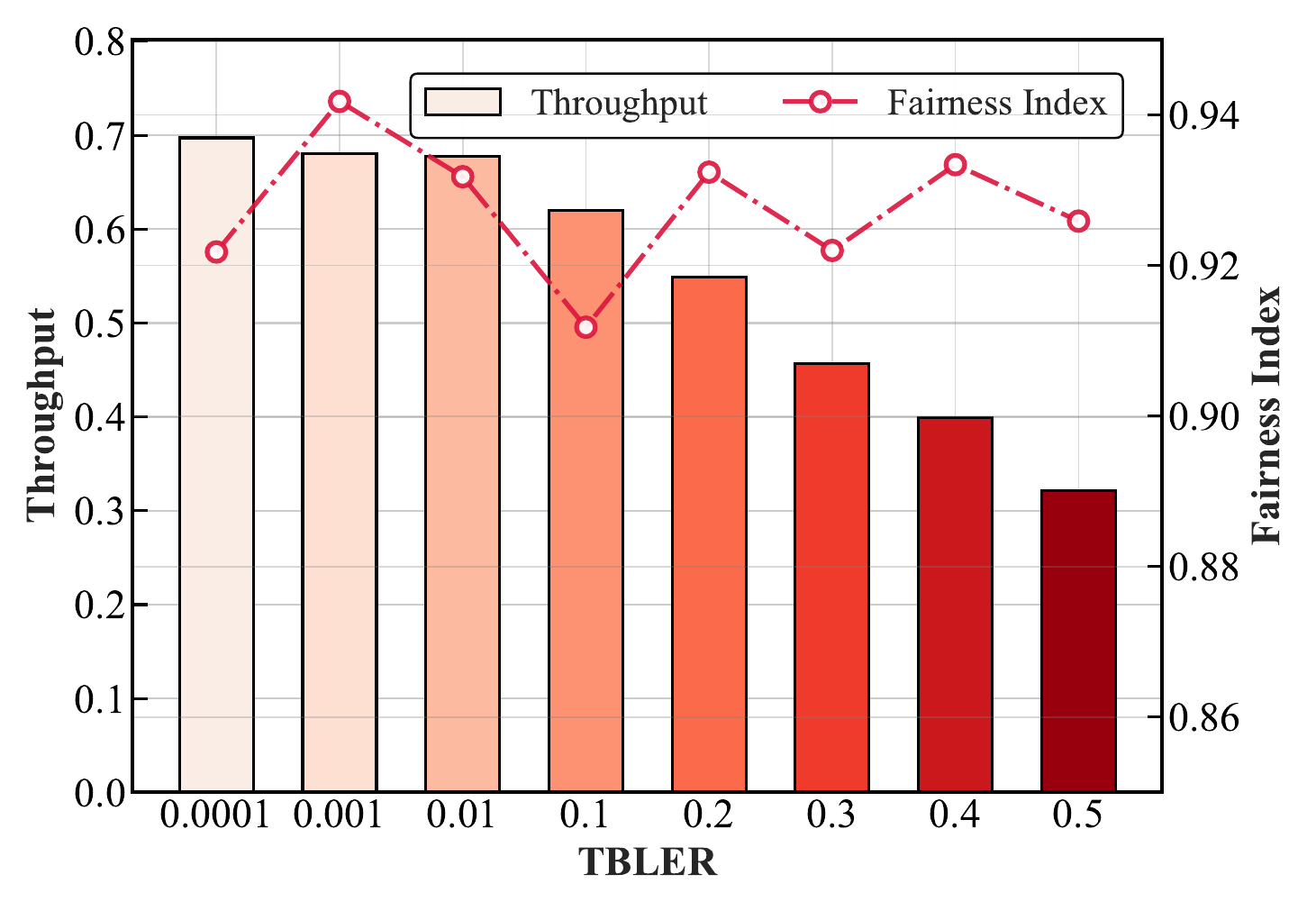}
	\caption{ Impact of TBLER on system throughput and fairness index.}
	\label{fig:diffTBLR}
\end{figure}
Fig. \ref{fig:diffTBLR} assesses the robustness of the proposed framework under varying TBLER conditions\footnote{\color{black}As standard practice in system-level link abstraction \cite{MP-maddpg,MP-scale}, TBLER serves to model residual MAC-layer packet erasures rather than decomposing PHY-layer sources.}. The results indicate that as the TBLER increases from $0.0001$ to $0.5$, the system throughput exhibits a smooth decline from approximately $0.7$ to $0.32$, in line with theoretical expectations. Notably, even under severe channel degradation where throughput is nearly halved, the fairness index consistently remains above $0.92$, thereby substantiating the robustness of the learned MAC protocol.

\section{Conclusion and Future Works}
In this paper, we have focused on the critical challenge of autonomously establishing generalizable and adaptive MAC protocols for dynamic wireless networks, where traditional MARL methods require costly retraining and redesign. We have proposed a novel LLM-empowered framework that recasts protocol emergence as a dynamic MFSG, capturing the intrinsic hierarchical interactions among network agents. By employing LLMs as policy networks within a PPO-driven RL loop, the framework natively handled variable-length inputs and outputs while preserving the exploratory learning required for protocol discovery. {\color{black}Theoretical analyses have guaranteed the existence of equilibria the local convergence of the proposed method under specific assumptions}. Simulation results have demonstrated that our method can robustly adapt to dynamic network conditions and consistently outperform various heuristic and conventional MAPPO baselines across diverse KPIs and environmental settings. {\color{black}Future work will investigate the longitudinal stability of distilled edge policies under repeated distillation–deployment cycles, with particular attention to teacher-anchored refresh strategies for mitigating potential policy drift.}

\bibliographystyle{IEEEtran}
\bibliography{ref}
\begin{IEEEbiography}[{\includegraphics[width=1in,height=1.25in,clip,keepaspectratio]{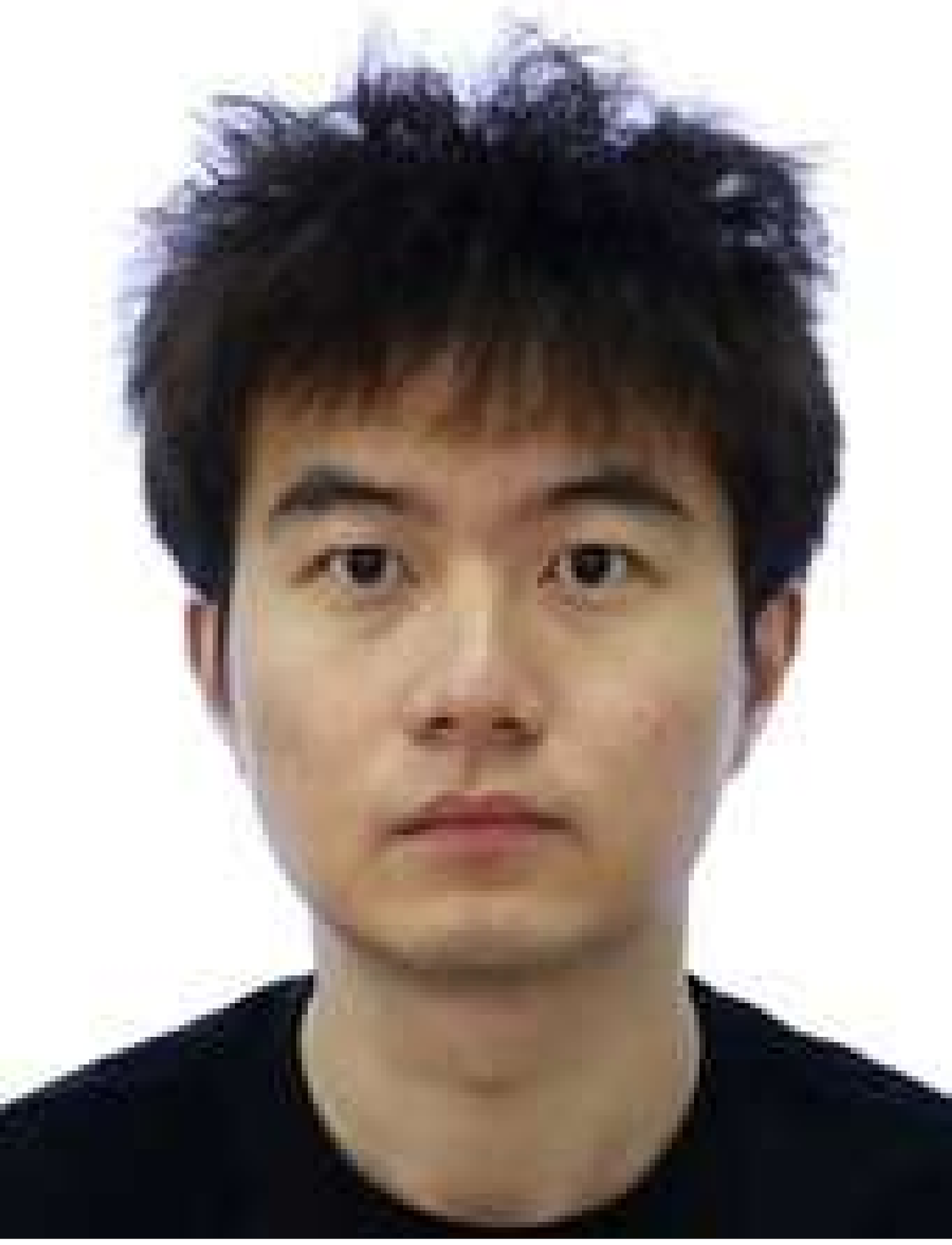}}]{Renxuan Tan}(Member, IEEE) received the
 B.E. degree in communication engineering from Sichuan University, Chengdu, China, in 2024. He is currently pursuing the Ph.D. degree with the College of Information Science and Electronic Engineering, Zhejiang University, Hangzhou, China. His research interests include reinforcement learning and large language models for wireless communications.
\end{IEEEbiography}
\vspace{-1cm}
\begin{IEEEbiography}[{\includegraphics[width=1in,height=1.25in,clip,keepaspectratio]{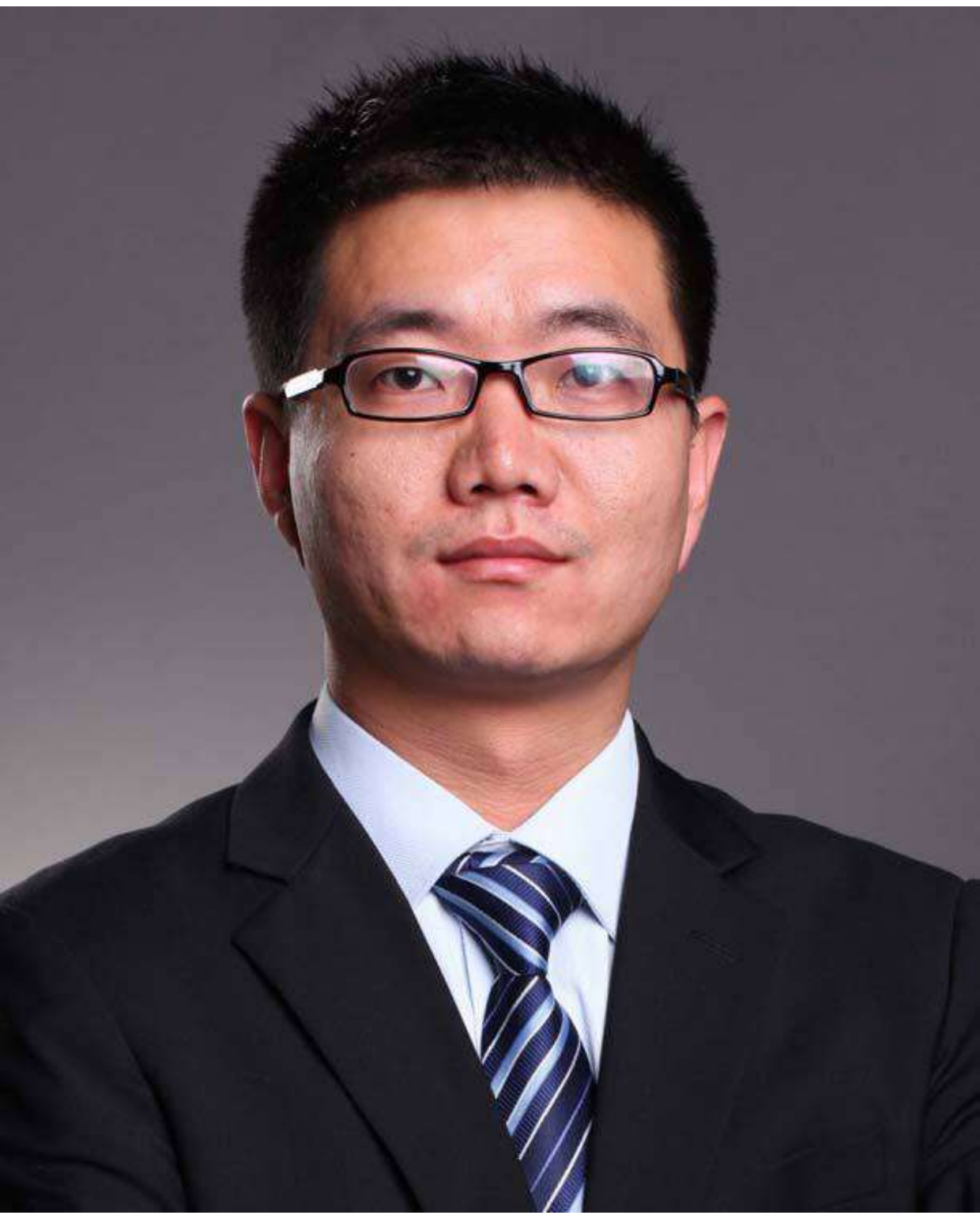}}]{Rongpeng Li}(Senior Member, IEEE) is currently an Associate Professor with the College of Information Science and Electronic Engineering, Zhejiang University. He received the B.E. degree from Xidian University, Xi’an, China, in June 2010, and the Ph.D. degree from Zhejiang University, Hangzhou, China, in June 2015. From August 2015 to September 2016, he was a Research Engineer with the Wireless Communication Laboratory, Huawei Technologies Company Ltd., Shanghai, China. He was a Visiting Scholar with the Department of Computer Science and Technology, University of Cambridge, Cambridge, U.K., from February 2020 to August 2020. His current research interests focus on networked intelligence for comprehensive efficiency (NICE).
\end{IEEEbiography}
\begin{IEEEbiography}[{\includegraphics[width=1in,height=1.25in,clip,keepaspectratio]{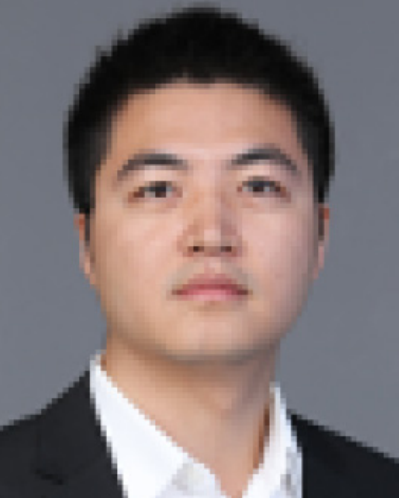}}]{Fei Wang} received the Doctoral degree from the University of Science and Technology of China, Hefei, China, in 2015. He is currently the Chief Researcher of Wireless Technology Lab, Huawei Technologies Co. Ltd., Shenzhen, China. His main research directions include 6G wireless network architecture, wireless distributed learning paradigm, federated learning, and large models.
\end{IEEEbiography}
\begin{IEEEbiography}[{\includegraphics[width=1in,height=1.25in,clip,keepaspectratio]{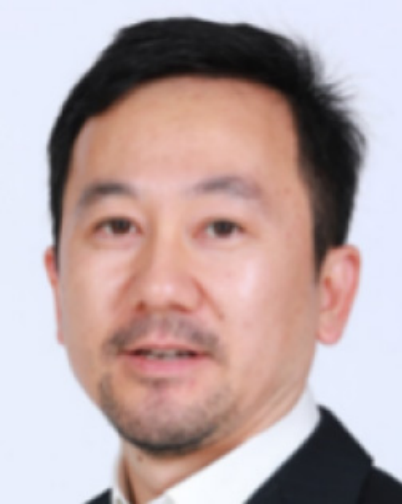}}]{Chenghui Peng} received the B.E. degree from the
University of Electronic Science and Technology of China, Chengdu, China, in 2001. He is an Expert Researcher of Wireless Technology Lab, Huawei Technologies Co. Ltd., Shenzhen, China. His main research directions include 6G wireless network architecture, task-oriented native intelligent architecture, mobile computing network, and wireless distributed learning paradigm. He has applied for more than 100 patents in the LTE, 5G, and 6G field.
\end{IEEEbiography}
\begin{IEEEbiography}[{\includegraphics[width=1in,height=1.25in,clip,keepaspectratio]{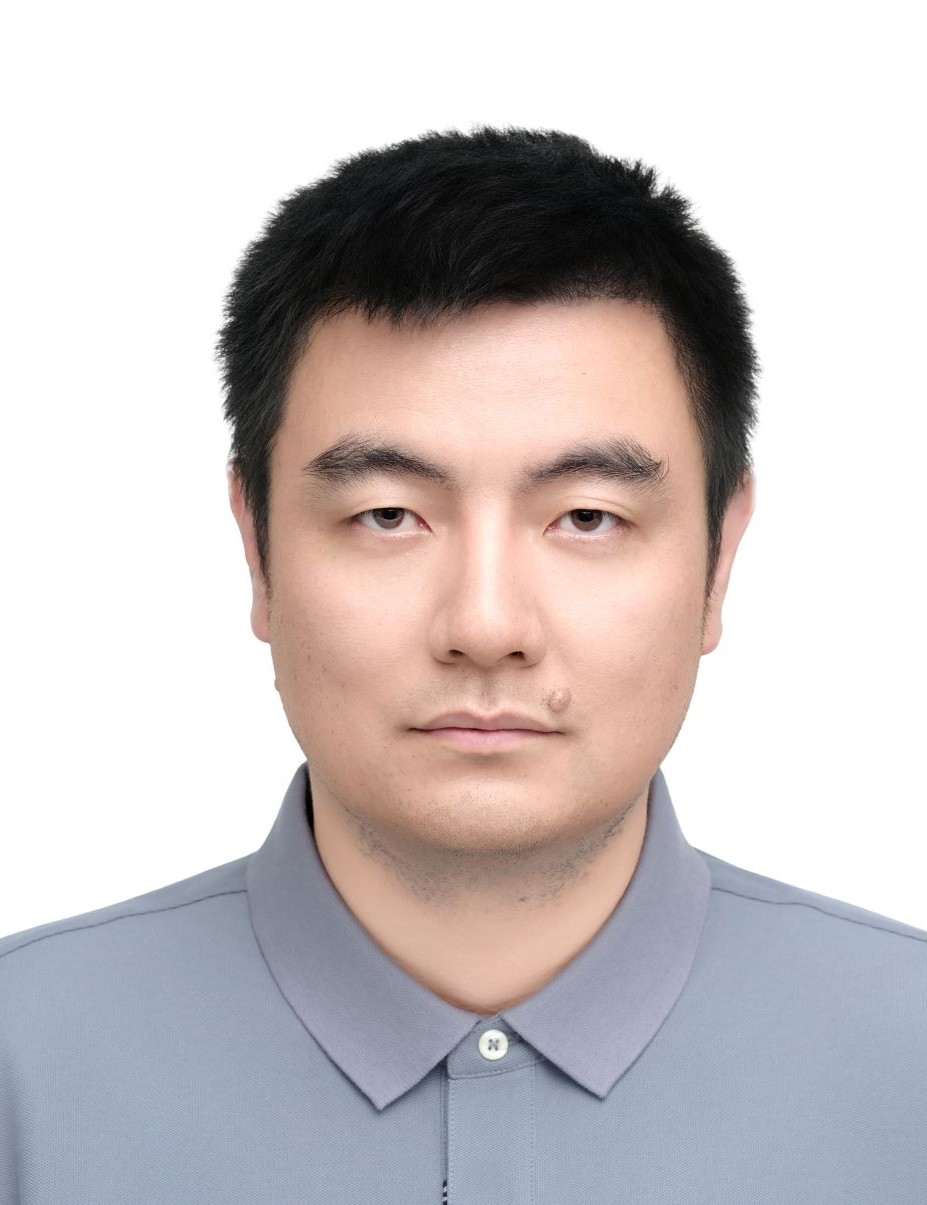}}]{Shaoyun Wu} is currently the Director of the Future Network Laboratory, the Chief 6G Network Architecture Expert, and a Senior Network Architect with Huawei Technologies Co., Ltd. He is primarily engaged in the research and system design of wireless communication networks. His research interests cover key technologies including connectivity, AI-native design, data services, computing services, security, and relevant protocols.
\end{IEEEbiography}
\begin{IEEEbiography}[{\includegraphics[width=1in,height=1.25in,clip,keepaspectratio]{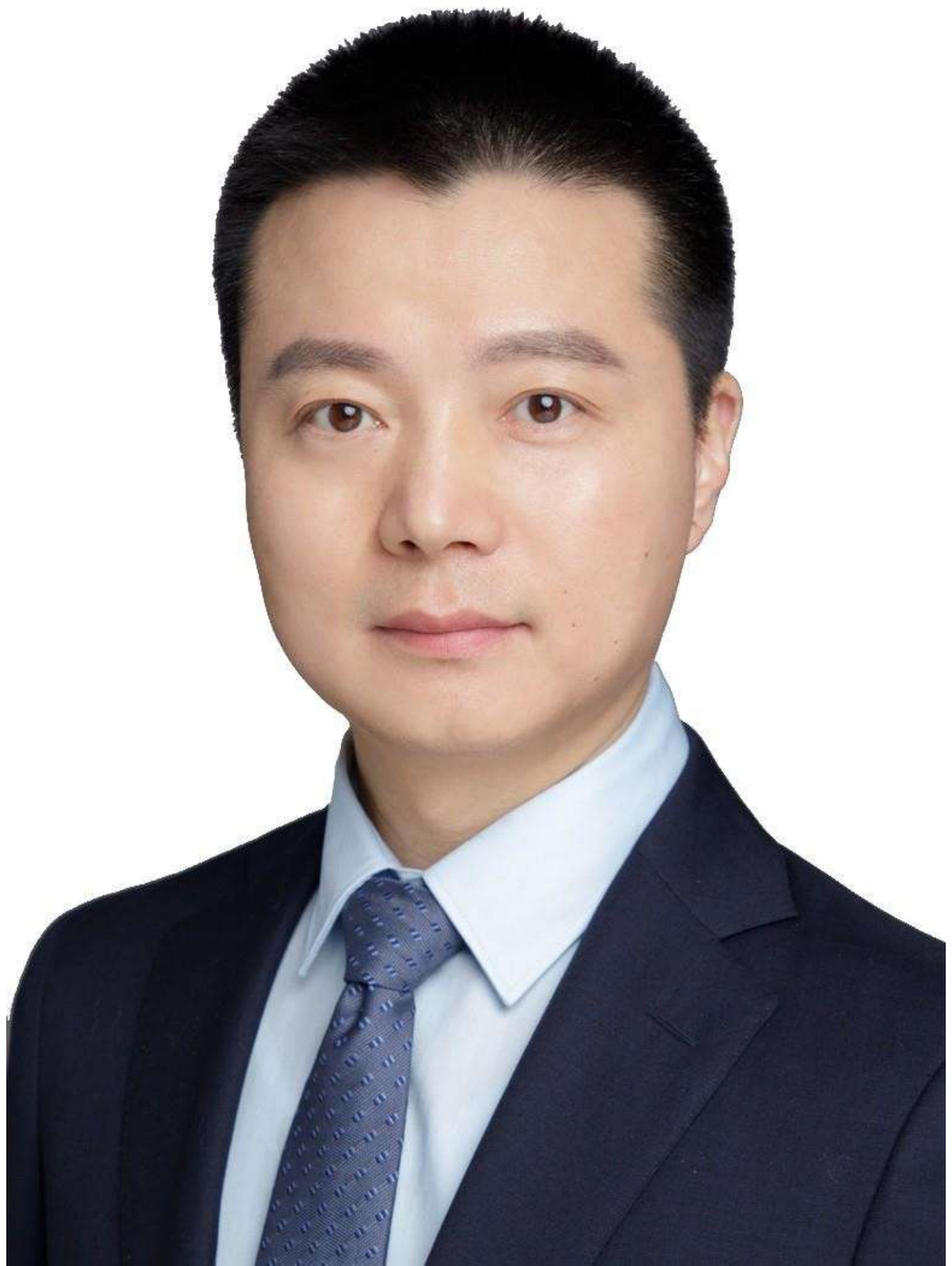}}]{Zhifeng Zhao}(Senior Member, IEEE) received the B.E. degree in computer science, the M.E. degree in communication and information systems, and the Ph.D. degree in communication and information systems from the PLA University of Science and Technology, Nanjing, China, in 1996, 1999, and 2002, respectively. From 2002 to 2004, he acted as a Post-Doctoral Researcher with Zhejiang University, Hangzhou, China, where his researches were focused on multimedia next-generation networks (NGNs) and softswitch technology for energy efficiency. Currently, he is with the Zhejiang Lab, Hangzhou as the Chief Engineering Officer. His research areas include software defined networks (SDNs), wireless network in 6G, computing networks, and collective intelligence. He is the Symposium Co-Chair of ChinaCom 2009 and 2010. He is the Technical Program Committee (TPC) Co-Chair of the 10th IEEE International Symposium on Communication and Information Technology (ISCIT 2010).
\end{IEEEbiography}
\begin{IEEEbiography}[{\includegraphics[width=1in,height=1.25in,clip,keepaspectratio]{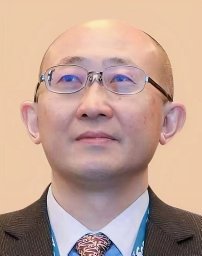}}]{Honggang Zhang}
(Fellow, IEEE) was a Professor with the College of Information Science and Electronic Engineering, Zhejiang University, Hangzhou, China. He was an Honorary Visiting Professor with the University of York, York, U.K., and an International Chair Professor of Excellence with the Universit\'{e} Europ\'{e}enne de Bretagne, Sup\'{e}lec, France. He is currently a Professor with the School of Computer Science and Engineering, Macau University of Science and Technology, Macau, China. His research interests include cognitive radio networks, semantic communications, green communications, machine learning, artificial intelligence, intelligent computing, and the internet of intelligence.

Dr. Zhang was a co-recipient of the 2021 IEEE Communications Society Outstanding Paper Award and the 2021 IEEE Internet of Things Journal (IoT-J) Best Paper Award. He served as the Chair for the Technical Committee on Cognitive Networks of the IEEE Communications Society from 2011 to 2012. He was the Founding Chief Managing Editor of \textit{Intelligent Computing} (Science Partner Journal). He was the leading Guest Editor of the Special Issues on Green Communications for \textit{IEEE Communications Magazine}. He served as a Series Editor for \textit{IEEE Communications Magazine} (Green Communications and Computing Networks Series) from 2015 to 2018. He is the Associate Editor-in-Chief of \textit{China Communications}.
\end{IEEEbiography}
\clearpage

\begin{appendices}
\section{Proofs}
For ease of reference, we first present a table summarizing the notations used throughout the proofs as Table \ref{tab:notations-appx}.

\subsection{Proof of Theorem 1}
\label{app: existence}
To establish the existence of an ESE, we first introduce two foundational lemmas.
\begin{lemma}[Weierstrass Theorem \cite{Weierstrass}]
\label{lemma:Weierstrass}
    A continuous function defined on a non-empty, compact set is guaranteed to attain its maximum and minimum values on that set.
\end{lemma}
\begin{lemma}[Implicit Function Theorem \cite{IFT}]
\label{lemma:implicit}
    Given a system $\mathscr{G}(\mathbf{x},\mathbf{y})=0$, if the Jacobian matrix of $\mathscr{G}$ with respect to $\mathbf{y}$, denoted by $\mathbf{J}_\mathscr{G}$, is non-singular at a point $(\mathbf{x},\mathbf{y})$ that satisfy the system, then there exist a unique, continuously differentiable function $h(\mathbf{x})$ in a neighborhood of $\mathbf{y}$ such that $\mathscr{G}(\mathbf{x},h(\mathbf{x}))=0$.
\end{lemma}

\begin{figure}[tb]
	\centering
	\includegraphics[width=\linewidth]{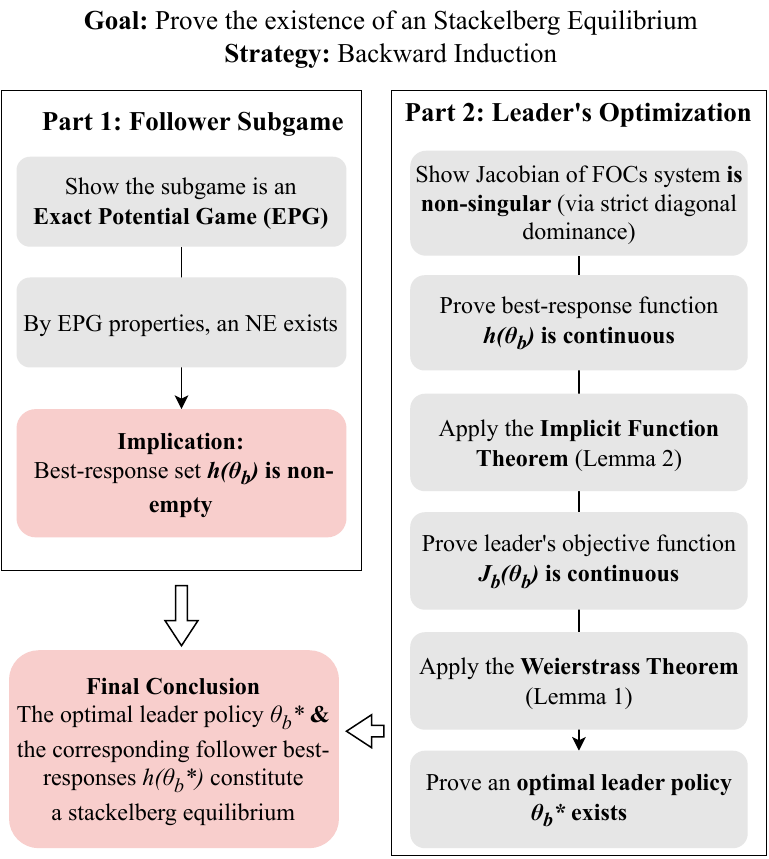}
	\caption{Proof roadmap for existence theorem}
	\label{fig: proof1}
\end{figure}

The mind map of the proof is illustrated in Fig. \ref{fig: proof1}. Inspired by backward induction, we hierarchically analyze the MFSG. We first prove the existence of NE in a multi-follower subgame with any follower numbers, and then establish the existence of a solution for the leader's problem. 
\begin{table}[tb]
	\caption{Notations Used in Appendix.}
	\label{tab:notations-appx}
	\begin{tabularx}{\columnwidth}{cX}
		\toprule
		Notation & Description  \\
		\midrule
        $\theta_b, \bm{\theta}_u$ & Specific policy parameters for the leader and followers\\
        $\bm{\varTheta}^{(e)}, \bm{\varTheta}^\ast$  & Joint policy parameters at epoch $e$ and equilibrium point, respectively\\
        $J_b(\cdot),J\subscriptiu(\cdot)$ & The expected long-term utility functions for the leader and follower $\UEn$\\
        $h(\theta_b)$ & The best-response of followers given the leader's policy $\theta_b$ \\
        $F^\prime\subscriptuet$ &  The interactive term in the follower's utility function, dependent on other followers' actions\\
        $\bm{a}^{\rm bit}$ & The joint transmission bitmaps $a^{\rm bit}\subscriptuet$ of all followers \\
        $\mathscr{E}(\cdot)$ & The potential function for follower subgame\\
        
		$\bm{\Omega}(\cdot)$ & The update gradient field for policy parameters\\
		$\varXi_m(\cdot)$ &The number of UEs selecting RBG $m$\\
        $\mathscr{C}_m(\cdot)$ & Payoff function at RBG $m$\\
        $N_m$ & The transmitted data volume at RBG $m$ \\ 
        $\mathscr{G}(\theta_b,h(\theta_b))$ &  The system of first-order conditions defining the followers' DNE given the leader's policy $\theta_b$\\
        $\varLambda(\bm{\varTheta})$ & The map of the discrete-time dynamical system governing the policy updates\\
        $\sigma(\cdot)$ & Singular value of a matrix\\
        $\mathbf{J}_{\mathscr{G}}$ & The Jacobian matrix of the FOC system $\mathscr{G}(\cdot)$\\
        $\mathbf{J}_{\bm \Omega}$ &The Jacobian matrix of the gradient field $\bm{\Omega}(\cdot)$ \\
        $\mathbf{M}(\bm{\varTheta})$ & The symmetric part of the Jacobian matrix $\mathbf{J}_{\bm \Omega}$\\
        $\kappa_1,\kappa_2$ & Positive constants derived from the singular values of $\mathbf{M}(\bm{\varTheta})$ and $\mathbf{J}_{\bm \Omega}$, related to the convergence rate\\
        $\iota_u$ &The time-scale separation factor for the followers' learning relative to the leader's \\
		\bottomrule
	\end{tabularx}
\end{table}
\begin{figure}[tb]
	\centering
	\includegraphics[width=0.9\linewidth]{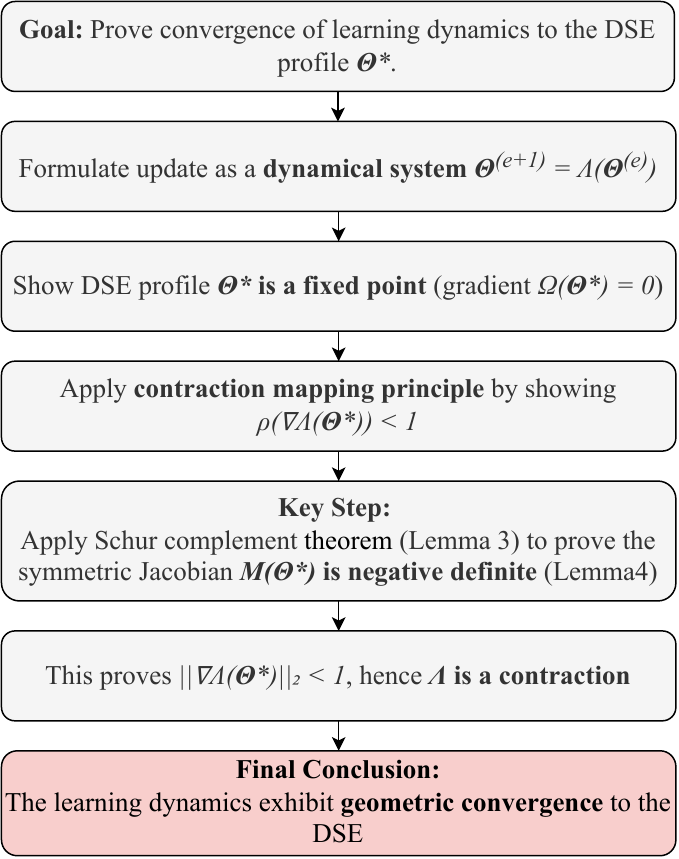}
	\caption{Proof roadmap for convergence theorem.}
	\label{fig: proof2}
\end{figure}
\subsubsection{Existence of NE in Follower Subgame}
Consider the subgame among followers under any fixed leader policy $\theta_b$ and any realized number of followers $\mathcal{I}_t=I$ drawn from the distribution $\mathbb{D}(I)$. We next show that this subgame is an exact potential game (EPG), which admits at least one NE \cite{9207866}. First, we focus on the interactive term $\xi^{\rm rec}\subscriptuet/\xi^{\rm tx}\subscriptuet$ from Eq. \eqref{eq:ue unity}. {\color{black}Note that although followers' actions collectively influence the leader's future observations across TTIs, within the instantaneous subgame at time $t$, the leader's broadcasted DCM $d_{i,t}$ acts as a realized, fixed parameter. Hence, $C_{i,t,u}$ is orthogonal to the concurrent actions of other followers $a_{-i,t,u}^{\rm bit}$, permitting its exact decoupling in the stage game.} To construct the potential function, we then reformulate this interactive utility $F'\subscriptuet$ as
\begin{equation}
	\label{eq:EPG ue}
	F'\subscriptuet (\bm{a}^{\rm bit};a\subscriptbst,I)= \sum_{m=1}^{M}a_{i,t,u,m}^{\rm bit}\Big(\prod_{j\in \mathcal{I}_t,j\neq i}(1-a_{j,t,u,m}^{\rm bit})N_m\Big),
\end{equation}
where $\bm{a}^{\rm bit}=(a^{\rm bit}\subscriptuet,a^{\rm bit}\subscriptuet[-i])$, $a^{\rm bit}\subscriptuet=(a_{i,t,u,1}^{\rm bit},\cdots,a_{i,t,u,M}^{\rm bit})$, and $N_m$ denotes data transmitted\footnote{We assume all UEs share the common $N_m$ for simplicity.} on RBG $m$. Based on this, we can define an exact potential function that captures the network throughput for the joint action profile $\bm{a}^{\rm bit}$
\begin{equation}
	\label{eq: potentil function}
	\mathscr{E}(\bm{a}^{\rm bit};a\subscriptbst,I) = \sum_{m=1}^{M}\sum_{j=1}^{\varXi_m(\bm{a}^{\rm bit})}\mathscr{C}_m(j),
\end{equation}
where $\varXi_m(\boldsymbol{a})$ is the number of UEs selecting RBG $m$, and the payoff function $\mathscr{C}_m(x) = N_m$ if $x=1$ and $0$ otherwise. The proposed potential function has a corresponding physical meaning, which is equal to the network throughput. Accordingly, it can be easily shown that for any UE $i$ and any unilateral change in its action from $a^{\rm bit}\subscriptuet$ to $\hat{a}^{\rm bit}\subscriptuet$, 
\begin{equation}
	\label{eq:EPG prove}
	\begin{aligned}
		&F'\subscriptuet(\hat{a}^{\rm bit}\subscriptuet, a^{\rm bit}\subscriptuet[-i];a\subscriptbst,I) - F'\subscriptuet(a^{\rm bit}\subscriptuet, a^{\rm bit}\subscriptuet[-i];a\subscriptbst,I) \\
		&= \mathscr{E}(\hat{a}^{\rm bit}\subscriptuet, a^{\rm bit}\subscriptuet[-i];a\subscriptbst,I) - \mathscr{E}(a^{\rm bit}\subscriptuet, a^{\rm bit}\subscriptuet[-i];a\subscriptbst,I),
	\end{aligned}
\end{equation}
confirming $\mathscr{E}$ as an exact potential function. {\color{black} Further incorporating the similarity term $C_{i,t,u}$ preserves the property. Since the broadcasted DCM $d_{i,t}$ is fixed within the stage game $t$, $C_{i,t,u}$ depends strictly on UE $i$'s own action. By defining a complete potential function $\mathscr{E}^* = \rho_1 \mathscr{E} + \rho_2 \sum_{k=1}^{I_t} C_{k,t,u}$, any unilateral deviation by UE $i$ yields a utility change $\Delta F_{i,t,u} = \rho_1 \Delta \mathscr{E} + \rho_2 \Delta C_{i,t,u} = \Delta \mathscr{E}^*$, which rigorously satisfies the EPG definition.} Therefore, for any leader policy $\theta_b$ and any finite number of followers $I$, there exists an NE among the follower subgame, which guarantees the set of followers' best-response policies $h(\theta_{b})$ is nonempty.

\subsubsection{Existence of an Optimal Universal Leader Policy}
We next analyze the leader's problem as shown in Eq. \eqref{eq: long-term bs unity}, which can be reformulated as the parameterized value function form
\begin{equation}
	\label{eq:111}
	\max_{\theta_b\in\Theta_b} J_b(\theta_b)=\mathbb{E}_{I\sim\mathbb{D}(I)}\left[ V_b(\theta_b,h(\theta_b),I)\right],
\end{equation}
where $V_b(\theta_b,h(\theta_b),I)$ is the long-term value function for a given instance $I$. By Lemma \ref{lemma:Weierstrass}, we just need to justify the continuity of $J_b$, which can be established by demonstrating the continuity of the value function $V_b$ for each $I$. The continuity of $V_b$, in turn, hinges on the continuity of the effective stage-game utility $F\subscriptbst(\theta_b,h(\theta_b), I)$, which depends on the followers' best-response function $h(\theta_b)$. 

Specifically, $h(\theta_b)$ is implicitly defined by the first-order conditions (FOCs) in Definition \ref{def:dse}, i,e,
\begin{equation}
	\label{eq: FOC}
	\mathscr{G}(\theta_b,h(\theta_b)) = \left(\nabla_{\theta_{\UEn[1]}}F\subscriptuet[1],\cdots,\nabla_{\theta_{\UEn[I]}}F\subscriptuet[I]\right)^\top = 0.
\end{equation}
According to Lemma \ref{lemma:implicit}, we discuss the Jacobian matrix of $\mathscr{G}$ with respect to $\bm{\theta}_u$ to show $h(\theta_b)$'s continuity, whcih is denoted by
\begin{equation}
	\label{eq:Jacob}
	\mathbf{J}_{\mathscr{G}} = \left[\nabla^2_{\theta_{\UEn[i]},\theta_{\UEn[j]}}F\subscriptuet(\theta_b,\bm{\theta}_u)\right]_{i,j=1}^I.
\end{equation} 
By Assumption \ref{assum: weak coupling} and the follower's second-order conditions (SOCs), we assert that $\mathbf{J}_\mathscr{G}$ is strictly diagonally dominant and thus non-singular, meeting the Lemma \ref{lemma:implicit}'s conditions. With $h(\theta_b)$ established as continuous and Assumption \ref{assum: soomth and compact}, it follows that $F\subscriptbst(\theta_b,h(\theta_b),I)$ is also continuous. For a discounted dynamic system, the continuity of the stage-game utility ensures that the corresponding discounted value function $V_b$ is also continuous with respect to $\theta_b$ \cite{PUTERMAN1990331}. Finally, since the value function $V_b$ is continuous for each $I$, the leader's objective $J_b$ --  being a finite weighted sum of these continuous functions over the distribution $\mathbb{D}(I)$ -- is itself continuous. With $J_b(\theta_b)$ confirmed as a continuous function over a compact set, the existence of an optimal universal leader policy $\theta_b^\ast$ is guaranteed \cite{Weierstrass}. Eventually, the combination of this optimal leader policy and the corresponding follower best responses constitutes an ESE.
\begin{figure*}[tb]
    \begin{equation}
		\label{eq:mat J omega}
		\mathbf{J}_{\bm{\Omega}} = 
		\left[
		\begin{array}{ll}
			\mathbf{J}_{11} & \mathbf{J}_{12} \\
			\mathbf{J}_{21} & \mathbf{J}_{22}
		\end{array}
		\right] 
		\!=\!\left[\begin{array}{c|ccc}
        \nabla_{\theta_b}^2 J_b & \nabla_{\theta_{u_1}, \theta_b}^2 J_b & \cdots & \nabla_{\theta_{u_I}, \theta_b}^2 J_b \\
        \hline \iota_u \nabla_{\theta_b, \theta_{u_1}}^2 J_{1, u} & \iota_u \nabla_{\theta_{u_1}}^2 J_{1, u} & \cdots & \iota_u \nabla_{\theta_{u_I}, \theta_{u_1}}^2 J_{1, u} \\
        \vdots & \vdots & \ddots & \vdots \\
        \iota_u \nabla_{\theta_b, \theta_{u_I}}^2 J_{I, u} & \iota_u \nabla_{\theta_{u_1}, \theta_{u_I}}^2 J_{I, u} & \cdots & \iota_u \nabla_{\theta_{u_I}}^2 J_{I, u}
        \end{array}\right]
        \!=\!
        \left[\begin{array}{cc}
			\nabla_{\theta_{b}}^2 J^{(e)}_b\left(\bm{\varTheta}^\ast\right) & \nabla_{\theta_{b}, \bm{\theta}_{u}}^2 J^{(e)}_b\left(\bm{\varTheta}^\ast\right) \\
			\iota_u \nabla_{\theta_{b}}\bm{\Omega}_u(\bm{\varTheta}^\ast)
			& 
			\iota_u \nabla_{\bm{\theta}_{u}}\bm{\Omega}_u(\bm{\varTheta}^\ast)
		\end{array}\right]
	\end{equation}
    \hrulefill
\end{figure*}

\subsection{Proof of Theorem 2}
\label{app: converge}
The mind map of the proof is illustrated in Fig. \ref{fig: proof2}. It is worth noting that Eq. \eqref{eq:stackel update rule} in the main text describes a discrete time dynamical system, $\bm{\varTheta}^{(e+1)} = \varLambda(\bm{\varTheta}^{(e)})$. The FOCs of Definition \ref{def:dse} yield $\bm{\Omega}(\bm{\varTheta}^\ast)=0$, which implies $\varLambda(\bm{\varTheta}^\ast)=\bm{\varTheta}^\ast$, establishing the DSE profile $\bm{\varTheta}^\ast$ as a fixed point \cite{fixedpont}. It is known from the theory of dynamical systems that if $\varLambda$ represents a contraction mapping proximate to a fixed point (i.e., spectral radius $\rho(\nabla\varLambda)<1$), then the system asymptotically converges to it \cite{dynamicsys}. We begin with some denotations and lemmas.
\begin{lemma}[Theorem 1.20 of Ref. \cite{horn2005basic}]
	\label{lemma: Schur theorem}
	Let $\mathbf{M}=
	\left[
	\begin{array}{ll}
		\mathbf{A} & \mathbf{B} \\
		\mathbf{B}^\top & \mathbf{D}
	\end{array}
	\right]$ be a symmetric matrix, where $\mathbf{A}$ is invertible.
	Then $\mathbf{M}$ is negative definite, i.e.,  $\mathbf{M} \prec 0$, if and only if both $\mathbf{A} \prec 0$  and the Schur-complement $\mathbf{D}-\mathbf{B}^{\top} \mathbf{A}^{-1} \mathbf{B} \prec 0$.
\end{lemma}
\begin{lemma}
	\label{lemma: M neg definite}
	Let the symmetric part of the Jacobian of the scaled gradient vector field $\bm{\Omega}(\bm{\varTheta})$ evaluated at $\bm{\varTheta}$ as $\mathbf{M}(\bm{\varTheta})=\frac{1}{2}\big(\mathbf{J}_{\bm{\Omega}}\left(\bm{\varTheta}\right)^\top+\mathbf{J}_{\bm{\Omega}}\left(\bm{\varTheta}\right)\big)$. The Jacobian of this gradient field, $\mathbf{J}_{\bm{\Omega}}(\bm \varTheta) = \nabla_{\bm \varTheta} \bm{\Omega}(\bm \varTheta)$, evaluated at point $\bm{\varTheta}^\ast$, is given by Eq. \eqref{eq:mat J omega}, where $\bm{\Omega}_u = \big(\nabla_{\theta_{\UEn[1]}} J^{(e)}\subscriptiu[1],\cdots,\nabla_{\theta_{\UEn[I]}} J^{(e)}\subscriptiu[I]\big)^\top$ is the followers part of gradient vector. We assure the block matrix $\mathbf{M}(\bm{\varTheta}^\ast)$ is negative definite.
\end{lemma}
See Appendix \ref{app: lemma2} for the entire proof.

Next, we just need to show that $\rho(\nabla\varLambda)<1$, i.e., considering its $L_2$-norm. Taking the gradient of dynamic system $\nabla \Lambda\left(\bm{\varTheta}^*\right)=\mathbf{I}+\alpha_b \mathbf{J}_{\bm{\Omega}}\left(\bm{\varTheta}^*\right)$, its squared $L_2$-norm 
\begin{align}
    & \Vert \mathbf{I}+\alpha_b \mathbf{J}_{\bm{\Omega}}(\bm{\varTheta}^\ast)\Vert_2^2\nonumber\\
    &=\sigma_{\max}\big(\left(\mathbf{I}+\alpha_b \mathbf{J}_{\bm{\Omega}}(\bm{\varTheta}^\ast)\right)^\top\left(\mathbf{I}+\alpha_b \mathbf{J}_{\bm{\Omega}}(\bm{\varTheta}^\ast)\right)\big) \nonumber\\
    & =\sigma_{\max}\left(\mathbf{I}+2\alpha_b \mathbf{M}(\bm{\varTheta}^\ast)+\alpha_b^2 \mathbf{J}_{\bm{\Omega}}^\top(\bm{\varTheta}^\ast) \mathbf{J}_{\bm{\Omega}}(\bm{\varTheta}^\ast)\right)\nonumber\\
    & \stackrel{(a)}{\leq} 1+2\alpha_b\sigma_{\max}\big(\mathbf{M}(\bm{\varTheta}^\ast)\big)+\alpha_b^2\sigma_{\max}\big(\mathbf{J}_{\bm{\Omega}}^\top(\bm{\varTheta}^\ast) \mathbf{J}_{\bm{\Omega}}(\bm{\varTheta}^\ast)\big)\nonumber\\
    & \stackrel{(b)}{=} \big(1-\frac{\kappa_1^2}{\kappa_2}\big),\nonumber
\end{align}
where $\sigma$ denotes the singular value. (a) implies Weyl's inequality \cite{horn1994topics}. Let $\kappa_1 = \sigma_{\min} (-\mathbf{M}(\bm{\varTheta}^\ast))$, $\kappa_2 =\sigma_{\max} (\mathbf{J}_{\bm{\Omega}}^\top\left(\bm{\varTheta}^\ast\right) \mathbf{J}_{\bm{\Omega}}\left(\bm{\varTheta}^\ast\right))$ and $\alpha_b = \frac{\kappa_1}{\kappa_2}$. By Lemma \ref{lemma: M neg definite}, $\mathbf{M}(\bm{\varTheta}^\ast) \prec 0$, its singular value are negative, so that both $\kappa_1,\kappa_2>0$. Furthermore, 
\begin{equation}
	\kappa_1^2 \leq \sigma_{\max}^2\big(-\mathbf{M}(\bm{\varTheta}^\ast)\big)\leq\sigma_{\max}\big(\mathbf{J}_{\bm{\Omega}}\left(\bm{\varTheta}^\ast\right)^\top \mathbf{J}_{\bm{\Omega}}\left(\bm{\varTheta}^\ast\right)\big) = \kappa_2,
\end{equation}
which means $\frac{\kappa_1^2}{\kappa_2} \in (0,1)$, and we claim that (b) holds.
Therefore, $\rho\big(\nabla\Lambda(\bm{\varTheta}^\ast)\big)\leq\sqrt{\big(1-\frac{\kappa_1^2}{\kappa_2}\big)}<1$, which establishes that $\nabla\Lambda(\bm{\varTheta}^\ast)$ is a contraction.

Consider the Taylor expansion of $\Lambda(\bm{\varTheta})$ around $\bm{\varTheta}^\ast$, we have
\begin{equation}
	\begin{aligned}
		&\Vert\bm{\varTheta}^{(e+1)}-\bm{\varTheta}^\ast\Vert_2 = \Vert\Lambda(\bm{\varTheta}^{(e)}) - \Lambda(\bm{\varTheta}^\ast)\Vert_2\\
		&=\Vert\big(\mathbf{I}+\alpha_b \mathbf{J}_{\bm{\Omega}}\left(\bm{\varTheta}^*\right)\big) (\bm{\varTheta}_{t}-\bm{\varTheta}^\ast) + o(\bm{\varTheta}_{t}-\bm{\varTheta}^\ast)\Vert_2\\
		&\leq \Vert\big(\mathbf{I}+\alpha_b \mathbf{J}_{\bm{\Omega}} \left(\bm{\varTheta}^*\right)\big)\Vert_2 \Vert (\bm{\varTheta}_{t}-\bm{\varTheta}^\ast)\Vert_2 + \Vert o(\bm{\varTheta}_{t}-\bm{\varTheta}^\ast)\Vert_2 \\
		&\approx\sqrt{\big(1-\frac{\kappa_1^2}{\kappa_2}\big)}\Vert (\bm{\varTheta}_{t}-\bm{\varTheta}^\ast)\Vert_2.
	\end{aligned}
\end{equation}
This leads to the geometric convergence rate $O\big(\sqrt{\big(1\!-\!\frac{\kappa_1^2}{\kappa_2}\big)^e}\big)$.
\subsection{Proof of Lemma \ref{lemma: M neg definite}}
	\label{app: lemma2}
Consistent with the formulation in Lemma \ref{lemma: Schur theorem}, we first examine the sub-blocks of $\mathbf{M}(\bm{\varTheta}^\ast)$. The top-left block, $\mathbf{A}=\frac{1}{2}(\mathbf{J}_{11}+\mathbf{J}_{11}^\top) = \nabla_{\theta_{b}}^2 J^{(e)}_b\left(\bm{\varTheta}^\ast\right)$, is the Hessian of the BS with respect to its strategy parameters. Similarly, he bottom-right block $\mathbf{D}_\iota = \iota_u \mathbf{D}= \frac{1}{2}(\mathbf{J}_{22}+\mathbf{J}_{22}^\top)$, and its $(j,k)$-th sub-block is denoted by $\frac{\iota_u}{2} \big(\nabla_{\theta\subscriptiu[j],\theta\subscriptiu[k]}^2 J^{(e)}\subscriptiu[j]  +  (\nabla_{\theta\subscriptiu[k],\theta\subscriptiu[j]}^2 J^{(e)}\subscriptiu[k])^\top \big)$. From the SOCs of the underlying SDE, it is established that $\mathbf{A}$ and the diagonal principal sub-blocks of $\mathbf{D}_\iota$ are negative definite. Assumption \ref{assum: weak coupling} stipulates that the norms of the cross second-order gradients between different UEs are negligible\footnote{See Appendix \ref{app: weak premise} for a detailed physical justification}. {\color{black}The weak-coupling premise is invoked around the expected Stackelberg equilibrium, where the learned DCM-UCM interaction has already formed a quasi-coordinated intent-response pattern.} Consequently, $\mathbf{D}_\iota$ is strictly diagonally dominant, which, in turn, implies that $\mathbf{D}_\iota$ is also negative definite. 

According to Lemma \ref{lemma: Schur theorem}, it is now sufficient to demonstrate that $\mathbf{D}_\iota-\mathbf{B}^{\top} \mathbf{A}^{-1} \mathbf{B} \prec 0$. Let $\mathbf{x}$ a non-zero vector, we analyze $\mathbf{x}^\top (\mathbf{D}_\iota-\mathbf{B}^{\top} \mathbf{A}^{-1} \mathbf{B})\mathbf{x} < 0$ as follow:
	\begin{subequations}
		\begin{align}
			& \mathbf{x}^\top \mathbf{D}_\iota \mathbf{x} < \mathbf{x}^\top (\mathbf{B}^{\top} \mathbf{A}^{-1} \mathbf{B})\mathbf{x} \label{Za}\\
			\Rightarrow & \iota_u\sigma_{\min}(-\mathbf{D}) > \sigma_{\max}(-\mathbf{B}^{\top} \mathbf{A}^{-1} \mathbf{B}) \label{Zb}\\
			\Rightarrow & \iota_u\sigma_{\min}(-\mathbf{D}) >  \Vert\mathbf{B}\Vert_2^2 \Vert\mathbf{A}^{-1}\Vert_2 \label{Zc}\\
			\Rightarrow &\iota_u\sigma_{\min}(-\mathbf{D})\sigma_{\min}(-\mathbf{A}) > \Vert\mathbf{B}\Vert_2^2\label{Zd}
		\end{align}
	\end{subequations}
Eq. \eqref{Zb} leverages Rayleigh quotient properties \cite{horn1994topics} to establish singular value bounds for the matrix quadratic terms. Eq. \eqref{Zc} then applies the submultiplicativity of the spectral norm \cite{horn1994topics}, yielding $\sigma_{\max}(-\mathbf{B}^{\top} \mathbf{A}^{-1} \mathbf{B})=\Vert\mathbf{B}^{\top} \mathbf{A}^{-1} \mathbf{B} \Vert_2 \leq \Vert\mathbf{B}\Vert_2^2 \Vert\mathbf{A}^{-1}\Vert_2$. Eq. \eqref{Zd} involves substituting the identity $\Vert \mathbf{A}^{-1}\Vert_2 = \sigma_{\min}^{-1}(-\mathbf{A})$ into Eq. \eqref{Zc}. An appropriate choice of $\iota_u$ then ensures the resulting inequality holds. Therefore, the Schur complement of $\mathbf{M}(\bm{\varTheta}^\ast)$ is also negative definite, confirming that $\mathbf{M}(\bm{\varTheta}^\ast)$ itself is negative definite.
\subsection{Proof of Corollary 1}
\label{app: corollary 1}
In the dynamic MFSG with stochastically varying followers, the epoch-wise gradient field is formulated as
\begin{equation}
	\bm{\Omega}(\bm{\varTheta}^{(e)}, \bar{I}_e) = \left( \nabla_{\theta_b} J^{(e)}_b, \iota_u \left( \frac{1}{\bar{I}_e} \sum_{i=1}^{\bar{I}_e} \nabla_{\theta_u} J^{(e)}\subscriptiu \right) \right)^\top,
\end{equation}
where $\bar{I}_e$ denotes the batch-wise expected number of followers. For convergence analysis, Eq. \eqref{eq:stackel update rule} is rewritten as the average dynamics
\begin{equation}
	\bm{\varTheta}^{(e+1)} = \bm{\varTheta}^{(e)} + \alpha_b\cdot\bar{\bm{\Omega}}(\bm{\varTheta}^{(e)}),
\end{equation}
where $\bar{\bm{\Omega}}(\bm{\varTheta}) = \mathbb{E}_{\bar{I}_e \sim \mathbb{D}(I)}[\bm{\Omega}(\bm{\varTheta}, \bar{I}_e)]$ is the averaged gradient vector. The pivotal step in the proof of Theorem \ref{thm:converge} is showing that the symmetric part of the Jacobian $\mathbf{M}(\bm{\varTheta}^\ast)$ is negative definite. Here, this key variable also transforms into its expected form. 
\begin{equation}
	\bar{\mathbf{M}}(\bm{\varTheta}^\ast) \triangleq \mathbb{E}_{\bar{I}_e}\left[\mathbf{M}(\bm{\varTheta}^\ast,\bar{I}_e)\right].
\end{equation}
Lemma \ref{lemma: M neg definite} ensures that $\mathbf{M}(\bm{\varTheta}^\ast, I)$ is negative definite for any realization $I$. Since expectations over the distribution are convex combinations of these matrices and the cone of negative definite matrices is convex, it follows that $\bar{\mathbf{M}}(\bm{\varTheta}^\ast)$ is negative definite. Let $\kappa_1 = \sigma_{\rm min}(-\bar{\mathbf{M}}(\bm{\varTheta}^*))$ and $\kappa_2 = \sigma_{\rm max}(\bar{\mathbf{J}}_{\bm{\Omega}}(\bm{\varTheta}^*)^{\top} \bar{\mathbf{J}}_{\bm{\Omega}}(\bm{\varTheta}^*))$, we obtain a convergence result similar to Theorem \ref{thm:converge}. Crucially, the validity of this entire analysis hinges on the LLM-enabled universal policy, whose fixed-dimensional parameter space $\bm{\varTheta}$ renders the expectation over Jacobians from different game instances mathematically coherent.

\subsection{Proof of Theorem 3}
\label{app: robustness proof}
We analyze the deviation by decomposing it into distributional drift and estimation error, by triangle equation:
\begin{equation}
    \label{eq: triangle decomp}
    | J_\mathcal{D} - \hat{J}^{(K)}_{\mathcal{D}'} | \le | J_\mathcal{D} - J_\mathcal{D}' | + | J_\mathcal{D}' - \hat{J}^{(K)}_{\mathcal{D}'} |.
\end{equation}
To bound the total violation probability $\mathbb{P}(| J_\mathcal{D} - \hat{J}^{(K)}_{\mathcal{D}'} | \geq \delta)$, we apply the union bound. Thus,
\begin{align}
\label{eq: union bound}
    &\mathbb{P}( |J_{\mathcal{D}} - \hat{J}_{\mathcal{D}'}^{(K)}| \geq \delta ) \leq \\
    &\mathbb{P}( | J_\mathcal{D} - J_\mathcal{D}' | \geq \frac{\delta}{2} ) + \mathbb{P}( | J_\mathcal{D}' - \hat{J}^{(K)}_{\mathcal{D}'} | \geq \frac{\delta}{2} ).\notag
\end{align}
We first examine the first term. Based on Assumption 4, $\mathcal{D}, \mathcal{D}' \sim {\rm Dir}(\chi_1,\chi_2,\cdots,\chi_{|\mathcal{S}|})$, the expected utility $J_{\mathcal{D}}(\boldsymbol{\theta}^*) = \sum_{s \in \mathcal{S}} \mathcal{D}(s) F_{t,b}(\boldsymbol{\theta}^*, s)$ is a linear functional of a Dirichlet-distributed random vector. Let $\mu_s = \mathbb{E}[\mathcal{D}(s)]$, by the properties of Dirichlet distribution \cite{ng2011dirichlet}, the covariance between any two components $\mathcal{D}(s)$ and $\mathcal{D}(s')$ is given by
\begin{equation}
    {\rm Cov}(\mathcal{D}(s), \mathcal{D}(s')) = 
    \begin{cases} 
        \frac{\mu_s(1-\mu_s)}{1+ \sum_{i=1}^{|\mathcal{S}|} \chi_i } & \text{if } s = s', \\ 
        \frac{-\mu_s \mu_{s'}}{1+\sum_{i=1}^{|\mathcal{S}|} \chi_i} & \text{if } s \neq s'. 
    \end{cases}
\end{equation}
The variance of $J_{\mathcal{D}}$ is derived by
\begin{align}
\label{eq: varJd}
    &{\rm Var}(J_{\mathcal{D}}) 
    = \sum_{s \in \mathcal{S}} \sum_{s' \in \mathcal{S}} F_{t,b}(s) F_{t,b}(s') \text{Cov}(\mathcal{D}(s), \mathcal{D}(s')) \notag\\
    &=  \sum_{s \in \mathcal{S}}F_{t,b}^2(s)\frac{\mu_s(1-\mu_s)}{1+\sum_{i=1}^{|\mathcal{S}|} \chi_i} \!+\!\sum_{s\neq s'}F_{t,b}(s)F_{t,b}(s')\frac{-\mu_s \mu_{s'}}{1+\sum_{i=1}^{|\mathcal{S}|} \chi_i}  \notag \\
    &= \frac{1}{1 + \sum_{i=1}^{|\mathcal{S}|} \chi_i} \Big[\sum_s\mu_sF_{t,b}^2(s)-\big(\sum_s\mu_sF_{t,b}(s)\big)^2 \Big]\notag\\
    &=\frac{1}{1 + \sum_{i=1}^{|\mathcal{S}|} \chi_i}\sum_{s \in \mathcal{S}} \mathbb{E}[\mathcal{D}(s)] \left( F_{t,b}(s) - \mathbb{E}[J_{\mathcal{D}}] \right)^2.
\end{align}
The summation term $\sum_{s} \mathbb{E}[\mathcal{D}(s)] ( F_{t,b}(s) - \mathbb{E}[J_{\mathcal{D}}] )^2$ represents the variance of the scalar values $F_{t,b}(s)$ with respect to the mean distribution $\mu$. Since $F_{t,b} \in [0, F_{\max}]$, we invoke Popoviciu's inequality, which states that for any random variable $X$ bounded in $[a, b]$, its variance satisfies ${\rm Var}(X) \leq \frac{1}{4}(a-b)^2$. Applying this to Eq. \eqref{eq: varJd}, we yield ${\rm Var}(J_{\mathcal{D}}) \le \frac{F_{\max}^2}{4(1 + \sum_{i=1}^{|\mathcal{S}|} \chi_i)}$. Given that $\mathcal{D}$ and $\mathcal{D}'$ are independent, we have
\begin{equation}
    {\rm Var}(|J_{\mathcal{D}} - J_{\mathcal{D}'}|) \le \frac{F_{\max}^2}{2(1 + \sum_{i=1}^{|\mathcal{S}|} \chi_i)}\le \frac{F_{\max}^2}{2(1 + |\mathcal{S}|\chi_{\epsilon})}.
\end{equation}
The second inequality holds as $\sum_{i=1}^{|\mathcal{S}|} \chi_i \ge |\mathcal{S}|\chi_{\epsilon}$. Applying Chebyshev’s inequality with threshold $\delta/2$:
\begin{equation}
\mathbb{P}\big( \left| J_{\mathcal{D}} - J_{\mathcal{D}'} \right| \ge \frac{\delta}{2} \big) \le \frac{2F_{\max}^2}{\delta^2 (1 + |\mathcal{S}|\chi_{\epsilon})}.    
\end{equation}
We next bound the estimation error term $| J_{\mathcal{D}'} - \hat{J}_{\mathcal{D}'}^{(K)} |$. The term $\hat{J}_{\mathcal{D}'}^{(K)} = \frac{1}{K} \sum_{k=1}^K F_{t,b}(s_k)$ is the empirical mean of $K$ independent random variables bounded in $[0, F_{\max}]$. By Hoeffding’s inequality, for a deviation of $\delta/2$:
\begin{equation}
    \mathbb{P}( | J_{\mathcal{D}'} - \hat{J}_{\mathcal{D}'}^{(K)} | \ge \frac{\delta}{2} ) \le 2 \exp \big( -\frac{K \delta^2}{2F_{\max}^2} \big).
\end{equation}
Finally, by the Eq. \eqref{eq: union bound}, the probability that the total deviation exceeds $\delta$ is upper-bounded by the sum of the tail probabilities of the constitutive terms, completing the proof:
\begin{equation}
    \mathbb{P}\big( |J_{\mathcal{D}} - \hat{J}_{\mathcal{D}'}^{(K)}| \geq \delta \big) \leq \frac{2 F_{\max}^2}{\delta^2(1 + |\mathcal{S}|\chi_{\epsilon})} + 2 \exp \big( - \frac{K \delta^2}{2 F_{\max}^2} \big).
\end{equation}
\section{Implementation Details}
\label{app: details setup}
In this section, we elaborate on some specific experimental setups and implementations, with all key experimental parameters and hyperparameters summarized in Table \ref{tab:sim_params}.
\begin{table}[t]
	\caption{Simulation Parameters.}
	\centering
    \label{tab:sim_params}
    \renewcommand{\arraystretch}{1.2}
	\begin{tabularx}{\linewidth}{X cr} 
		\toprule
		\textbf{Parameter} & \textbf{Symbol} & \textbf{Value}\\
		\midrule 
		Number of UEs & $I$ & $\{3,4,5\}$ \\
		Transport block size (bit) & $\xi^{\rm gen}$ & 256\\
		Buffer capacity of each UE & $B_{\rm cap}$ & $\infty$ \\
		TTI duration & \textrm{TTI} & $5\times10^{-3}$ s \\
		The arrive rate of dPDU & $p_a$ & $[0.05,0.8]$ \\
		Number of RBGs & $M$ & $5$ \\
		Reward discount factor & $\gamma$ & $0.95$ \\
        GAE discount factor & $\lambda$ & $0.9$ \\
		Transport block error rate & TBLER & $10^{-3}$ \\
		Clipping parameter of PPO & $\epsilon$ & $0.5$ \\
		Actor learning rate & $\alpha,\beta$  & $5\times 10^{-5}$ \\
        Critic learning rate & $\beta$  & $5\times 10^{-4}$ \\
        Subcarriers per RB & $N^{\rm rb}_{\rm sc}$  & $12$\\
		Symbols per slot &  $N^{\rm sh}_{\rm symbol}$  & $14$\\
        Numerology &     & $1$(30kHz)\\
        Channel changing frequency &  & $[1,10]$ ms \\
  		Total training epochs & $e^{\max}$ & $2000$ \\
  		Weighting parameter &$\rho_1,\rho_2,\varepsilon$ & $8,10,0.5$ \\
        Value loss coef. & & $0.8$ \\
        Entropy coef. & & $0.05$ \\
		PPO update epochs &   & $5$\\
		  Mini-batch size &  & $128$ \\
		\bottomrule
	\end{tabularx}
	
\end{table}
\subsection{Data Generation Mechanism}
Unlike supervised learning approaches that rely on static offline datasets, our framework generates simulation data online through real-time interactions between the LLM agents and the UDTS environment. Each UE and BS agent performs $5$ local PPO updates after collecting every $20$ episodes, with an episode horizon of $24$ and a PPO batch size of $32$.
Over the course of $e^{\rm max}=2,000$ training epochs, the agents collect approximately $9.6\times 10^5$ interaction steps. The input prompts are dynamically constructed from numerical observations as defined in Eq. \eqref{eq: pag embedding}, ensuring the agents learn strictly from environmental feedback without reliance on external domain-specific corpora.
\subsection{Details of Channel Dynamics}
To ensure the fidelity of the simulation regarding time-varying wireless environments, we assume the channel quality evolution of each UE as a discrete-time Markov chain (DTMC) within a finite state space $\mathcal{S} = \{\text{poor, medium, good}\}$. The evolution of the channel state is governed by a transition probability matrix $\mathbf{P}$, 
\begin{equation}
\label{eq:DTMC}
    \mathbf{P} = 
    \begin{bmatrix}
        0.85 & 0.10 & 0.05 \\
        0.05 & 0.90 & 0.05 \\
        0.05 & 0.10 & 0.85
    \end{bmatrix}
\end{equation}
where $P_{mn} = \mathbb{P}(c_{i,t+1,u} = s_n | c\subscriptuet = s_m)$, and the state changing frequency is distributed in $[1, 10]$ ms. This formulation captures the essential temporal correlations of fading channels and abstracts the continuous physical layer into semantic states, ensuring that the environment exhibits realistic stability intervals rather than chaotic, uncorrelated fluctuations. To abstract the MCS selection, we ideally assign discrete spectral efficiency values $\nu_{i,t} \in \{0.15, 0.45, 0.85\}$ bits/s/Hz to the poor, medium, and good states, respectively\footnote{This configuration is specifically selected to accommodate the transmission of small, fixed-size data payloads $\xi^{gen}=256$ bits. These values correspond to the low-index MCS entries specified in 3GPP TS 38.214 \cite{3GPP38214}, representing a robust transmission regime typical of cell-edge users or power-limited IoT devices.}. {\color{black}Crucially, the simulation incorporates imperfect and asymmetric CSI. The BS observes an estimated state $\hat{c}_{i,t,u}$, which deviates from the true local state $c_{i,t,u}$ due to the temporal misalignment between channel updates. Specifically, the true channel state evolves according to the DTMC in Eq. \eqref{eq:DTMC}, with state transitions occurring at random intervals uniformly sampled from $[1, 10]$ ms, whereas the BS can only acquires its CSI at the boundaries of each scheduling TTI ($5$ ms). This mismatch directly resulting in CSI aging and an inherent information asymmetry between the BS (leader) and UEs (followers).}
\subsection{Details of User Behavior}
\label{sec: user behavior}
 We consider only two distinct types of user behavior, with the difference being manifested exclusively in their decision-making processes. Under our method and MAPPO-S/G baselines, the agentic UEs are governed by a learnable policy $\pi\subscriptiu$ that synthesizes transmission actions $a\subscriptuet^{\rm bit}$ based on the POMDP framework. In contrast, under the heuristic baseline, we assume the legacy UEs adhere to an enhanced S-ALOHA with an initial transmission probability $p_{\rm tx}^{\rm init}=0.5$. The transmission probability evolves recursively based on the feedback. Upon a collision, the retransmission probability is reduced via a backoff factor $\delta_{\rm bo}=0.95$ (i.e., $p_{\rm tx}^{\rm new} = p_{\rm tx}^{\rm curr} \cdot \delta_{\rm bo}$), subject to a minimum floor $p_{\rm tx}^{\rm min}=0.1$, and resets to $p_{\rm tx}^{\rm init}$ immediately upon a successful transmission. Regardless of the control logic, all UEs manage data queues via a strict FIFO discipline, where successfully acknowledged bits are removed from the queue head as shown in Section \ref{sec: system model}.

\subsection{Details of Traffic Generation}
We formally define the traffic generation process as a discrete-time stochastic sequence. Let $\Gamma\subscriptit$ denote the arriving payload (in bits) for UE $u_i$ at time slot $t$, and let the dPDU size $\xi^{\rm gen} =256$. The three packet generation models used in our simulation are specified as follows:
\begin{itemize}
    \item Bernoulli Arrival Process: The arrival follows an independent and identically distributed (i.i.d.) Bernoulli sequence with parameter $p_a$. The mathematical expression is
    \begin{equation}
    \Gamma\subscriptit = \xi^{\rm gen} \cdot \mathbb{I}(X_{i,t} < p_a),
    \end{equation}
    where $X_{i,t} \sim \mathcal{U}[0,1]$ is a uniform random variable sampled at each slot, and $\mathbb{I}(\cdot)$ is the indicator function. By default, the simulation cyclically assigns the values $p_a\in\{0.1,0.3,0.7\}$ to UEs to model heterogeneous traffic loads.
    \item Periodic Arrival Process: Traffic is generated at strict, regular intervals defined by a period $T_{\rm per}=3$. The arrival function is defined using modulo arithmetic
    \begin{equation}
        \Gamma\subscriptit = \xi^{\rm gen} \cdot \mathbb{I}(t\pmod{T_{\rm per}} \equiv 0).
    \end{equation}
    \item Burst Arrival Process: Unlike the single-packet periodic model, we model the arrival as a discrete impulse train of packet batches. The arrival payload $\Gamma\subscriptit$ is expressed as a summation of weighted impulses occurring at a fixed on-off cycle $T_{\rm on/off}$
    \begin{equation}
        \Gamma\subscriptit=\sum_{n=0}^{\infty} \xi^{\rm agg} \cdot \delta[t - n \cdot T_{\rm on/off}],
    \end{equation}
    where $\delta[\cdot]$ is impulse function, and $\xi^{\rm agg} = K \cdot \xi^{\rm gen}$ represents the aggregated burst payload. We configure $K=3$ and $T_{\rm on/off}=5$ in our setting.
\end{itemize}
The mathematical models defined above are selected to comprehensively capture the heterogeneous traffic profiles typical of 5G/6G networks \cite{embb,shen2023five}. Specifically, the Bernoulli, periodic, and burst arrival processes inherently correspond to the standardized traffic patterns of massive machine-type communications (mMTC), ultra-reliable low-latency communications (URLLC), and enhanced mobile broadband (eMBB) services, respectively \cite{popovski20185g}. By evaluating the protocol against these diverse generation patterns, we verify its robust adaptability across the full spectrum of next-generation service requirements.
\subsection{Implementation of MAPPO and G-MAC Baselines}
We implement the MAPPO and G-MAC baselines under the centralized training with decentralized execution (CTDE) paradigm. While both methods operate on identical observation and action spaces, derived from the system model in Section \ref{sec: system model}, they differ fundamentally in the neural architecture of the BS agent to address the generalization challenge.
\subsubsection{UE Policy Implementation}
The UE agent employs a shared policy network across all users to facilitate number generalization and efficient learning \cite{FeasibleMAC, Miuccio2022LearningGW}. To mitigate partial observability, the input to the policy consists of a history stack ($\text{horizon}=3$) containing the local buffer status, estimated spectral efficiency, and previous actions. This input is processed by an MLP backbone with four hidden layers of sizes $[128, 128, 256, 256]$ and using ReLU activations, followed by a linear layer that maps the features to a 512-dimensional latent representation. These features are then distributed to three distinct output heads, yielding: a discrete transmission indicator $a\subscriptuet^{\rm tx} \in \{0,1\}$, a discrete UCM $u\subscriptit$ selected from a vocabulary of size $5$, and crucially, a resource block bitmap $a\subscriptuet^{\rm bit} \in \{0,1\}^M$. To handle the combinatorial complexity of selecting non-contiguous RBGs, we model the bitmap policy using a Bernoulli distribution head, which outputs $M$ independent probabilities simultaneously, rather than treating the bitmap as a categorical distribution over $2^M$ options. 
\subsubsection{BS Policy Implementation}
The implementation of the BS policy diverges significantly between MAPPO-G/S and G-MAC. The MAPPO baseline utilizes a monolithic MLP architecture designed for a fixed maximum UE capacity ($I_{\max}$). Its backbone is a $5$-layer MLP with hidden sizes of $[128, 128, 256, 256]$ and a final output dimension of $512$, which ingests a concatenated vector of the global resource state history and the flattened observation histories of all users. The network terminates in a multi-head structure consisting of $I_{\max}$ parallel linear layers, where the $i$-th head independently outputs the DCM logits for the $i$-th UE, and the DCM vocabulary size is set to $7$. For the MAPPO-G variant, which evaluates generalization, we employ zero-padding or truncation when the actual UE number $I_t$ differs from the training configuration $I_{\max}$.

In contrast, G-MAC adopts a sequential architecture to achieve generalization across varying UE numbers \cite{GMAC}. Instead of a fixed-input MLP, the G-MAC BS policy decomposes the joint decision-making process into a sequence of user-wise steps modeled by a gated recurrent unit (GRU). Specifically, a global encoder first processes the RBG state history into a context vector. For each $\UEn$, a user-specific encoder projects their feature history into a $256$-dimensional embedding. This embedding is concatenated with a learned embedding of the DCM action taken for the previous UE, forming a $512$-dimensional input for the GRU cell (hidden size $256$). Both encoders contain two hidden layers of size $[256, 256]$ and a $256$-dimensional linear layer. This auto-regressive structure allows the BS to unroll the policy for an arbitrary sequence length $I_t$ while sharing parameters, thereby enabling the handling of environment shift without architectural changes.

\subsubsection{Training Details}
To ensure efficient and stable training, we employ a multi-threaded environment sampling strategy that parallelizes data collection across multiple CPU cores. For each PPO update, we aggregate $200$ rollouts from the parallel environments to form a training batch. We utilize the Adam optimizer with a learning rate of $2 \times 10^{-4}$ and a minibatch size of $1024$. The training process is constrained to a maximum of $10,000$ updates, incorporating an early stopping mechanism. The optimization process also uses a discount factor of $\gamma=0.99$, a GAE parameter $\lambda=0.95$, and a PPO clipping ratio of $\epsilon=0.2$.
\subsection{Configuration of Vanilla Transformer Experiments}
\label{app: vanilla transformer}
The vanilla transformer baseline employs a standard Transformer architecture initialized from scratch, processing numerical tokens without incorporating pre-trained language. Both the policy and value network are implemented as a $6$-layer Transformer decoder with 8 attention heads, a hidden embedding dimension of $256$, and a feed-forward dimension of $1,024$, utilizing standard sinusoidal positional encodings and Xavier uniform initialization. Training utilizes the PPO algorithm optimized via Adam with a learning rate of $3 \times 10^{-4}$, and it takes the same number of episodes. For the environmental settings, we configure $p_a=0.63$ and initialize the buffer to $10,000$, while other key parameters follow the default configuration. The computational overhead of using different policy backbones is summarized in Table \ref{tab:computation_cost}.
\begin{table*}[tb]
  \centering
  \caption{Performance comparison under heterogeneous scenarios where a subset of agentic UEs, operating under the Stackelberg game framework, coexist with legacy UEs using heuristic access. The results are averaged over three independent runs.}
  \label{tab:heterogenous_ues}
  \renewcommand{\arraystretch}{1.2}
  \begin{tabular}{c c c c c c c}
    \toprule
    \multirow{2}{*}{\textbf{Experimental Group}} & \multicolumn{2}{c}{$p_a = 0.3$} & \multicolumn{2}{c}{$p_a = 0.5$} & \multicolumn{2}{c}{$p_a = 0.7$} \\
    \cmidrule(lr){2-3} \cmidrule(lr){4-5} \cmidrule(lr){6-7}
    & Throughput & Fairness Index & Throughput & Fairness Index & Throughput & Fairness Index \\
    \midrule
    5 Agentic          & $0.633 \pm 0.067$ & $0.953 \pm 0.022$ & $1.021 \pm 0.037$ & $0.960 \pm 0.028$ & $1.282 \pm 0.047$ & $0.963 \pm 0.030$ \\
    4 Agentic + 1 Legacy& $0.536 \pm 0.009$ & $0.872 \pm 0.012$ & $0.727 \pm 0.062$ & $0.863 \pm 0.030$ & $0.810 \pm 0.080$ & $0.897 \pm 0.093$ \\
    3 Agentic + 2 Legacy& $0.503 \pm 0.076$ & $0.896 \pm 0.009$ & $0.492 \pm 0.078$ & $0.898 \pm 0.019$ & $0.519 \pm 0.084$ & $0.897 \pm 0.043$ \\
    1 Agentic + 4 Legacy& $0.411 \pm 0.036$ & $0.933 \pm 0.017$ & $0.451 \pm 0.030$ & $0.766 \pm 0.111$ & $0.378 \pm 0.050$ & $0.866 \pm 0.147$ \\
    5 Legacy           & $0.402 \pm 0.021$ & $0.871 \pm 0.046$ & $0.297 \pm 0.005$ & $0.632 \pm 0.149$ & $0.326 \pm 0.063$ & $0.651 \pm 0.185$ \\
    \bottomrule
  \end{tabular}
\end{table*}
\begin{table}[htbp]
  \centering
  \caption{Sensitivity analysis on fairness-throughput trade-off via reward shaping. Results are averaged over three independent runs.}
  \label{tab:varepsilon sensitivity}
  \setlength{\tabcolsep}{10pt}
  \begin{threeparttable}
  \begin{tabular}{ccc}
    \toprule
    \textbf{Method} & \textbf{Throughput (Mbps)} & \textbf{Fairness Index} \\
    \midrule
    \multicolumn{3}{l}{\textit{Baselines}} \\
    Max Weight & $2.688 \pm 0.050$ & $0.385 \pm 0.010$ \\
    Round Robin & $1.547 \pm 0.030$ & $0.993 \pm 0.001$ \\
    S-ALOHA & $1.364 \pm 0.090$ & $0.953 \pm 0.020$ \\
    \midrule
     \multicolumn{3}{l}{\textit{LLM-Scheduler}} \\
     Ours ($\varepsilon=0.0$) & $\mathbf{2.912 \pm 0.007}$ & $0.287 \pm 0.021$ \\
     Ours ($\varepsilon=0.1$) & $2.849 \pm 0.030$ & $0.328 \pm 0.019$ \\
      Ours ($\varepsilon=0.3$) & $2.650 \pm 0.040$ & $0.403 \pm 0.019$ \\
      Ours ($\varepsilon=0.4$) & $2.553 \pm 0.035$ & $0.492 \pm 0.011$ \\
      Ours ($\varepsilon=0.5$) & $2.275 \pm 0.016$ & $0.537 \pm 0.018$ \\
      Ours ($\varepsilon=0.8$) & $1.629 \pm 0.034$ & $0.952 \pm 0.013$ \\
      Ours ($\varepsilon=1.0$) & $1.495 \pm 0.005$ & $\mathbf{1.000 \pm 0.001}$ \\
    \bottomrule
  \end{tabular}
  \begin{tablenotes}
      \footnotesize
      \item[1] Max-Weight (MW): A throughput-optimal policy that prioritizes UEs based on the product of their queue length and spectral efficiency.
      \item[2] Round-Robin (RR): A fairness-centric policy that cyclically allocates resources to UEs regardless of their channel states, guaranteeing equal access opportunities.
      \item[3] S-ALOHA: A contention-based heuristic policy, the specific implementation of which is shown in the Appendix \ref{sec: user behavior}. 
    \end{tablenotes}
  \end{threeparttable}
\end{table}

\begin{table*}[htb]
\centering
\caption{Computational overhead under different model scales, with both the number of UEs and the number of RBGs set as $5$. The training metrics are measured on NVIDIA A800.}
\label{tab:computation_cost}
\renewcommand{\arraystretch}{1.05}
\begin{tabular}{l|c|cc|cc|c}
\toprule
\multirow{2}{*}{\textbf{Model}} & \textbf{Model Scale} & \multicolumn{2}{c|}{\textbf{Training Cost (Offline)}} & \multicolumn{2}{c|}{\textbf{Inference Cost (Online)}} & \textbf{Performance} 
\\ 
\cline{2-7} 
 & \textbf{Params}  & \textbf{Peak Memory} & \textbf{Time per episode} & \textbf{Decision Delay} & \textbf{Active Memory} & \textbf{System Utility} 
\\ 
\hline
Vanilla-Transformer & 6.4M  & 886 MB & 1.76 s  &  1.83 ms  & 223 MB  & 0.56  \\

GPT-2-Small & 124M  & 26.78 GB & 4.76 s  & 6.23 ms  & 2102 MB  &  0.66 \\
GPT-2-Medium & 345M  & 36.75 GB & 5.36 s  & 8.13 ms  & 3769 MB  &  0.83 \\
GPT-2-Large & 774M  & 50.12 GB & 6.56 s  & 10.63 ms  & 7424 MB  &  0.80 \\
Llama-3.1 & 1.0B  & 54.46 GB & 6.72 s  &  12.85 ms  & 10.81 GB  &  0.79 \\ 
\toprule
\end{tabular}
\end{table*}
\section{Additional Experimental Results}
\subsection{Impact of Heterogeneous UE}
We further evaluate the adaptability of the proposed framework in a heterogeneous environment where the Stackelberg game is engaged by only a subset of the network entities. In this setup, agentic UEs continue to interact with the BS via the proposed LLM-based signaling to optimize their strategies, while coexisting legacy UEs operate independently using a rigid heuristic policy. As illustrated in Table \ref{tab:heterogenous_ues}, the system exhibits graceful degradation rather than abrupt failure as the prevalence of non-cooperative legacy nodes increases. Specifically, under a moderate traffic load ($p_a=0.3$), the transition from a fully agentic network to a mixed environment (4 Agentic + 1 Legacy UE) incurs an acceptable performance degradation, with throughput and fairness decreasing by approximately $14\%$ and $8.5\%$, respectively. A partial deployment with even a single agentic UE is sufficient to yield perceptible throughput gains over the pure legacy baseline, particularly under high traffic loads of $p_a=0.7$. These empirical results substantiate that the learned leader-follower equilibrium remains effective and beneficial even when applied to a restricted subset of users within a heterogeneous network. 
\subsection{Impact of Weighting Parameters}
To rigorously assess the trade-off between network efficiency and fairness, Table \ref{tab:varepsilon sensitivity} details a sensitivity analysis on the weighting parameter $\varepsilon$ (defined in Eq. \eqref{bs:ue unity}) under a controlled stress-test environment\footnote{To amplify the conflict between throughput and fairness explicitly, we construct a static stress test scenario. We let five UEs be initialized with sufficiently large buffers ($b\subscriptit=30,000$) to ensure continuous contention, and are assigned fixed, highly heterogeneous spectral efficiencies of $\nu\subscriptit\in [0.15,0.45, 0.85, 1.0, 2.0]$ bits/s/Hz. This setup necessitates a harder decision between maximizing system rate and preventing starvation than dynamic random-arrival settings.}. In the efficiency-dominant regime ($\varepsilon=0$), our framework acts similarly to the throughput-optimal MW (max-weight) policy by prioritizing users with better channel conditions, surpassing the MW by approximately $4.3\%$. Conversely, as $\varepsilon$ increases towards 1.0, the learned policy progressively shifts focus towards perfect fairness, eventually achieving a JFI of $1.0$ that matches the fairness-centric RR (round-robin) algorithm. Crucially, at a balanced setting of $\varepsilon=0.5$, the framework maintains a high fairness index of $0.735$ while delivering $2.2$ Mbps throughput, representing an approximately $40\%$ performance gain over the heuristic S-ALOHA. {\color{black}This confirms that our protocol does not rigidly sacrifice efficiency for fairness; instead, it provides a flexible, learnable continuum that allows operators to navigate the Pareto frontier between the high-throughput extremes of MW and the high-fairness extremes of RR.}
\subsection{Analysis of Policy Distillation}
\label{app: distillation}
\begin{figure}[b]
    \centering
    \includegraphics[width=0.8\linewidth]{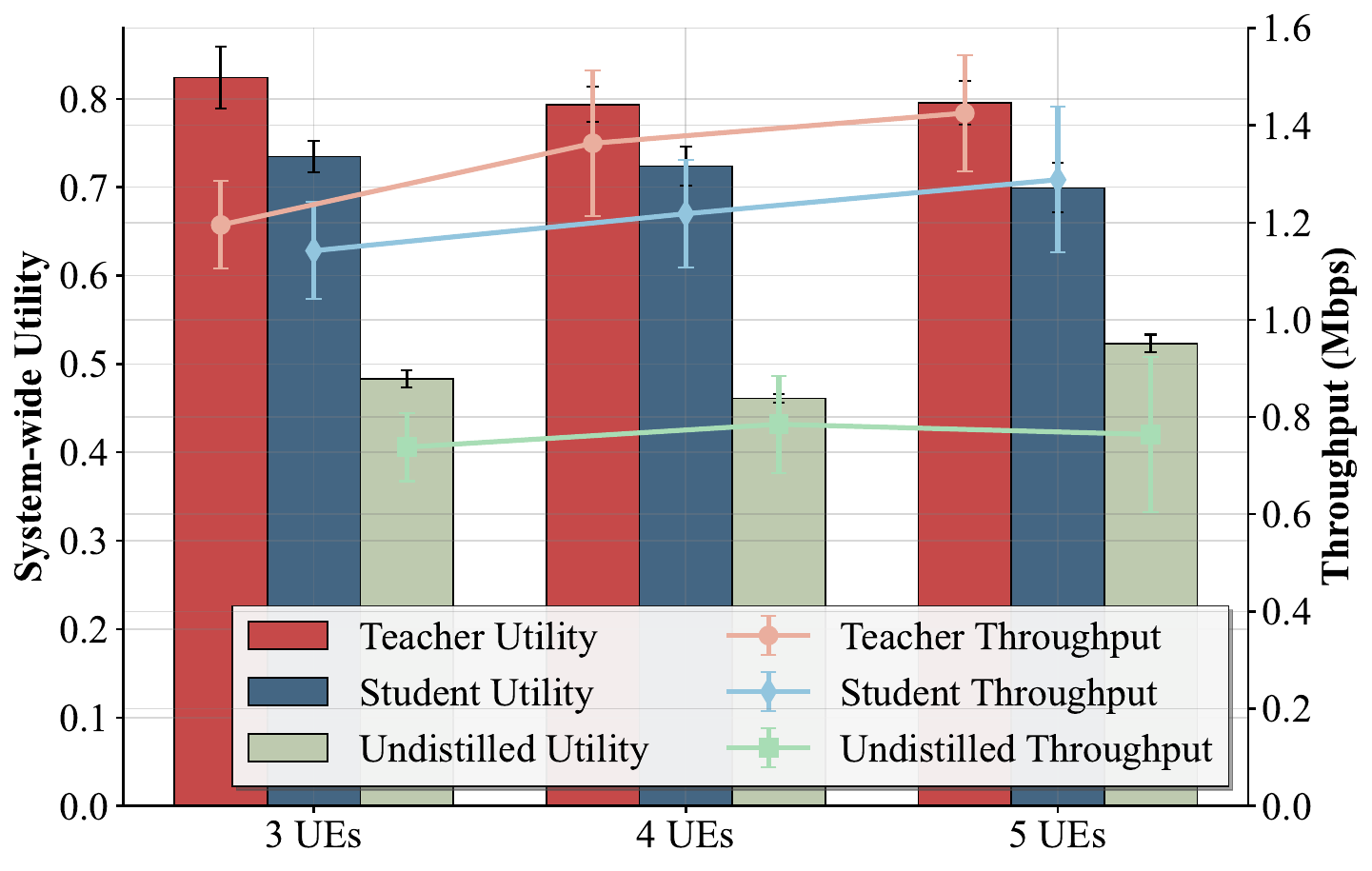}
    \caption{Performance validation of policy distillation in terms of the system-wide utility and throughput over three independent runs. We employ the Llama-3.1-1B (Teacher) to supervise the lightweight vanilla Transformer (Student), defined in Section V-B-6, via trajectory-based joint policy-value optimization.}
    \label{fig:distillation}
\end{figure}
To proactively mitigate potential deployment constraints, we conduct exploratory experiments on distilling LLMs' emergent capabilities into a more compact architecture. As illustrated in Fig. \ref{fig:distillation}, {\color{black}a single distilled student retains the teacher's generalization capabilities across various UE scenarios without requiring retraining, successfully achieving approximately $92-95\%$ of the system-wide utility and throughput and surpassing the undistilled baseline.} Crucially, as quantified in Table \ref{tab:computation_cost}, this structural compression slashes inference latency from $12.85$ ms to $1.83$ ms, offering a promising path toward meeting the strict timing constraints of MAC protocols. This knowledge distillation paradigm secures the superior exploration and generalization capabilities of LLMs, which scratch-trained lightweight models often fail to achieve, while maintaining the execution efficiency essential for practical deployment.
\section{Additional Discussions}
\subsection{\color{black}Scalability and Architectural Evolution}
\label{app: Scalability}
While our empirical results in Table \ref{tab:multi_cell_scalability} indicate that the learned LLM-based policies can transfer to multi-cell scenarios via basic hierarchical resource partitioning, this HCN evaluation is intended as a scalability-transfer study rather than a formal multi-leader MFSG. In the current setting, the CBS only serves as an exogenous tier-0 coordinator that partitions the global RBG pool using fixed heuristic policies, without participating in joint learning or strategic optimization. Conditioned on a given CBS partition, each DBS operates over its assigned RBG subset and the corresponding DBS-UE subsystem reduces to the local MFSG analyzed in Section IV-E. The shared edge UEs coupling under the fixed-partition setting is absorbed into the stochastic local environment rather than modeled as an endogenous DBS-level strategic interaction. Therefore, the existence and local convergence guarantees in Section IV-E apply to each DBS-local MFSG under a fixed partition, while a fully coupled multi-cell guarantee is not claimed. A rigorous large-scale extension may be formulated as a two-tier nested Stackelberg game, with the CBS as a super-leader, DBSs as local leaders, and UEs as followers. Preserving convergence in such a hierarchy would require additional regularity conditions. Future work may further combine this formulation with mean-field games \cite{mean-filed1,mean-field2} or HFL \cite{9054634}, where collective state distributions or a ``Macro-Aggregator, Micro-Learner'' paradigm can improve scalability and distributed adaptability.

\subsection{Discussion on Flexible Interaction Models}
The proposed MFSG formulation relies on the inherent leader-follower hierarchy of cellular uplink standards (e.g., 3GPP TS 36.321 \cite{3GPP36321}), where the BS holds the grant authority. However, we acknowledge that this model may not fully capture the complexities of purely heterogeneous or ad-hoc networks. For scenarios requiring high degrees of peer-to-peer autonomy, coalition game approaches or mixed games become more applicable than the strict Stackelberg paradigm. These models allow UEs to form cooperative clusters for interference mitigation or load balancing. In our future work, we will explore a hybrid framework that nests a horizontal coalition game (among UEs) within the vertical Stackelberg game (between BS and UEs), thereby reconciling centralized control with distributed cooperative incentives.
\end{appendices}
\subsection{Feasibility and Practical Deployment}
\label{sec:feasibility_discussion}
The practical implementation of LLM-empowered MAC protocols hinges on addressing computational cost, communication overhead, and inference latency constraints. Regarding computational feasibility, the intensive training process can be offloaded to cloud clusters with abundant power and computing resources (e.g., edge servers with high-end GPUs), while model compression techniques\cite{Zhu2023ASO} and hardware evolution \cite{xiao2025understandinglargelanguagemodels} are rapidly bridging the gap between theoretical frameworks and real-world deployment. For example, recent benchmarks on consumer-grade system-on-chips (SoCs), such as Snapdragon 8 Elite Gen 5 \cite{Qualcomm2025Snapdragon8Elite} and Apple A19 \cite{Apple2025iPhone17ProA19Pro}, demonstrate the capability to execute 7–13B parameter models locally. For ultra-dense networks, as discussed in Appendix \ref{app: Scalability}, future architectures can leverage hierarchical MAC protocols \cite{MP-new} or mean-field games \cite{mean-field2, mean-filed1} to further abstract individual interactions, thereby maintaining performance stability without prohibitive computational growth.

As for communication overhead, a distinction must be drawn between internal agentic reasoning and physical transmission. While the input state space utilizes verbose textual descriptions to facilitate semantic generalization, the output signaling (i.e., UCMs and DCMs) is strictly confined by the PAG to a discrete, low-dimensional subspace. Specifically, the BS's macroscopic scheduling intent is physically transmitted as a sequence of $M$ token indices, yielding a core payload of $ M \times \lceil \log_2(I_t + 1) \rceil$. In our experimental configuration with $5$ RBGs and $5$ UEs, this corresponds to a compact payload of approximately $15$ bits. Even when scaling to larger scenarios (e.g., $M=20, I_t=20$, resulting in approximately $100$ bits), the payload remains manageable. Crucially, the base payload fits comfortably within a single 5G NR control channel element (CCE), which comprises 72 resource elements (REs) and can carry over 100 encoded bits under QPSK modulation \cite{cce1,3GPP38211}. For larger payloads, standard CCE aggregation techniques (e.g., aggregation level 2 or 4) can be employed. This implies that the signaling overhead does not require new physical channels or excessive bandwidth and can be seamlessly embedded within standard 5G NR PDCCH structures with high reliability and negligible transmission latency.

{\color{black}Given the stringent latency requirements inherent to MAC protocols, our proposed framework is primarily envisioned for offline protocol discovery and near-real-time control updates. In this deployment paradigm, the comprehensive LLM operates in the cloud or during an offline phase to discover and adapt optimal protocols \cite{10597596}. Once a stable policy emerges, its behavior can be distilled into a lightweight, specialized neural network (student) \cite{kim2025resilientllmempoweredsemanticmac,Kim2024KnowledgeDF} or a symbolic rule set \cite{prolog} that runs efficiently on specific hardware with microsecond-level latency. We have preliminarily explored this concept in Appendix \ref{app: distillation}, where our lightweight student model can effectively inherit the generalization capability of the teacher. Furthermore, complementary algorithmic optimizations (e.g., model pruning and quantization) and the integration of AI-specialized accelerators (e.g., NPUs and FPGAs \cite{chen2025characterizingmobilesocaccelerating,li2025largelanguagemodelinference}) can be leveraged to further reduce inference time and accelerate the feasibility of real-world deployment.}

\subsection{\color{black}Discussion on Semantic PAG-Constrained Protocols}
\label{app: semantic}
We clarify that a semantic protocol within this framework denotes a task-oriented representation rather than the standalone lexical interpretability of surface tokens. As detailed in Section IV-B, semantic comprehension occurs during the input phase, where raw numerical states are transformed into language-generalized prompts (Eq. 11) for contextual reasoning. Consequently, while our PAG design (Section IV-C) constrains outputs to numerical $\mathcal{W}_b^-$ and $\mathcal{W}_u^-$ tokens, these are not mere heuristic values. Rather, they are structured, executable actions distilled from the LLM's semantic reasoning. This aligns with recent LLM-empowered MAC designs, which parse natural-language contexts into structured, executable action fields \cite{kim2025resilientllmempoweredsemanticmac}, and with language-guided emergent communication, where task-grounded messages are quantized symbols rather than unconstrained text \cite{Kim2024KnowledgeDF, Semantic-paper}. Translating these emergent representations into human-readable formal logic (e.g., via ProbLog extraction \cite{prolog}) remains a distinct, orthogonal direction for post-hoc interpretability rather than a prerequisite for semantic protocol emergence.
\subsection{\color{black}Physical Justification of the Weak-Coupling Premise}
\label{app: weak premise}
While Assumption 3 mathematically structures the local geometry of the gradient field, it is fundamentally an emergent property of the proposed Stackelberg game rather than an arbitrary restriction. In conventional non-coordinated MAC protocols (e.g., ALOHA), dense contention inevitably leads to strong inter-UE coupling, rendering cross-Hessians significant. However, in our hierarchical MFSG, the leader (BS) actively orchestrates resource access by broadcasting allocation intents (DCMs). Near the equilibrium $\bm{\varTheta}^*$, the BS policy $\theta_b^*$ tend to output highly orthogonal intents to maximize system utility, and the followers' best-response policies $\theta_u^*$ are also largely aligned with these intents to maximize Eq. 2. Consequently, at $\bm{\varTheta}^*$, the agents' action probability are highly concentrated on disjoint sub-spaces of the available RBGs. Because the UEs are effectively segregated into orthogonal resources, a marginal perturbation in UE $i$'s policy near $\bm{\varTheta}^*$ has a limited impact on the marginal utility of UE $j$. Thus, the cross-coupling is substantially weakened near the equilibrium, providing an operational justification for Assumption 3.
\subsection{\color{black}Limitation}
One limitation is our reliance on a simplified OFDMA uplink baseline without complex mobility or inter-cell interference. Additionally, CSI imperfection is modeled as an aggregate BS-UE observation mismatch rather than a decomposed full physical layer process. Future work should involve comprehensive system-level simulations and explicitly parameterize estimation error, reciprocity calibration error, and feedback/processing delay to support PHY-MAC co-design analysis. Furthermore, we have not yet conducted hardware-in-the-loop (HIL) validation or full end-to-end timing profiling on commercial NPU/FPGA-based edge hardware. In practical deployment, the full LLM teacher is expected to be replaced by a distilled, quantized, or cached lightweight executor in the PDCCH slot-critical loop.

\end{document}